\documentclass[ijoc,nonblindrev]{informs3} % current default for manuscript submission

\OneAndAHalfSpacedXII % current default line spacing
\usepackage{natbib}
 \bibpunct[, ]{(}{)}{,}{a}{}{,}%
 \def\bibfont{\small}%
 \def\bibsep{\smallskipamount}%
 \def\bibhang{24pt}%
 \def\newblock{\ }%
 \def\BIBand{and}%

\TheoremsNumberedThrough     % Preferred (Theorem 1, Lemma 1, Theorem 2)
\EquationsNumberedThrough    % Default: (1), (2), ...
\usepackage{amsmath}
\usepackage[ruled, vlined,linesnumbered]{algorithm2e}
\usepackage{url}
\usepackage{gnuplot-lua-tikz}
\usepackage{tikz}
\usetikzlibrary{plotmarks}
\usetikzlibrary{decorations.text,arrows.meta,calc}
\usetikzlibrary{patterns}
\usetikzlibrary{shapes}
\usetikzlibrary{shapes.misc}
\usepackage{multirow}
\usepackage{xcolor}
\usepackage{subcaption}
\usepackage{fullpage}
\usepackage{accents}
\usepackage{makecell}
\usepackage{placeins}
\usepackage{adjustbox,lipsum}

\DeclareMathOperator*{\lexmax}{lex\,max}
\DeclareMathOperator*{\lexmin}{lex\,min}
\DeclareMathOperator*{\std}{std}
\DeclareMathOperator*{\arglexmin}{arg\,lex\,min}
\DeclareMathOperator*{\conv}{conv}
\DeclareMathOperator*{\UBFinderPoint}{UB-Finder-Point}
\DeclareMathOperator*{\UBFinderLine}{UB-Finder-Line}
\DeclareMathOperator*{\ExploreTriangle}{Explore-Triangle}
\DeclareMathOperator*{\LBFinderTriangle}{LB-Finder-Triangle}
\DeclareMathOperator*{\LBFinderRectangle}{LB-Finder-Rectangle}
\DeclareMathOperator*{\WSMethod}{Weighted-Sum-Method}
\DeclareMathOperator*{\LDetector}{Line-Detector}
\DeclareMathOperator*{\SplitTriangle}{Split-Triangle}
\DeclareMathOperator*{\FindNDP}{Find-NDP}

\newtheorem{obs}{Observation}

\usepackage{amssymb}

\begin{document}
%%%%%%%%%%%%%%%%

%\setcounter{page}{1}
%\DOI{10.1287/xxxx.0000.0000}%


%\RUNTITLE{}

% \TITLE{Bundling Information Goods of Decreasing Value}
\TITLE{Learning to Project in Multi-Objective Binary
	Linear Programming}

\ARTICLEAUTHORS{%
\AUTHOR{Alvaro Sierra-Altamiranda}
\AFF{Department of Industrial and Management System Engineering, University of South Florida, Tampa, FL, 33620 USA, \EMAIL{amsierra@mail.usf.edu},
\URL{http://www.eng.usf.edu/$\sim$amsierra/}}
\AUTHOR{Hadi Charkhgard}
\AFF{Department of Industrial and Management System Engineering, University of South Florida, Tampa, FL, 33620 USA, \EMAIL{hcharkhgard@usf.edu}, \URL{http://www.eng.usf.edu/$\sim$hcharkhgard/}}
\AUTHOR{Iman Dayarian}
\AFF{Culverhouse College of Business, The University of Alabama, Tuscaloosa, AL 35487 USA, \EMAIL{idayarian@cba.ua.edu},
	\URL{https://culverhouse.ua.edu/news/directory/iman-dayarian/}}
\AUTHOR{Ali Eshragh}
\AFF{School of Mathematical and Physical Sciences, The University of Newcastle, Callaghan, NSW 2308 Australia, \EMAIL{ali.eshragh@newcastle.edu.au},
	\URL{https://www.newcastle.edu.au/profile/ali-eshragh}}
\AUTHOR{Sorna Javadi}
\AFF{Department of Industrial and Management Systems Engineering, University of South Florida, Tampa, FL, 33620 USA, \EMAIL{javadis@mail.usf.edu}}
} % end of the block

\ABSTRACT{%
In this paper, we investigate the possibility of improving the performance of multi-objective optimization solution approaches using machine learning techniques. Specifically, we focus on multi-objective binary linear programs and employ one of the most effective and recently developed criterion space search algorithms, the so-called KSA, during our study. This algorithm computes all nondominated points of a problem with $p$ objectives by searching on a projected criterion space, i.e., a $(p-1)$-dimensional criterion apace. We present an effective and fast learning approach to identify on which projected space the KSA should work. We also present several generic features/variables that can be used in machine learning techniques for identifying the best projected space. Finally, we present an effective bi-objective optimization based heuristic for selecting the best subset of the features to overcome the issue of overfitting in learning. Through an extensive computational study over 2000 instances of tri-objective Knapsack and Assignment problems, we demonstrate that an improvement of up to 12\% in time can be achieved by the proposed learning method compared to a random selection of the projected space.}%

%\KEYWORDS{deterministic inventory theory; infinite linear programming duality; 
%  existence of optimal policies; semi-Markov decision process; cyclic schedule}

\HISTORY{}

\maketitle

	\section{Introduction}
	\label{sec:Introduction}
	Many real-life optimization problems involve multiple objective functions and they can be stated as follows:
	\begin{equation}
	\label{eq:MO}
	\min_{\boldsymbol{x} \in \mathcal{X}} \ \{z_1(\boldsymbol{x}),\dots,z_p(\boldsymbol{x})\},
	\end{equation}
	where $\mathcal{X}\subseteq\mathbb{R}^n$ represents the set of feasible solutions of the problem, and $z_1(\boldsymbol{x}),\dots, z_p(\boldsymbol{x})$ are $p$ objective functions. Because objectives are often competing in a multi-objective optimization problem, an ideal feasible solution that can optimize all objectives at the same time does not often exist in practice. Hence, when solving such a problem, the goal is often generating some (if not all) \textit{efficient} solutions, i.e., a feasible solution in which it is impossible to improve the value of one objective without making the value of any other objective worse.
	
	The focus of this study is on Multi-Objective Binary Linear Programs
	(MOBLPs), i.e., multi-objective optimization problems in which all decision variables are binary and all objective
	functions and constraints are linear. In the last few years, significant advances have been made in the development of effective
	algorithms for solving MOBLPs, see for instance \citet{Boland2015BBM,Boland2015TSM, Boland2016LSM, Boland2016QSM, Dachert2012, Dachert2013,Fattahi2017, Kirlik2013, Koksalan2010, Lokman2012, Ozlen2013, PRZYBYLSKI2017, PRZYBYLSKI2010, Soylu2016}, and \citet{Vincent2012}. Many of the recently developed algorithms fall into the category of criterion space search algorithms, i.e., those that work in the space of objective functions' values. Hence, such algorithms are specifically designed to find all \textit{nondominated points} of a multi-objective optimization problem, i.e., the image of an efficient solution in the criterion space is being referred to as a nondominated point. After computing each nondominated point, criterion space search algorithms remove the proportion of the criterion space dominated by the obtained nondominated point and search for not-yet-found nondominated points in the remaining space.
	
	In general, to solve a multi-objective optimization problem, criterion space search algorithms solve a sequence of single-objective optimization problems. Specifically, when solving a problem with $p$ objective functions, many criterion space search algorithms first attempt to transform the problem into a sequence of problems with $(p-1)$ objectives \citep{Boland2016QSM}. In other words, they attempt to compute all nondominated points by discovering their projections in a $(p-1)$-dimensional criterion space. Evidently, the same process can be applied recursively until a sequence of single-objective optimization problems are generated. For example, to solve each problem with $(p-1)$ objectives, a sequence of problems with $(p-2)$ objectives can be solved.
	
	Overall, there are at least two possible ways to apply the projection from a higher dimensional criterion space (for example $p$) to a criterion space with one less dimension (for example $p-1$):
	\begin{itemize}
		\item \textit{Weighted Sum Projection}: A typical approach used in the literature \citep[see for instance][]{Ozlen2009} is to select one of the objective functions available in the higher dimension (for example $z_1(\boldsymbol{x})$) and remove it after adding it with some strictly positive weight to the other objective functions. In this case, by imposing different bounds for $z_1(\boldsymbol{x})$ and/or the value of the other objective functions, a sequence of optimization problems with $p-1$ objectives will be generated.
		\item \textit{Lexicographical Projection}: We first note that a \textit{lexicographical optimization problem} is a two-stage optimization problem that attempts to optimize a set of objectives, the so-called \textit{secondary} objectives, over the set of solutions that are optimal for another objective, the so-called \textit{primary} objective. The first stage in the lexicographical optimization problem is a single-objective optimization problem as it optimizes the primary goal. The second stage, however, can be a multi-objective optimization problem as it optimizes the secondary objectives. Based on this definition, another typical approach (see for instance \cite{Ozlen2013}) for projection is to select one of the objective functions available in the higher dimension (for example $z_1(\boldsymbol{x})$) and simply remove it. In this case, by imposing different bounds for $z_1(\boldsymbol{x})$ and/or the value of the other objective functions, a sequence of lexicographical optimization problems should be solved in which $z_1(x)$ is the primary objective and the remaining $p-1$ objectives are secondary objectives. 
	\end{itemize} 
	
	In light of the above,  which objective function should be selected for doing a projection and how to do a projection are two typical questions that can be asked when developing a criterion space search algorithm. So, by this observation, there are many possible ways to develop a criterion space search algorithm and some of which may perform better for some instances. So, the underlying research question of this study is that whether Machine Learning (ML) techniques can help us answer the above questions for a given class of instances of a multi-objective objective optimization problem?   
	
	It is worth mentioning that, in recent years, similar questions have been asked in the field of single-objective optimization. For example, ML techniques have been successfully implemented for the purpose of \textit{variable selection} and \textit{node selection} in branch-and-bound algorithms (see for instance \citet{Khalil2016,Alvarez2017,Sabharwal2012,He2014,Khalil2017}). However, still the majority of the algorithmic/theoretical studies in the field of ML have been focused on using optimization models and algorithms to enhance ML techniques and not the other way around (see for instance \citet{Roth2005, Bottou2010, Le2011, Sra2012, Snoek2012, Bertsimas2016}). 
	
	In general, to the best of our knowledge, there are no studies in the literature that address the problem of enhancing multi-objective optimization algorithms using ML.  In this study, as the first attempt, we focus only on the simplest and most high-level question that can be asked, that is for a given instance of MOBLP with $p$ objective functions which objective should be removed for reducing the dimension of the criterion space to $p-1$ in order to minimize the solution time?
	
	It is evident that if one can show that ML is even valuable for such a high-level question then deeper questions can be asked and explored that can possibly improve the solution time significantly. In order to answer the above question, we employ one of the effective state-of-the-art algorithms in the literature of multi-objective optimization, the so-called KSA which is developed by \cite{Kirlik2013}. This algorithm uses the lexicographical projection for reducing the $p$-dimensional criterion space to $p-1$, and then it recursively reduces the dimension from $p-1$ to $1$ by using a special case of the weighted sum projection in which all the weights are equal to one.       
	
	Currently, the default objective function for conducting the projection from $p$-dimensional criterion space to $p-1$ is the first objective function (or better say random because one can change the order of the objective functions in an input file). So, a natural question is that does it really matter which objective function is selected for such a projection? To answer this question, we conducted a set of experiments by using a C++ implementation of the KSA which is publicly available in \url{http://home.ku.edu.tr/~moolibrary} and recorded the number of ILPs solved (\#ILPs) and the computational time (in seconds). We generated 1000 instances (200 per class) of tri-objective Assignment Problem (AP) and 1000 instances (200 per class) of Knapsack Problem (KP) based on the procedure described by \cite{Kirlik2013}. Table \ref{tab:MotivationTable} shows the impact of projecting based on the worst and best objective function using the KSA where \#ILPs is the number of single-objective integer linear programs solved. Numbers reported in this table are averages over 200 instances. 
	
	\begin{table}[ht]
		\caption{Projecting based on different objectives using the KSA.}
		\centering
		\resizebox{\textwidth}{!}{ 
			\begin{tabular}{cccccccccccc}
				\cline{1-3} \cline{5-6} \cline{8-9} \cline{11-12}
				\multicolumn{1}{|c|}{\multirow{2}{*}{\textbf{Type}}} & \multicolumn{1}{c|}{\multirow{2}{*}{\textbf{\#Objectives}}} & \multicolumn{1}{c|}{\multirow{2}{*}{\textbf{\#Variables}}} & \multicolumn{1}{c|}{} & \multicolumn{2}{c|}{\textbf{Projecting worst objective}} & \multicolumn{1}{c|}{} & \multicolumn{2}{c|}{\textbf{Projecting best objective}} & \multicolumn{1}{c|}{} & \multicolumn{2}{c|}{\textbf{\%Decrease}} \\ \cline{5-6} \cline{8-9} \cline{11-12} 
				\multicolumn{1}{|c|}{} & \multicolumn{1}{c|}{} & \multicolumn{1}{c|}{} & \multicolumn{1}{c|}{} & \multicolumn{1}{c|}{\textbf{Run time (s.)}} & \multicolumn{1}{c|}{\textbf{\#ILPs}} & \multicolumn{1}{c|}{} & \multicolumn{1}{c|}{\textbf{Run time (s.)}} & \multicolumn{1}{c|}{\textbf{\#ILPs}} & \multicolumn{1}{c|}{} & \multicolumn{1}{c|}{\textbf{Run time (s.)}} & \multicolumn{1}{c|}{\textbf{\#ILPs}} \\ \cline{1-3} \cline{5-6} \cline{8-9} \cline{11-12} 
				&  &  &  &  &  &  &  &  &  &  &  \\ \cline{1-3} \cline{5-6} \cline{8-9} \cline{11-12} 
				\multicolumn{1}{|c|}{\multirow{5}{*}{\textbf{AP}}} & \multicolumn{1}{c|}{\multirow{5}{*}{3}} & \multicolumn{1}{c|}{$20 \times 20$} & \multicolumn{1}{c|}{} & \multicolumn{1}{c|}{351.56} & \multicolumn{1}{c|}{5,122.67} & \multicolumn{1}{c|}{} & \multicolumn{1}{c|}{337.03} & \multicolumn{1}{c|}{5,015.39} & \multicolumn{1}{c|}{} & \multicolumn{1}{c|}{4.31\%} & \multicolumn{1}{c|}{4.61\%} \\ \cline{3-3} \cline{5-6} \cline{8-9} \cline{11-12} 
				\multicolumn{1}{|c|}{} & \multicolumn{1}{c|}{} & \multicolumn{1}{c|}{$25 \times 25$} & \multicolumn{1}{c|}{} & \multicolumn{1}{c|}{948.21} & \multicolumn{1}{c|}{9,685.69} & \multicolumn{1}{c|}{} & \multicolumn{1}{c|}{912.07} & \multicolumn{1}{c|}{9,500.52} & \multicolumn{1}{c|}{} & \multicolumn{1}{c|}{3.96\%} & \multicolumn{1}{c|}{4.40\%} \\ \cline{3-3} \cline{5-6} \cline{8-9} \cline{11-12} 
				\multicolumn{1}{|c|}{} & \multicolumn{1}{c|}{} & \multicolumn{1}{c|}{$30 \times 30$} & \multicolumn{1}{c|}{} & \multicolumn{1}{c|}{2,064.34} & \multicolumn{1}{c|}{16,294.37} & \multicolumn{1}{c|}{} & \multicolumn{1}{c|}{1,988.64} & \multicolumn{1}{c|}{16,019.64} & \multicolumn{1}{c|}{} & \multicolumn{1}{c|}{3.81\%} & \multicolumn{1}{c|}{4.01\%} \\ \cline{3-3} \cline{5-6} \cline{8-9} \cline{11-12} 
				\multicolumn{1}{|c|}{} & \multicolumn{1}{c|}{} & \multicolumn{1}{c|}{$35 \times 35$} & \multicolumn{1}{c|}{} & \multicolumn{1}{c|}{4,212.69} & \multicolumn{1}{c|}{26,161.34} & \multicolumn{1}{c|}{} & \multicolumn{1}{c|}{4,050.44} & \multicolumn{1}{c|}{25,592.13} & \multicolumn{1}{c|}{} & \multicolumn{1}{c|}{4.01\%} & \multicolumn{1}{c|}{4.27\%} \\ \cline{3-3} \cline{5-6} \cline{8-9} \cline{11-12} 
				\multicolumn{1}{|c|}{} & \multicolumn{1}{c|}{} & \multicolumn{1}{c|}{$40 \times 40$} & \multicolumn{1}{c|}{} & \multicolumn{1}{c|}{6,888.45} & \multicolumn{1}{c|}{35,737.21} & \multicolumn{1}{c|}{} & \multicolumn{1}{c|}{6,636.79} & \multicolumn{1}{c|}{35,061.41} & \multicolumn{1}{c|}{} & \multicolumn{1}{c|}{3.79\%} & \multicolumn{1}{c|}{3.97\%} \\ \cline{1-3} \cline{5-6} \cline{8-9} \cline{11-12} 
				&  &  &  &  &  &  &  &  &  &  &  \\ \cline{1-3} \cline{5-6} \cline{8-9} \cline{11-12} 
				\multicolumn{1}{|c|}{\multirow{5}{*}{\textbf{KP}}} & \multicolumn{1}{c|}{\multirow{5}{*}{3}} & \multicolumn{1}{c|}{60} & \multicolumn{1}{c|}{} & \multicolumn{1}{c|}{270.34} & \multicolumn{1}{c|}{3,883.45} & \multicolumn{1}{c|}{} & \multicolumn{1}{c|}{212.41} & \multicolumn{1}{c|}{3,861.68} & \multicolumn{1}{c|}{} & \multicolumn{1}{c|}{27.27\%} & \multicolumn{1}{c|}{1.05\%} \\ \cline{3-3} \cline{5-6} \cline{8-9} \cline{11-12} 
				\multicolumn{1}{|c|}{} & \multicolumn{1}{c|}{} & \multicolumn{1}{c|}{70} & \multicolumn{1}{c|}{} & \multicolumn{1}{c|}{813.31} & \multicolumn{1}{c|}{6,182.04} & \multicolumn{1}{c|}{} & \multicolumn{1}{c|}{638.87} & \multicolumn{1}{c|}{6,158.20} & \multicolumn{1}{c|}{} & \multicolumn{1}{c|}{27.31\%} & \multicolumn{1}{c|}{0.82\%} \\ \cline{3-3} \cline{5-6} \cline{8-9} \cline{11-12} 
				\multicolumn{1}{|c|}{} & \multicolumn{1}{c|}{} & \multicolumn{1}{c|}{80} & \multicolumn{1}{c|}{} & \multicolumn{1}{c|}{1,740.80} & \multicolumn{1}{c|}{9,297.96} & \multicolumn{1}{c|}{} & \multicolumn{1}{c|}{1,375.58} & \multicolumn{1}{c|}{9,265.09} & \multicolumn{1}{c|}{} & \multicolumn{1}{c|}{26.55\%} & \multicolumn{1}{c|}{0.74\%} \\ \cline{3-3} \cline{5-6} \cline{8-9} \cline{11-12} 
				\multicolumn{1}{|c|}{} & \multicolumn{1}{c|}{} & \multicolumn{1}{c|}{90} & \multicolumn{1}{c|}{} & \multicolumn{1}{c|}{5,109.56} & \multicolumn{1}{c|}{14,257.32} & \multicolumn{1}{c|}{} & \multicolumn{1}{c|}{3,917.32} & \multicolumn{1}{c|}{14,212.35} & \multicolumn{1}{c|}{} & \multicolumn{1}{c|}{30.44\%} & \multicolumn{1}{c|}{0.63\%} \\ \cline{3-3} \cline{5-6} \cline{8-9} \cline{11-12} 
				\multicolumn{1}{|c|}{} & \multicolumn{1}{c|}{} & \multicolumn{1}{c|}{100} & \multicolumn{1}{c|}{} & \multicolumn{1}{c|}{10,451.97} & \multicolumn{1}{c|}{19,420.06} & \multicolumn{1}{c|}{} & \multicolumn{1}{c|}{7,780.96} & \multicolumn{1}{c|}{19,366.88} & \multicolumn{1}{c|}{} & \multicolumn{1}{c|}{34.33\%} & \multicolumn{1}{c|}{0.57\%} \\ \cline{1-3} \cline{5-6} \cline{8-9} \cline{11-12} 
			\end{tabular}}
			\label{tab:MotivationTable}
		\end{table}
		
		We observe that, on average, the running time can be reduced up to $34\%$ while the \#ILPs can be improved up to $4\%$. This numerical study clearly shows the importance of projection in the solution time. Hence, it is certainly worth studying ML techniques in predicting the best objective function for projecting, the so-called \textit{learning to project}.  So, our main contribution in this study is to introduce an ML framework to simulate the selection of the best objective function to project. We collect data from each objective function and their interactions with the decision space to create features. Based on the created features, an easy-to-evaluate function is learned to emulate the classification of the projections.  Another contribution of this study is developing a simple but effective bi-objective optimization-based heuristic approach to select the best subset of features to overcome the issue of overfitting.  We show that the accuracy of the proposed prediction model can reach up to around 72\%, which represents up to 12\% improvement in solution time.  
		
		The rest of this paper is organized as follows. In Section \ref{sec:Preliminaries}, some useful concepts and notations about multi-objective optimization are introduced and also a high-level description of the KSA is given. In Section \ref{sec:Methodology}, we provide a high-level description of our proposed  machine learning framework and its three main components. In Section~\ref{sec:preorder}, the first component of the framework, which is a pre-ordering approach for changing the order of the objective functions in an input file, is explained. In Section~\ref{sec:Features}, the second component of the framework that includes features and labels are explained. In Section~\ref{sec:bestfeatures}, the third/last component of the framework, which is a bi-objective optimization based heuristic for selecting the best subset of features, is introduced. In Section \ref{sec:Experimentation}, we provide a comprehensive computational study. Finally, in Section \ref{sec:Conclusions}, we provide some concluding remarks.
		
		\section{Preliminaries}
		\label{sec:Preliminaries}
		A {\em Multi-Objective Binary Linear Program} (MOBLP) is a problem of the form \eqref{eq:MO} in which $\mathcal{X}:=\big\{\boldsymbol{x}\in \{0,1\}^{n}:A\boldsymbol{x}\le \boldsymbol{b}\big\}$ represents the \textit{feasible set in the decision space}, $A\in \mathbb{R}^{m\times n}$, and $\boldsymbol{b}\in \mathbb{R}^{m}$.  It is assumed that $\mathcal{X}$ is \textit{bounded} and $z_i(\boldsymbol{x})=\boldsymbol{c}^\intercal_i\boldsymbol{x}$  where $\boldsymbol{c}_i\in \mathbb{R}^{n}$ for $i=1,2,\dots,p$ represents a linear objective function. The image $\mathcal{Y}$ of $\mathcal{X}$ under vector-valued function
		$\boldsymbol{z}:=(z_1,z_2,\dots,z_p)^\intercal$ represents the \textit{feasible set in the objective/criterion space}, that is $\mathcal{Y} := \{\boldsymbol{o}\in \mathbb{R}^p: \boldsymbol{o}=\boldsymbol{z}(\boldsymbol{x})$ for all $\boldsymbol{x}\in \mathcal{X} \}$. Throughout this article, vectors are always column-vectors and denoted in bold font.
		
		\begin{definition}
			\label{def:nondominant}
			A feasible solution $\boldsymbol{x}\in \mathcal{X}$ is called \textit{efficient}
			or \textit{Pareto optimal}, if there is no other $\boldsymbol{x}'\in \mathcal{X}$
			such that $z_i(\boldsymbol{x}')\le z_i(\boldsymbol{x})$ for $ i=1,\dots,p$ and $\boldsymbol{z}(\boldsymbol{x}') \ne \boldsymbol{z}(\boldsymbol{x})$.
			If $\boldsymbol{x}$ is efficient, then $\boldsymbol{z}(\boldsymbol{x})$ is called a \textit{nondominated point}. The set of all efficient solutions $\boldsymbol{x} \in \mathcal{X}$ is
			denoted by $\mathcal{X}_E$. The set of all nondominated points $
			\boldsymbol{z}(\boldsymbol{x}) \in \mathcal{Y}$ for some $\boldsymbol{x} \in \mathcal{X}_E$ is denoted by
			$\mathcal{Y}_N$ and referred to as the \textit{nondominated
				frontier}.\label{def:2}
		\end{definition}
		
		Overall, multi-objective optimization is concerned with finding all nondominated points, i.e., an exact representation of the elements of $\mathcal{Y}_N$. The set of nondominated points of a MOBLPs is finite (since by assumption $\mathcal{X}$ is bounded). However, due to the existence of \textit{unsupported} nondominated points, i.e., those nondominated points that cannot be obtained by optimizing any positive weighted summation of the objective functions over the feasible set, computing all nondominated points is challenging. One of the effective criterion space search algorithms for MOBLPs is the KSA and its high-level description is provided next.

		%\subsection{The Kirlik \& Say\i n Algorithm (KSA)}
		%\label{subsec:algorithm}
		The KSA is basically a variation of the  $\varepsilon$-constraint method for generating the entire nondominated frontier of multi-objective integer linear programs. In each iteration, this algorithm solves the following lexicographical optimization problem in which the first stage is:
		
		\begin{equation*}
		\hat{\boldsymbol{x}} \in \argmin\big\{z_1(\boldsymbol{x}): \boldsymbol{x} \in \mathcal{X},\ z_i(\boldsymbol{x})\le u_i \ \ \forall i\in\{2,\dots,p\} \big\},
		\end{equation*}
		where $u_2,\dots,u_p$ are user-defined upper bounds. If $\hat{\boldsymbol{x}}$ exists, i.e., the first stage is feasible, then the following second-stage problem will be solved:
		\begin{equation*}
		\hat{\boldsymbol{x}}^*  \in \argmin\big\{\sum_{i=2}^pz_i(\boldsymbol{x}): \boldsymbol{x} \in \mathcal{X},\ z_1(\boldsymbol{x})\le z_1(\hat{\boldsymbol{x}}),\ z_i(\boldsymbol{x})\le u_i\ \ \forall i\in\{2,\dots,p\}  \big\}.
		\end{equation*}
		
		The algorithm computes all nondominated points by imposing different values on $u_2,\dots, u_p$ in each iteration. Interested readers may refer to \cite{Kirlik2013} for further details about how values of $u_2,\dots, u_p$ will be updated in each iteration. It is important to note that in the first stage users can replace the objective function $z_1(\boldsymbol{x})$ with any other arbitrary objective function, i.e.,  $z_j(\boldsymbol{x})$ where $j\in\{1,\dots.p\}$, and change the objective function of the second stage accordingly, i.e., $\sum_{i=1:\\ i\ne j}^pz_i(\boldsymbol{x})$. As shown in Introduction, on average, the running time can decrease up to $34\%$ by choosing the right objective function for the first stage. So, the goal of the proposed machine learning technique in this study is to identify the best choice.	
		
		As an aside, we note that to be consistent with our explanation of the lexicographic and/or weighted sum projections in Introduction, the lexicographic optimization problem of the KSA is presented slightly differently in this section. Specifically, \cite{Kirlik2013} use the following optimization problem instead of the second-stage problem (mentioned above):
		
		\begin{equation*}
		\hat{\boldsymbol{x}}^*  \in \argmin\big\{\sum_{i=1}^pz_i(\boldsymbol{x}): \boldsymbol{x} \in \mathcal{X},\ z_1(\boldsymbol{x})= z_1(\hat{\boldsymbol{x}}),\ z_i(\boldsymbol{x})\le u_i\ \ \forall i\in\{2,\dots,p\}  \big\}.
		\end{equation*}
		
		However, one can easily observe that these two formulations are equivalent. In other words, the lexicographic optimization problem introduced in this section is a just different representation of the one proposed by \cite{Kirlik2013}.   
		
		\section{Machine learning framework}
		\label{sec:Methodology}
		We now introduce our ML framework for learning to project in MOBLPs. Our proposed framework is based on Multi-class Support Vector Machine (MSVM). In this application, MSVM learns a function $f:\Phi\rightarrow\Omega$ from a training set to predict which objective function will have the best performance in the first stage of the KSA (for a MOBLP instance), where $\Phi$ is the feature map domain describing the MOBLP instance and $\Omega:=\{1,2,\dots,p\}$ is the domain of the labels. A label $y\in\Omega$ indicates the index of the objective function that should be selected. We do not explain MSVM in this study but interested readers may refer to \cite{Crammer2001} and \cite{Tsochantaridis2004} for details. We used the publicly available implementation of MSVM in this study which can be found in \url{https://goo.gl/4hLjyq}. It is worth mentioning that we have used MSVM mainly because it was performing well during the course of this study. In Section~\ref{subsec: RandomForest}, we also report results obtained by replacing MSVM with Random Forest \citep{breiman2001random,prinzie2008random} to show the performance of another learning technique in our proposed framework. Also, we provide more reasons in Section~\ref{subsec: RandomForest} about why MSVM is used in this study. Overall, the proposed ML framework contains three main components: 
		\begin{itemize}
			\item \textit{Component 1}: It is evident that by changing the order of the objective functions of a MOBLP instance in an input file, the instance remains the same. Therefore, in order to increase the stability of the prediction of MSVM, we propose an approach to pre-order the objective functions of each MOBLP instance in an input file before feeding it to MSVM (see Section~\ref{sec:preorder}). 
			\item \textit{Component 2}: We propose several generic features that can be used to describe each MOBLP instance. A high-level description of the features can be found in Section~\ref{sec:Features} and their detailed descriptions can be found in Appendix \ref{app:Features}. 
			\item \textit{Component 3}: We propose a bi-objective heuristic approach (see Section~\ref{sec:bestfeatures}) for selecting the best subset of features for each class of MOBLP instances (which are AP and KP in this study). Our numerical results show that our approach selects around 15\% of features based on the training set for each class of MOBLP instances in practice. Note that identifying the best subset of features is helpful for overcoming the issue of overfitting and improving the prediction accuracy \citep{ANZIAM2019,Tibshirani94}.
		\end{itemize}
		
		The proposed ML framework uses the above components for training purposes. A detailed discussion on the accuracy of the proposed framework on a testing set (for each class of MOBLP instances) is given in Section~\ref{sec:Experimentation}.  
		
		\section{A pre-ordering approach for objective functions}
		\label{sec:preorder}
		It is obvious that by changing the order of objective functions in an input file corresponding to an instance, a new instance will not be generated. In other words, only the instance is represented differently in that case and hence its nondominated frontier will remain the same. This suggests that the vector of features that will be extracted for any instance should be independent of the order of the objective functions. To address this issue, we propose to perform a pre-ordering (heuristic) approach before giving an instance to MSVM for training or testing purposes. That is, when users provide an instance, we first change its input file by re-ordering the objective functions before feeding it to the MSVM. Obviously, this somehow stabilizes the prediction accuracy of the proposed ML framework. 
		
		In light of the above, let 
		
		$$\boldsymbol{\tilde{x}}:=(\frac{1}{{\sum_{i=1}^{p}|c_{i1}|}+1},\dots, \frac{1}{{\sum_{i=1}^{p}|c_{in}|}+1}).$$ 
		
		In the proposed approach, we re-order the objective functions in an input file in a non-decreasing order of $\boldsymbol{c}_1^\intercal\boldsymbol{\tilde{x}}, \boldsymbol{c}_2^\intercal\boldsymbol{\tilde{x}},\dots,\boldsymbol{c}_p^\intercal\boldsymbol{\tilde{x}}$. Intuitively, $\boldsymbol{c}_i^\intercal\boldsymbol{\tilde{x}}$ can be viewed as the normalization score for objective function $i\in\{1,\dots, p\}$. In the rest of this paper, the vector $\boldsymbol{c}_i$ for $i=1,2,\dots,p$ is assumed to be ordered according to the proposed approach.
		%the normalized vector $\boldsymbol{\tilde{c}}:=\{\boldsymbol{o}\in\mathbb{R}^p:o_i=\boldsymbol{c}_i\boldsymbol{\tilde{x}}\}$. Finally, $\boldsymbol{\tilde{c}}$ is organized in non-increasing order to create $\boldsymbol{\hat{c}}$, i.e., $\boldsymbol{\hat{c}}=\{\tilde{c}_i\ge\dots\ge\tilde{c}_j,i,j\in\{1,2,\dots,p\}, i\ne j\}$ and the objective functions are ordered based on the correspondent values in $\boldsymbol{\tilde{c}}$.
		
		\section{Features and label describing a MOBLP instance}
		\label{sec:Features}
		This section provides a high-level explanation of the features that we create to describe a MOBLP instance and also how each instance is labeled. To the best of our knowledge, there are no studies that introduce features to describe multi-objective instances, and hence the proposed features are new. 
		
		%Basically, we compute for each instance a pair $(\boldsymbol{\phi},y)$ where $\boldsymbol{\phi}$ is the vector of features and $y$ is its corresponding label. 
		
		\subsection{Features}
		The efficiency of our proposed ML approach relies on the features describing a MOBLP instance. In other words, the features should be easy to compute and effective. Based on this observation, we create only \textit{static} features, i.e., those that are computed once using just the information provided by the MOBLP instance. Note that we only consider static features because the learning process and the decision on which objective function to select for projection (in the KSA) have to take place before solving the MOBLP instance. Overall, due to the nature of our research, most of our features describe the objective functions of the instances. We understand that the objective functions by themselves are a limited source of information to describe an instance. Therefore, we also consider establishing some relationships between the objective functions and the other characteristics of the instance in order to improve the reliability of our features.
		
		In light of the above, a total of  $5p^2+106p-50$ features are introduced for describing each instance of MOBLP.  As an aside, because in our computational study $p=3$, we have 313 features in total. Some of these features rely on the characteristics of the objective functions such as the magnitude and the number of positive, negative and zero coefficients. We also consider features that establish a relationship between the objective functions using some normalization techniques, e.g., the pre-ordering approach used to order the objective functions (see Section \ref{sec:preorder}). Other features are created based on some mathematical and statistical computations that link the objective functions with the technological coefficients and the right-hand-side values. 
		
		We also define features based on the area of the projected criterion space i.e., the corresponding ($p-1$)-dimensional criterion space,  that needs to be explored when one of the objectives is selected for conducting the projection. Note that, to compute such an area, several single-objective binary linear programming problems need to be solved. However, in order to reduce the complexity of the features extraction, we compute an approximation of the area-to-explore by optimizing the linear relaxation of the problems. Additionally, we create features in which the basic characteristics of an instance are described, e.g., size, number of variables, and number of constraints. 
		
		The main idea of the features is to generate as much information as possible in a simple way. We accomplish this by computing all the proposed features in polynomial time for a given instance. The features are normalized using a t-statistic score. Normalization is performed by aggregating subsets of features computed from a similar source. Finally, the values of the normalized feature matrix are distributed approximately between -1 and 1. Interested readers can find a detailed explanation of the features in Appendix \ref{app:Features}.
		
		\subsection{Labels}
		Based on our research goal, i.e., simulating the selection of the best objective, we propose a multi-class integer labeling scheme for each instance, where $y\in\Omega$ is the label of the instance and $\Omega=\{1,2,\dots,p\}$ is the domain of the labels. The value of $y^i$ classifies the instance $i$ with a label that indicates the index of the best objective function for projection based on the running time of the KSA (when generating the entire nondominated frontier of the instance). The label of each instance is assigned as follows:
		\begin{equation}
		\label{eq:label}
		y^i\in \argmin_{j\in \{1,\dots,p\}}\{\text{RunningTime}^i_j\},
		\end{equation}
		%\begin{align}
		%	y^i_j&=\begin{cases}
		%	1,& \text{if } \#\text{ILPs}_j= \#\text{ILPs}^* \text{ and } y^i_k=0,\ \ \ \forall k<j \\
		%	0,              & \text{otherwise}
		%	\end{cases}, &\forall\ j=1.\dots,p.
		%\end{align}
		where $\text{RunningTime}^i_j$ is the running time of the instance $i$ when the objective function $j$ is used for projecting the criterion space. 

		\section{Best subset selection of features}
		\label{sec:bestfeatures}
		It is easy to observe that by introducing more (linearly independent) features and re-training an ML model (to optimality) its prediction error, i.e., $error=1-accuracy$, on the training set decreases and eventually it becomes zero. This is because in that case we are providing a larger degree of freedom to the ML model. However, this is not necessarily the case for the testing set. In other words, by introducing more features, the ML model that will be obtained is often overfitted to the training set and does not perform well on the testing set. So, the underlying idea of the best subset selection of features is to avoid the issue of overfitting.  However, the key point is that in a real-world scenario we do not have access to the testing set. So, selecting the best subset of features should be done based the information obtained from the training set. 
		
		In light of the above, studying the trade-off between the number of features and the prediction error of an ML model on the training set is helpful for selecting the best subset of features \citep{ANZIAM2019}. However, computing such a tradeoff using exact methods is difficult in practice since the total number of subsets (of features) is an exponential function of the number of features. Therefore, in this section, we introduce a bi-objective optimization-based heuristic for selecting the best subset of features. The proposed approach has two phases: 
		\begin{itemize}
			\item \textit{Phase I}: In the first phase, the algorithm attempts to approximate the tradeoff. Specifically, the algorithm computes an approximated nondominated frontier of a bi-objective optimization problem in which its conflicting objectives are minimizing the number of features and minimizing the prediction error on the training set. 
			\item \textit{Phase II}:  In the second phase, the algorithm selects one of the approximated nondominated point and its corresponding MSVM model to be used for prediction on the testing set.
		\end{itemize}
		
		We first explain Phase I. To compute the approximated nondominated frontier, we run MSVM iteratively on a training set. In each iteration, one approximated nondominated point will be generated. The approximated nondominated point obtained in iteration $t$ is denoted by $(k_t,e_t)$ where $k_t$ is the number of features in the corresponding prediction model (obtained by MSVM) and $e_t$ is the prediction error of the corresponding model on the training set.
		
		To compute the first nondominated point, the proposed approach/algorithm assumes that all features are available and it runs the MVSM to obtain the parameters of the prediction model. We denote by  $W^t$ the parameter of the prediction model obtained by MSVM in iteration $t$. Note that $W^t$ is a $p\times k_t$ matrix where $p$ is the number of objectives. Now consider an arbitrary iteration $t$. The algorithm will explore the parameters of the prediction model obtained in the previous iteration by MSVM, i.e., $W^{t-1}$, and will remove the least important feature based on $W^{t-1}$. Hence, because of removing one feature, we have that $k_t=k_{t-1}-1$. Specifically, each column of matrix $W^{t-1}$ is associated to a feature. Therefore, the algorithm computes the standard deviation of each column independently. The feature with the minimum standard deviation will be selected and removed in iteration $t$. Note that MSVM creates a model for each objective function and that is the reason that matrix $W^{t-1}$ has $p$ rows. So, if the standard deviation of a column in the matrix is zero then we know that the corresponding feature is contributing exactly the same in all $p$ models and therefore it can be removed. So, we observe that the standard deviation plays an important role in identifying the least important feature. 
		
		Overall, after removing the least important feature, MSVM should be run again for computing $W^{t}$ and $e_t$. The algorithm for computing the approximated nondominated frontier terminates as soon as $k^t=0$.  A detailed description of the algorithm for computing the approximated nondominated frontier can be found in Algorithm \ref{alg:Alg1}.       
		
		\begin{algorithm}
			\small
			\DontPrintSemicolon
			\SetKwInOut{Input}{input}
			\Input{Training set, The set of features}
			$Queue.create(Q)$\\
			$t\leftarrow 1$\\
			$k_t\leftarrow $ The initial number of features\\
			\While{$k_t\ne 0$}{
				\If{$t\ge 1$}{
					Find the least important feature from $W^{t-1}$ and remove it from the set of features \\
					$k_t\leftarrow k_{t-1}-1$			
				}
				Compute $W^t$ by applying MSVM on the training set using the current set of features \\
				Compute $e_t$ by applying the obtained prediction model associated with $W^t$ on the training set\\	$Q.add\big((k_t,e_t)\big)$\\
				$t\leftarrow t+1$
			}
			\Return $Q$
			\caption{Phase I: Computing an approximated nondominated frontier}
			\label{alg:Alg1}
		\end{algorithm}
		In the second phase, we select an approximated nondominated point. However, before doing so, it is worth mentioning that MSVM can take a long time to compute $W^t$ in each iteration of Algorithm~\ref{alg:Alg1}. So, to avoid this issue, users usually terminate MSVM before it reaches to an optimal solution by imposing some termination conditions including a relative optimality gap tolerance and adjusting the so-called \textit{regularization} parameter (see \cite{Crammer2001} and \cite{Tsochantaridis2004} for details). In this study, we set the tolerance to 0.1 and the regularization parameter to $5\times10^4$ (since we numerically observed that MSVM performs better in this case). Such limitations obviously impact the prediction error that will be obtained on the training set, i.e., $e^t$. So, some of the points that will be reported by Algorithm \ref{alg:Alg1} may dominate each other. Therefore, in Phase II, we first remove the dominated points. In the remaining of this section we assume that there is no dominated point in the approximated nondominated frontier.   
		
		Next, the proposed approach selects an approximated nondominated point that has the minimum Euclidean distance from the (imaginary) \textit{ideal} point, i.e., an imaginary point in the criterion space that has the minimum number of features as well as the minimum prediction error. Such a technique is a special case of optimization over the frontier \citep{Abbas2006,Jorge2009,Boland2016OLF,Sierra2018}. We note that in bi-objective optimization, the ideal point can be computed easily based on the endpoints of the (approximated) nondominated frontier. Let $(k',e')$ and $(k'',e'')$ be the two endpoints in which $k'<k''$ and $e'>e''$. In this case, the ideal point is $(k',e'')$. Note too that because the first and second objectives have different scales, in this study, we first normalize all approximated nondominated points before selecting a point. Let $(k,e)$ be an arbitrary approximated nondominated point.  After normalization, this point will be as follows:
		$$
		(\frac{k-k'}{k''-k'},\frac{e-e''}{e'-e''}).
		$$
		
		Observe that the proposed normalization technique ensures that the value of each component of a point will be between 0 and 1. As a consequence, in this case, the nominalized ideal point will be always $(0,0)$. We will discuss about the effectiveness of our proposed best subset selection approach in the next section.
		\section{A computational study}
		\label{sec:Experimentation}
		In this section, we conduct an extensive computational study to evaluate the performance of the KSA when the proposed ML technique is used for learning to project. We generate 1000 tri-objective AP instances and also 1000 tri-objective KP instances based on the procedures described by \cite{Kirlik2013}. Since there are three objectives, we compute the entire representation of the nondominated frontier three times for each instance using the KSA; In each time,  a different objective function will be selected for projection. We employ CPLEX 12.7 as the single-objective binary linear programming solver. All computational experiments are carried out on a Dell PowerEdge R630 with two Intel Xeon E5-2650 2.2 GHz 12-Core Processors (30MB), 128GB RAM, and the RedHat Enterprise Linux 6.8 operating system. We only use a single thread for all experiments.
		
		Our experiments are divided into three parts. In the first part, we run our approach over the entire set of instances using 80\% of the data as the training set and 20\% of the data as the testing set. The second part evaluates the prediction models obtained in the first part on a reduced testing set. In other words, the training set is as same as the one in the first part but the testing set is smaller. Specifically, if it does not really matter which objective function to be selected for projection (in terms of solution time) for an instance in the testing set of the first part then we remove it. Obviously one can think of such instances as tie cases. In the third part of our experiments, we extend the concept of reduced testing set to the training set. That is, we remove not only the tie cases from the testing set but also from the training set. In general, the goal of reducing testing set and/or training set is to improve the overall accuracy of the prediction model.
		
		At the end of the computational study, we replace MSVM by Random Forest in the proposed ML framework to show the performance of another learning technique. We note that in this computational study, we do not report any time for our proposed ML framework because the aggregated time of generating the features, learning process, and predictions for all 1000 instances of a class of optimization problem, i.e., AP and KP, is around 50 seconds. This implies that on average almost 0.05 seconds are spent on each instance. 
		
		\subsection{Complete training and testing sets}
		\label{subsec:CompTraingTesting}
		The first part of our experiments are done on the entire training and testing sets. For each class of optimization problems, i.e., KP and AP,  the proposed subset selection approach is run on its corresponding training set. The proposed approach obtains the best subset of features and its corresponding prediction model for each class of instances. Before providing detailed explanations about the accuracy of such a prediction model on the testing set, it is necessary to show that the proposed bi-objective optimization approach for selecting the best subset of features is indeed effective. 
		
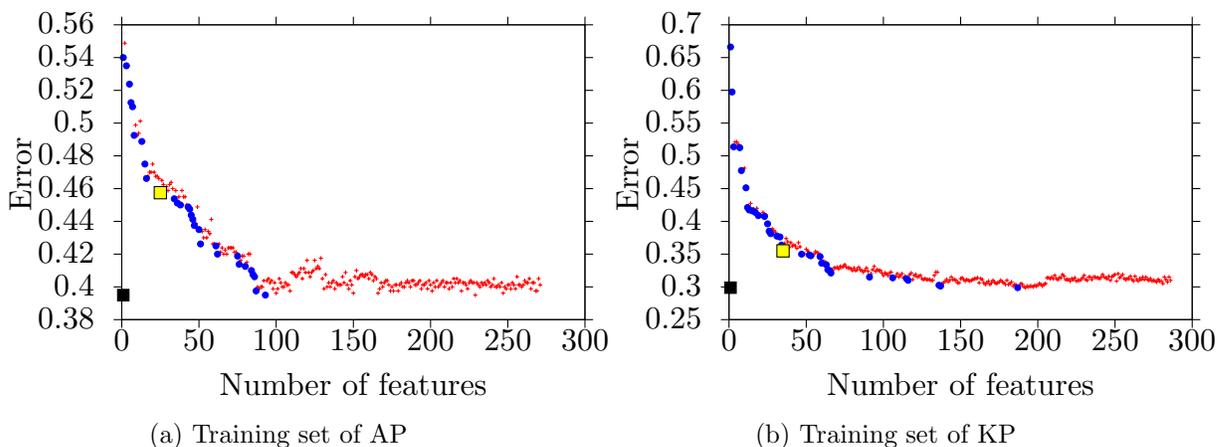
\begin{figure}[ht]
			\begin{center}
				\begin{minipage}{0.45\linewidth}
					\begin{center}
						\begin{tikzpicture}[scale=0.60,
						axis/.style={ ->, >=stealth'},line/.style={very thick},]
						\path (0.000,0.000) rectangle (12.500,8.750);
						\gpcolor{color=gp lt color border}
						\gpsetlinetype{gp lt border}
						\gpsetdashtype{gp dt solid}
						\gpsetlinewidth{1.00}
						\draw[gp path] (1.684,1.165)--(1.504,1.165);
						\draw[gp path] (11.947,1.165)--(12.127,1.165);
						\node[gp node right] at (1.320,1.165) {$0.38$};
						\draw[gp path] (1.684,1.891)--(1.504,1.891);
						\draw[gp path] (11.947,1.891)--(12.127,1.891);
						\node[gp node right] at (1.320,1.891) {$0.4$};
						\draw[gp path] (1.684,2.618)--(1.504,2.618);
						\draw[gp path] (11.947,2.618)--(12.127,2.618);
						\node[gp node right] at (1.320,2.618) {$0.42$};
						\draw[gp path] (1.684,3.344)--(1.504,3.344);
						\draw[gp path] (11.947,3.344)--(12.127,3.344);
						\node[gp node right] at (1.320,3.344) {$0.44$};
						\draw[gp path] (1.684,4.070)--(1.504,4.070);
						\draw[gp path] (11.947,4.070)--(12.127,4.070);
						\node[gp node right] at (1.320,4.070) {$0.46$};
						\draw[gp path] (1.684,4.797)--(1.504,4.797);
						\draw[gp path] (11.947,4.797)--(12.127,4.797);
						\node[gp node right] at (1.320,4.797) {$0.48$};
						\draw[gp path] (1.684,5.523)--(1.504,5.523);
						\draw[gp path] (11.947,5.523)--(12.127,5.523);
						\node[gp node right] at (1.320,5.523) {$0.5$};
						\draw[gp path] (1.684,6.249)--(1.504,6.249);
						\draw[gp path] (11.947,6.249)--(12.127,6.249);
						\node[gp node right] at (1.320,6.249) {$0.52$};
						\draw[gp path] (1.684,6.976)--(1.504,6.976);
						\draw[gp path] (11.947,6.976)--(12.127,6.976);
						\node[gp node right] at (1.320,6.976) {$0.54$};
						\draw[gp path] (1.684,7.702)--(1.504,7.702);
						\draw[gp path] (11.947,7.702)--(12.127,7.702);
						\node[gp node right] at (1.320,7.702) {$0.56$};
						\draw[gp path] (1.684,1.165)--(1.684,0.985);
						\draw[gp path] (1.684,7.702)--(1.684,7.882);
						\node[gp node center] at (1.684,0.677) {$0$};
						\draw[gp path] (3.395,1.165)--(3.395,0.985);
						\draw[gp path] (3.395,7.702)--(3.395,7.882);
						\node[gp node center] at (3.395,0.677) {$50$};
						\draw[gp path] (5.105,1.165)--(5.105,0.985);
						\draw[gp path] (5.105,7.702)--(5.105,7.882);
						\node[gp node center] at (5.105,0.677) {$100$};
						\draw[gp path] (6.816,1.165)--(6.816,0.985);
						\draw[gp path] (6.816,7.702)--(6.816,7.882);
						\node[gp node center] at (6.816,0.677) {$150$};
						\draw[gp path] (8.526,1.165)--(8.526,0.985);
						\draw[gp path] (8.526,7.702)--(8.526,7.882);
						\node[gp node center] at (8.526,0.677) {$200$};
						\draw[gp path] (10.237,1.165)--(10.237,0.985);
						\draw[gp path] (10.237,7.702)--(10.237,7.882);
						\node[gp node center] at (10.237,0.677) {$250$};
						\draw[gp path] (11.947,1.165)--(11.947,0.985);
						\draw[gp path] (11.947,7.702)--(11.947,7.882);
						\node[gp node center] at (11.947,0.677) {$300$};
						\draw[gp path] (1.684,7.702)--(1.684,1.165)--(11.947,1.165)--(11.947,7.702)--cycle;
						\node[gp node center,rotate=-270] at (-0.476,4.433) {Error};
						\node[gp node center] at (6.815,-0.315) {Number of features};
						%\node[gp node center,scale=0.8,font={\fontsize{14.0pt}{16.8pt}\selectfont}] at (6.815,8.287) {\textbf{Training set error AP instances}};
						\gpcolor{rgb color={1.000,0.000,0.000}}
						\gpsetpointsize{2.00}
						\gppoint{gp mark 1}{(1.752,7.295)}
						\gppoint{gp mark 1}{(1.821,6.794)}
						\gppoint{gp mark 1}{(1.992,5.479)}
						\gppoint{gp mark 1}{(2.026,5.251)}
						\gppoint{gp mark 1}{(2.060,5.298)}
						\gppoint{gp mark 1}{(2.095,5.567)}
						\gppoint{gp mark 1}{(2.163,5.116)}
						\gppoint{gp mark 1}{(2.266,4.343)}
						\gppoint{gp mark 1}{(2.300,4.433)}
						\gppoint{gp mark 1}{(2.334,4.433)}
						\gppoint{gp mark 1}{(2.368,4.615)}
						\gppoint{gp mark 1}{(2.402,4.433)}
						\gppoint{gp mark 1}{(2.437,4.343)}
						\gppoint{gp mark 1}{(2.471,4.343)}
						\gppoint{gp mark 1}{(2.505,4.295)}
						\gppoint{gp mark 1}{(2.573,4.252)}
						\gppoint{gp mark 1}{(2.608,4.161)}
						\gppoint{gp mark 1}{(2.642,4.027)}
						\gppoint{gp mark 1}{(2.676,4.114)}
						\gppoint{gp mark 1}{(2.710,4.027)}
						\gppoint{gp mark 1}{(2.745,4.161)}
						\gppoint{gp mark 1}{(2.779,4.208)}
						\gppoint{gp mark 1}{(2.813,4.070)}
						\gppoint{gp mark 1}{(2.881,4.027)}
						\gppoint{gp mark 1}{(2.950,3.889)}
						\gppoint{gp mark 1}{(3.018,4.027)}
						\gppoint{gp mark 1}{(3.052,3.889)}
						\gppoint{gp mark 1}{(3.087,3.889)}
						\gppoint{gp mark 1}{(3.121,3.707)}
						\gppoint{gp mark 1}{(3.326,3.300)}
						\gppoint{gp mark 1}{(3.360,3.664)}
						\gppoint{gp mark 1}{(3.463,3.119)}
						\gppoint{gp mark 1}{(3.497,2.981)}
						\gppoint{gp mark 1}{(3.531,3.162)}
						\gppoint{gp mark 1}{(3.566,2.981)}
						\gppoint{gp mark 1}{(3.600,3.119)}
						\gppoint{gp mark 1}{(3.634,3.072)}
						\gppoint{gp mark 1}{(3.668,3.388)}
						\gppoint{gp mark 1}{(3.702,2.843)}
						\gppoint{gp mark 1}{(3.737,2.843)}
						\gppoint{gp mark 1}{(3.839,2.843)}
						\gppoint{gp mark 1}{(3.873,2.661)}
						\gppoint{gp mark 1}{(3.908,2.756)}
						\gppoint{gp mark 1}{(3.942,2.618)}
						\gppoint{gp mark 1}{(3.976,2.708)}
						\gppoint{gp mark 1}{(4.010,2.618)}
						\gppoint{gp mark 1}{(4.044,2.756)}
						\gppoint{gp mark 1}{(4.079,2.756)}
						\gppoint{gp mark 1}{(4.113,2.756)}
						\gppoint{gp mark 1}{(4.147,2.756)}
						\gppoint{gp mark 1}{(4.181,2.661)}
						\gppoint{gp mark 1}{(4.216,2.661)}
						\gppoint{gp mark 1}{(4.318,2.480)}
						\gppoint{gp mark 1}{(4.352,2.574)}
						\gppoint{gp mark 1}{(4.387,2.574)}
						\gppoint{gp mark 1}{(4.455,2.436)}
						\gppoint{gp mark 1}{(4.489,2.618)}
						\gppoint{gp mark 1}{(4.523,2.618)}
						\gppoint{gp mark 1}{(4.694,1.891)}
						\gppoint{gp mark 1}{(4.729,1.848)}
						\gppoint{gp mark 1}{(4.763,1.891)}
						\gppoint{gp mark 1}{(4.797,1.891)}
						\gppoint{gp mark 1}{(4.831,2.029)}
						\gppoint{gp mark 1}{(4.900,2.029)}
						\gppoint{gp mark 1}{(4.934,2.073)}
						\gppoint{gp mark 1}{(4.968,1.982)}
						\gppoint{gp mark 1}{(5.002,2.029)}
						\gppoint{gp mark 1}{(5.037,1.848)}
						\gppoint{gp mark 1}{(5.071,1.982)}
						\gppoint{gp mark 1}{(5.105,1.753)}
						\gppoint{gp mark 1}{(5.139,1.848)}
						\gppoint{gp mark 1}{(5.173,2.073)}
						\gppoint{gp mark 1}{(5.208,1.982)}
						\gppoint{gp mark 1}{(5.242,1.891)}
						\gppoint{gp mark 1}{(5.276,1.891)}
						\gppoint{gp mark 1}{(5.310,1.982)}
						\gppoint{gp mark 1}{(5.344,1.935)}
						\gppoint{gp mark 1}{(5.379,1.935)}
						\gppoint{gp mark 1}{(5.413,1.801)}
						\gppoint{gp mark 1}{(5.447,2.211)}
						\gppoint{gp mark 1}{(5.481,2.211)}
						\gppoint{gp mark 1}{(5.516,2.164)}
						\gppoint{gp mark 1}{(5.550,2.116)}
						\gppoint{gp mark 1}{(5.584,2.164)}
						\gppoint{gp mark 1}{(5.618,2.211)}
						\gppoint{gp mark 1}{(5.652,2.298)}
						\gppoint{gp mark 1}{(5.687,2.073)}
						\gppoint{gp mark 1}{(5.721,2.298)}
						\gppoint{gp mark 1}{(5.755,2.211)}
						\gppoint{gp mark 1}{(5.789,2.480)}
						\gppoint{gp mark 1}{(5.823,2.211)}
						\gppoint{gp mark 1}{(5.858,2.116)}
						\gppoint{gp mark 1}{(5.892,2.211)}
						\gppoint{gp mark 1}{(5.926,2.254)}
						\gppoint{gp mark 1}{(5.960,2.480)}
						\gppoint{gp mark 1}{(5.994,2.298)}
						\gppoint{gp mark 1}{(6.029,2.345)}
						\gppoint{gp mark 1}{(6.063,2.298)}
						\gppoint{gp mark 1}{(6.097,2.527)}
						\gppoint{gp mark 1}{(6.131,2.073)}
						\gppoint{gp mark 1}{(6.166,1.848)}
						\gppoint{gp mark 1}{(6.200,2.164)}
						\gppoint{gp mark 1}{(6.234,1.891)}
						\gppoint{gp mark 1}{(6.268,1.982)}
						\gppoint{gp mark 1}{(6.302,1.891)}
						\gppoint{gp mark 1}{(6.337,1.982)}
						\gppoint{gp mark 1}{(6.371,2.029)}
						\gppoint{gp mark 1}{(6.405,1.848)}
						\gppoint{gp mark 1}{(6.439,1.982)}
						\gppoint{gp mark 1}{(6.473,1.935)}
						\gppoint{gp mark 1}{(6.508,1.935)}
						\gppoint{gp mark 1}{(6.542,1.935)}
						\gppoint{gp mark 1}{(6.576,2.116)}
						\gppoint{gp mark 1}{(6.610,1.982)}
						\gppoint{gp mark 1}{(6.644,2.211)}
						\gppoint{gp mark 1}{(6.679,1.935)}
						\gppoint{gp mark 1}{(6.713,1.848)}
						\gppoint{gp mark 1}{(6.747,2.029)}
						\gppoint{gp mark 1}{(6.781,1.935)}
						\gppoint{gp mark 1}{(6.816,1.753)}
						\gppoint{gp mark 1}{(6.850,2.116)}
						\gppoint{gp mark 1}{(6.884,2.254)}
						\gppoint{gp mark 1}{(6.918,2.073)}
						\gppoint{gp mark 1}{(6.952,2.116)}
						\gppoint{gp mark 1}{(6.987,2.164)}
						\gppoint{gp mark 1}{(7.021,2.164)}
						\gppoint{gp mark 1}{(7.055,2.116)}
						\gppoint{gp mark 1}{(7.089,2.164)}
						\gppoint{gp mark 1}{(7.123,2.164)}
						\gppoint{gp mark 1}{(7.158,2.073)}
						\gppoint{gp mark 1}{(7.192,2.211)}
						\gppoint{gp mark 1}{(7.226,2.164)}
						\gppoint{gp mark 1}{(7.260,2.029)}
						\gppoint{gp mark 1}{(7.294,2.116)}
						\gppoint{gp mark 1}{(7.329,2.073)}
						\gppoint{gp mark 1}{(7.363,1.753)}
						\gppoint{gp mark 1}{(7.397,1.848)}
						\gppoint{gp mark 1}{(7.431,1.982)}
						\gppoint{gp mark 1}{(7.465,1.891)}
						\gppoint{gp mark 1}{(7.500,1.982)}
						\gppoint{gp mark 1}{(7.534,2.029)}
						\gppoint{gp mark 1}{(7.568,2.029)}
						\gppoint{gp mark 1}{(7.602,1.935)}
						\gppoint{gp mark 1}{(7.637,1.801)}
						\gppoint{gp mark 1}{(7.671,1.891)}
						\gppoint{gp mark 1}{(7.705,1.801)}
						\gppoint{gp mark 1}{(7.739,1.935)}
						\gppoint{gp mark 1}{(7.773,1.935)}
						\gppoint{gp mark 1}{(7.808,1.801)}
						\gppoint{gp mark 1}{(7.842,1.935)}
						\gppoint{gp mark 1}{(7.876,1.982)}
						\gppoint{gp mark 1}{(7.910,1.891)}
						\gppoint{gp mark 1}{(7.944,1.891)}
						\gppoint{gp mark 1}{(7.979,1.982)}
						\gppoint{gp mark 1}{(8.013,2.029)}
						\gppoint{gp mark 1}{(8.047,1.982)}
						\gppoint{gp mark 1}{(8.081,1.891)}
						\gppoint{gp mark 1}{(8.115,2.029)}
						\gppoint{gp mark 1}{(8.150,2.029)}
						\gppoint{gp mark 1}{(8.184,1.935)}
						\gppoint{gp mark 1}{(8.218,1.935)}
						\gppoint{gp mark 1}{(8.252,1.982)}
						\gppoint{gp mark 1}{(8.287,2.029)}
						\gppoint{gp mark 1}{(8.321,1.848)}
						\gppoint{gp mark 1}{(8.355,1.935)}
						\gppoint{gp mark 1}{(8.389,1.935)}
						\gppoint{gp mark 1}{(8.423,1.982)}
						\gppoint{gp mark 1}{(8.458,1.848)}
						\gppoint{gp mark 1}{(8.492,2.029)}
						\gppoint{gp mark 1}{(8.526,1.935)}
						\gppoint{gp mark 1}{(8.560,1.801)}
						\gppoint{gp mark 1}{(8.594,1.982)}
						\gppoint{gp mark 1}{(8.629,1.891)}
						\gppoint{gp mark 1}{(8.663,2.029)}
						\gppoint{gp mark 1}{(8.697,1.891)}
						\gppoint{gp mark 1}{(8.731,1.935)}
						\gppoint{gp mark 1}{(8.765,1.982)}
						\gppoint{gp mark 1}{(8.800,2.029)}
						\gppoint{gp mark 1}{(8.834,1.891)}
						\gppoint{gp mark 1}{(8.868,1.982)}
						\gppoint{gp mark 1}{(8.902,1.982)}
						\gppoint{gp mark 1}{(8.937,1.848)}
						\gppoint{gp mark 1}{(8.971,1.982)}
						\gppoint{gp mark 1}{(9.005,1.891)}
						\gppoint{gp mark 1}{(9.039,2.073)}
						\gppoint{gp mark 1}{(9.073,1.982)}
						\gppoint{gp mark 1}{(9.108,2.073)}
						\gppoint{gp mark 1}{(9.142,1.982)}
						\gppoint{gp mark 1}{(9.176,2.029)}
						\gppoint{gp mark 1}{(9.210,1.801)}
						\gppoint{gp mark 1}{(9.244,1.982)}
						\gppoint{gp mark 1}{(9.279,1.982)}
						\gppoint{gp mark 1}{(9.313,1.935)}
						\gppoint{gp mark 1}{(9.347,1.848)}
						\gppoint{gp mark 1}{(9.381,1.982)}
						\gppoint{gp mark 1}{(9.415,2.029)}
						\gppoint{gp mark 1}{(9.450,1.982)}
						\gppoint{gp mark 1}{(9.484,1.982)}
						\gppoint{gp mark 1}{(9.518,1.710)}
						\gppoint{gp mark 1}{(9.552,1.982)}
						\gppoint{gp mark 1}{(9.587,1.935)}
						\gppoint{gp mark 1}{(9.621,1.891)}
						\gppoint{gp mark 1}{(9.655,1.982)}
						\gppoint{gp mark 1}{(9.689,1.935)}
						\gppoint{gp mark 1}{(9.723,1.982)}
						\gppoint{gp mark 1}{(9.758,1.935)}
						\gppoint{gp mark 1}{(9.792,2.029)}
						\gppoint{gp mark 1}{(9.826,1.982)}
						\gppoint{gp mark 1}{(9.860,1.891)}
						\gppoint{gp mark 1}{(9.894,1.935)}
						\gppoint{gp mark 1}{(9.929,2.073)}
						\gppoint{gp mark 1}{(9.963,1.891)}
						\gppoint{gp mark 1}{(9.997,1.891)}
						\gppoint{gp mark 1}{(10.031,1.935)}
						\gppoint{gp mark 1}{(10.065,1.891)}
						\gppoint{gp mark 1}{(10.100,1.982)}
						\gppoint{gp mark 1}{(10.134,1.753)}
						\gppoint{gp mark 1}{(10.168,1.891)}
						\gppoint{gp mark 1}{(10.202,1.848)}
						\gppoint{gp mark 1}{(10.237,1.935)}
						\gppoint{gp mark 1}{(10.271,1.935)}
						\gppoint{gp mark 1}{(10.305,1.848)}
						\gppoint{gp mark 1}{(10.339,1.982)}
						\gppoint{gp mark 1}{(10.373,1.935)}
						\gppoint{gp mark 1}{(10.408,1.848)}
						\gppoint{gp mark 1}{(10.442,2.116)}
						\gppoint{gp mark 1}{(10.476,1.935)}
						\gppoint{gp mark 1}{(10.510,1.935)}
						\gppoint{gp mark 1}{(10.544,1.891)}
						\gppoint{gp mark 1}{(10.579,1.935)}
						\gppoint{gp mark 1}{(10.613,2.029)}
						\gppoint{gp mark 1}{(10.647,1.982)}
						\gppoint{gp mark 1}{(10.681,1.982)}
						\gppoint{gp mark 1}{(10.715,2.073)}
						\gppoint{gp mark 1}{(10.750,1.848)}
						\gppoint{gp mark 1}{(10.784,1.982)}
						\gppoint{gp mark 1}{(10.818,1.982)}
						\gppoint{gp mark 1}{(10.852,1.801)}
						\gppoint{gp mark 1}{(10.886,1.801)}
						\gppoint{gp mark 1}{(10.921,2.073)}
						\gppoint{gp mark 1}{(10.955,1.935)}
						\gpcolor{rgb color={0.000,0.000,1.000}}
						\gpsetpointsize{2.80}
						\gppoint{gp mark 7}{(1.718,6.976)}
						\gppoint{gp mark 7}{(1.787,6.794)}
						\gppoint{gp mark 7}{(1.855,6.387)}
						\gppoint{gp mark 7}{(1.889,5.977)}
						\gppoint{gp mark 7}{(1.923,5.886)}
						\gppoint{gp mark 7}{(1.958,5.251)}
						\gppoint{gp mark 7}{(2.129,5.116)}
						\gppoint{gp mark 7}{(2.197,4.615)}
						\gppoint{gp mark 7}{(2.231,4.295)}
						\gppoint{gp mark 7}{(2.539,3.980)}
						\gppoint{gp mark 7}{(2.847,3.845)}
						\gppoint{gp mark 7}{(2.916,3.751)}
						\gppoint{gp mark 7}{(2.984,3.707)}
						\gppoint{gp mark 7}{(3.155,3.664)}
						\gppoint{gp mark 7}{(3.189,3.616)}
						\gppoint{gp mark 7}{(3.223,3.482)}
						\gppoint{gp mark 7}{(3.258,3.388)}
						\gppoint{gp mark 7}{(3.292,3.253)}
						\gppoint{gp mark 7}{(3.395,3.162)}
						\gppoint{gp mark 7}{(3.429,2.843)}
						\gppoint{gp mark 7}{(3.771,2.799)}
						\gppoint{gp mark 7}{(3.805,2.618)}
						\gppoint{gp mark 7}{(4.250,2.574)}
						\gppoint{gp mark 7}{(4.284,2.393)}
						\gppoint{gp mark 7}{(4.421,2.345)}
						\gppoint{gp mark 7}{(4.558,2.254)}
						\gppoint{gp mark 7}{(4.592,2.164)}
						\gppoint{gp mark 7}{(4.626,2.116)}
						\gppoint{gp mark 7}{(4.660,1.801)}
						\gppoint{gp mark 7}{(4.866,1.710)}
						\gpcolor{rgb color={1.000,1.000,0.000}}
						\gpsetpointsize{5.40}
						\gppoint{gp mark 5}{(2.539,3.980)}
						\gpcolor{rgb color={0.000,0.000,0.000}}
						\gpsetpointsize{5.80}
						\gppoint{gp mark 4}{(2.539,3.980)}
						\gpsetpointsize{5.40}
						\gppoint{gp mark 5}{(1.718,1.710)}
						\gpcolor{color=gp lt color border}
						\draw[gp path] (1.684,7.702)--(1.684,1.165)--(11.947,1.165)--(11.947,7.702)--cycle;
						%% coordinates of the plot area
						\gpdefrectangularnode{gp plot 1}{\pgfpoint{1.684cm}{1.165cm}}{\pgfpoint{11.947cm}{7.702cm}}
						\end{tikzpicture}
						\subcaption{Training set of AP}
						\label{fig:TF1}
					\end{center}				
				\end{minipage}
				\hfil
				\begin{minipage}{0.45\linewidth}
					\begin{center}
						\begin{tikzpicture}[scale=0.60,
						axis/.style={ ->, >=stealth'},line/.style={very thick},]
						\path (0.000,0.000) rectangle (12.500,8.750);
						\gpcolor{color=gp lt color border}
						\gpsetlinetype{gp lt border}
						\gpsetdashtype{gp dt solid}
						\gpsetlinewidth{1.00}
						\draw[gp path] (1.684,1.165)--(1.504,1.165);
						\draw[gp path] (11.947,1.165)--(12.127,1.165);
						\node[gp node right] at (1.320,1.165) {$0.25$};
						\draw[gp path] (1.684,1.891)--(1.504,1.891);
						\draw[gp path] (11.947,1.891)--(12.127,1.891);
						\node[gp node right] at (1.320,1.891) {$0.3$};
						\draw[gp path] (1.684,2.618)--(1.504,2.618);
						\draw[gp path] (11.947,2.618)--(12.127,2.618);
						\node[gp node right] at (1.320,2.618) {$0.35$};
						\draw[gp path] (1.684,3.344)--(1.504,3.344);
						\draw[gp path] (11.947,3.344)--(12.127,3.344);
						\node[gp node right] at (1.320,3.344) {$0.4$};
						\draw[gp path] (1.684,4.070)--(1.504,4.070);
						\draw[gp path] (11.947,4.070)--(12.127,4.070);
						\node[gp node right] at (1.320,4.070) {$0.45$};
						\draw[gp path] (1.684,4.797)--(1.504,4.797);
						\draw[gp path] (11.947,4.797)--(12.127,4.797);
						\node[gp node right] at (1.320,4.797) {$0.5$};
						\draw[gp path] (1.684,5.523)--(1.504,5.523);
						\draw[gp path] (11.947,5.523)--(12.127,5.523);
						\node[gp node right] at (1.320,5.523) {$0.55$};
						\draw[gp path] (1.684,6.249)--(1.504,6.249);
						\draw[gp path] (11.947,6.249)--(12.127,6.249);
						\node[gp node right] at (1.320,6.249) {$0.6$};
						\draw[gp path] (1.684,6.976)--(1.504,6.976);
						\draw[gp path] (11.947,6.976)--(12.127,6.976);
						\node[gp node right] at (1.320,6.976) {$0.65$};
						\draw[gp path] (1.684,7.702)--(1.504,7.702);
						\draw[gp path] (11.947,7.702)--(12.127,7.702);
						\node[gp node right] at (1.320,7.702) {$0.7$};
						\draw[gp path] (1.684,1.165)--(1.684,0.985);
						\draw[gp path] (1.684,7.702)--(1.684,7.882);
						\node[gp node center] at (1.684,0.677) {$0$};
						\draw[gp path] (3.395,1.165)--(3.395,0.985);
						\draw[gp path] (3.395,7.702)--(3.395,7.882);
						\node[gp node center] at (3.395,0.677) {$50$};
						\draw[gp path] (5.105,1.165)--(5.105,0.985);
						\draw[gp path] (5.105,7.702)--(5.105,7.882);
						\node[gp node center] at (5.105,0.677) {$100$};
						\draw[gp path] (6.816,1.165)--(6.816,0.985);
						\draw[gp path] (6.816,7.702)--(6.816,7.882);
						\node[gp node center] at (6.816,0.677) {$150$};
						\draw[gp path] (8.526,1.165)--(8.526,0.985);
						\draw[gp path] (8.526,7.702)--(8.526,7.882);
						\node[gp node center] at (8.526,0.677) {$200$};
						\draw[gp path] (10.237,1.165)--(10.237,0.985);
						\draw[gp path] (10.237,7.702)--(10.237,7.882);
						\node[gp node center] at (10.237,0.677) {$250$};
						\draw[gp path] (11.947,1.165)--(11.947,0.985);
						\draw[gp path] (11.947,7.702)--(11.947,7.882);
						\node[gp node center] at (11.947,0.677) {$300$};
						\draw[gp path] (1.684,7.702)--(1.684,1.165)--(11.947,1.165)--(11.947,7.702)--cycle;
						\node[gp node center,rotate=-270] at (-0.476,4.433) {Error};
						\node[gp node center] at (6.815,-0.315) {Number of features};
						%\node[gp node center,scale=0.8,font={\fontsize{14.0pt}{16.8pt}\selectfont}] at (6.815,8.287) {\textbf{Training set error KP instances}};
						\gpcolor{rgb color={1.000,0.000,0.000}}
						\gpsetpointsize{2.00}
						\gppoint{gp mark 1}{(1.821,5.105)}
						\gppoint{gp mark 1}{(1.855,5.105)}
						\gppoint{gp mark 1}{(1.889,5.051)}
						\gppoint{gp mark 1}{(1.992,4.489)}
						\gppoint{gp mark 1}{(2.026,4.524)}
						\gppoint{gp mark 1}{(2.163,3.743)}
						\gppoint{gp mark 1}{(2.231,3.617)}
						\gppoint{gp mark 1}{(2.300,3.635)}
						\gppoint{gp mark 1}{(2.368,3.489)}
						\gppoint{gp mark 1}{(2.402,3.472)}
						\gppoint{gp mark 1}{(2.437,3.544)}
						\gppoint{gp mark 1}{(2.505,3.489)}
						\gppoint{gp mark 1}{(2.642,3.181)}
						\gppoint{gp mark 1}{(2.676,3.144)}
						\gppoint{gp mark 1}{(2.710,3.109)}
						\gppoint{gp mark 1}{(2.779,3.036)}
						\gppoint{gp mark 1}{(2.916,2.853)}
						\gppoint{gp mark 1}{(2.950,2.963)}
						\gppoint{gp mark 1}{(2.984,2.891)}
						\gppoint{gp mark 1}{(3.018,2.763)}
						\gppoint{gp mark 1}{(3.052,2.872)}
						\gppoint{gp mark 1}{(3.087,2.799)}
						\gppoint{gp mark 1}{(3.121,2.818)}
						\gppoint{gp mark 1}{(3.155,2.799)}
						\gppoint{gp mark 1}{(3.189,2.746)}
						\gppoint{gp mark 1}{(3.223,2.872)}
						\gppoint{gp mark 1}{(3.258,2.780)}
						\gppoint{gp mark 1}{(3.326,2.618)}
						\gppoint{gp mark 1}{(3.360,2.618)}
						\gppoint{gp mark 1}{(3.395,2.727)}
						\gppoint{gp mark 1}{(3.429,2.690)}
						\gppoint{gp mark 1}{(3.531,2.581)}
						\gppoint{gp mark 1}{(3.566,2.618)}
						\gppoint{gp mark 1}{(3.600,2.673)}
						\gppoint{gp mark 1}{(3.634,2.654)}
						\gppoint{gp mark 1}{(3.668,2.635)}
						\gppoint{gp mark 1}{(3.771,2.455)}
						\gppoint{gp mark 1}{(3.805,2.455)}
						\gppoint{gp mark 1}{(3.976,2.255)}
						\gppoint{gp mark 1}{(4.010,2.310)}
						\gppoint{gp mark 1}{(4.044,2.310)}
						\gppoint{gp mark 1}{(4.079,2.327)}
						\gppoint{gp mark 1}{(4.113,2.291)}
						\gppoint{gp mark 1}{(4.147,2.345)}
						\gppoint{gp mark 1}{(4.181,2.327)}
						\gppoint{gp mark 1}{(4.216,2.345)}
						\gppoint{gp mark 1}{(4.250,2.382)}
						\gppoint{gp mark 1}{(4.284,2.327)}
						\gppoint{gp mark 1}{(4.318,2.310)}
						\gppoint{gp mark 1}{(4.352,2.310)}
						\gppoint{gp mark 1}{(4.387,2.291)}
						\gppoint{gp mark 1}{(4.421,2.255)}
						\gppoint{gp mark 1}{(4.455,2.310)}
						\gppoint{gp mark 1}{(4.489,2.255)}
						\gppoint{gp mark 1}{(4.523,2.272)}
						\gppoint{gp mark 1}{(4.558,2.327)}
						\gppoint{gp mark 1}{(4.592,2.272)}
						\gppoint{gp mark 1}{(4.626,2.255)}
						\gppoint{gp mark 1}{(4.660,2.218)}
						\gppoint{gp mark 1}{(4.694,2.255)}
						\gppoint{gp mark 1}{(4.729,2.310)}
						\gppoint{gp mark 1}{(4.763,2.237)}
						\gppoint{gp mark 1}{(4.831,2.182)}
						\gppoint{gp mark 1}{(4.866,2.272)}
						\gppoint{gp mark 1}{(4.900,2.345)}
						\gppoint{gp mark 1}{(4.934,2.237)}
						\gppoint{gp mark 1}{(4.968,2.272)}
						\gppoint{gp mark 1}{(5.002,2.255)}
						\gppoint{gp mark 1}{(5.037,2.255)}
						\gppoint{gp mark 1}{(5.071,2.199)}
						\gppoint{gp mark 1}{(5.105,2.146)}
						\gppoint{gp mark 1}{(5.139,2.199)}
						\gppoint{gp mark 1}{(5.173,2.182)}
						\gppoint{gp mark 1}{(5.208,2.146)}
						\gppoint{gp mark 1}{(5.242,2.199)}
						\gppoint{gp mark 1}{(5.276,2.218)}
						\gppoint{gp mark 1}{(5.344,2.092)}
						\gppoint{gp mark 1}{(5.379,2.127)}
						\gppoint{gp mark 1}{(5.413,2.182)}
						\gppoint{gp mark 1}{(5.447,2.109)}
						\gppoint{gp mark 1}{(5.481,2.146)}
						\gppoint{gp mark 1}{(5.516,2.182)}
						\gppoint{gp mark 1}{(5.550,2.092)}
						\gppoint{gp mark 1}{(5.584,2.127)}
						\gppoint{gp mark 1}{(5.687,2.164)}
						\gppoint{gp mark 1}{(5.721,2.199)}
						\gppoint{gp mark 1}{(5.755,2.054)}
						\gppoint{gp mark 1}{(5.789,2.146)}
						\gppoint{gp mark 1}{(5.823,2.164)}
						\gppoint{gp mark 1}{(5.858,2.109)}
						\gppoint{gp mark 1}{(5.892,2.127)}
						\gppoint{gp mark 1}{(5.926,2.146)}
						\gppoint{gp mark 1}{(5.960,2.073)}
						\gppoint{gp mark 1}{(5.994,2.054)}
						\gppoint{gp mark 1}{(6.029,2.073)}
						\gppoint{gp mark 1}{(6.063,2.054)}
						\gppoint{gp mark 1}{(6.097,2.054)}
						\gppoint{gp mark 1}{(6.131,2.054)}
						\gppoint{gp mark 1}{(6.166,2.037)}
						\gppoint{gp mark 1}{(6.200,2.037)}
						\gppoint{gp mark 1}{(6.234,2.146)}
						\gppoint{gp mark 1}{(6.268,2.218)}
						\gppoint{gp mark 1}{(6.302,2.218)}
						\gppoint{gp mark 1}{(6.405,1.964)}
						\gppoint{gp mark 1}{(6.439,1.981)}
						\gppoint{gp mark 1}{(6.473,2.019)}
						\gppoint{gp mark 1}{(6.508,2.054)}
						\gppoint{gp mark 1}{(6.542,2.000)}
						\gppoint{gp mark 1}{(6.576,2.037)}
						\gppoint{gp mark 1}{(6.610,2.109)}
						\gppoint{gp mark 1}{(6.644,2.054)}
						\gppoint{gp mark 1}{(6.679,2.054)}
						\gppoint{gp mark 1}{(6.713,2.037)}
						\gppoint{gp mark 1}{(6.747,1.981)}
						\gppoint{gp mark 1}{(6.781,1.981)}
						\gppoint{gp mark 1}{(6.816,2.073)}
						\gppoint{gp mark 1}{(6.850,2.092)}
						\gppoint{gp mark 1}{(6.884,2.054)}
						\gppoint{gp mark 1}{(6.918,2.019)}
						\gppoint{gp mark 1}{(6.952,2.054)}
						\gppoint{gp mark 1}{(6.987,2.037)}
						\gppoint{gp mark 1}{(7.021,2.019)}
						\gppoint{gp mark 1}{(7.055,2.037)}
						\gppoint{gp mark 1}{(7.089,2.054)}
						\gppoint{gp mark 1}{(7.123,2.073)}
						\gppoint{gp mark 1}{(7.158,2.073)}
						\gppoint{gp mark 1}{(7.192,2.092)}
						\gppoint{gp mark 1}{(7.226,1.981)}
						\gppoint{gp mark 1}{(7.260,2.073)}
						\gppoint{gp mark 1}{(7.294,2.054)}
						\gppoint{gp mark 1}{(7.329,1.981)}
						\gppoint{gp mark 1}{(7.363,1.947)}
						\gppoint{gp mark 1}{(7.397,1.981)}
						\gppoint{gp mark 1}{(7.431,2.019)}
						\gppoint{gp mark 1}{(7.465,1.947)}
						\gppoint{gp mark 1}{(7.500,1.981)}
						\gppoint{gp mark 1}{(7.534,2.037)}
						\gppoint{gp mark 1}{(7.568,2.000)}
						\gppoint{gp mark 1}{(7.602,1.964)}
						\gppoint{gp mark 1}{(7.637,1.981)}
						\gppoint{gp mark 1}{(7.671,2.000)}
						\gppoint{gp mark 1}{(7.705,2.037)}
						\gppoint{gp mark 1}{(7.739,2.000)}
						\gppoint{gp mark 1}{(7.773,1.981)}
						\gppoint{gp mark 1}{(7.808,1.947)}
						\gppoint{gp mark 1}{(7.842,2.037)}
						\gppoint{gp mark 1}{(7.876,1.947)}
						\gppoint{gp mark 1}{(7.910,1.964)}
						\gppoint{gp mark 1}{(7.944,2.000)}
						\gppoint{gp mark 1}{(7.979,1.909)}
						\gppoint{gp mark 1}{(8.013,1.947)}
						\gppoint{gp mark 1}{(8.047,1.928)}
						\gppoint{gp mark 1}{(8.115,1.909)}
						\gppoint{gp mark 1}{(8.150,1.964)}
						\gppoint{gp mark 1}{(8.184,1.947)}
						\gppoint{gp mark 1}{(8.218,1.891)}
						\gppoint{gp mark 1}{(8.252,1.874)}
						\gppoint{gp mark 1}{(8.287,1.909)}
						\gppoint{gp mark 1}{(8.321,1.891)}
						\gppoint{gp mark 1}{(8.355,1.909)}
						\gppoint{gp mark 1}{(8.389,1.909)}
						\gppoint{gp mark 1}{(8.423,1.891)}
						\gppoint{gp mark 1}{(8.458,1.928)}
						\gppoint{gp mark 1}{(8.492,1.909)}
						\gppoint{gp mark 1}{(8.526,1.928)}
						\gppoint{gp mark 1}{(8.560,1.964)}
						\gppoint{gp mark 1}{(8.594,1.947)}
						\gppoint{gp mark 1}{(8.629,1.909)}
						\gppoint{gp mark 1}{(8.663,1.947)}
						\gppoint{gp mark 1}{(8.697,1.964)}
						\gppoint{gp mark 1}{(8.731,2.092)}
						\gppoint{gp mark 1}{(8.765,2.092)}
						\gppoint{gp mark 1}{(8.800,2.073)}
						\gppoint{gp mark 1}{(8.834,2.109)}
						\gppoint{gp mark 1}{(8.868,2.073)}
						\gppoint{gp mark 1}{(8.902,2.146)}
						\gppoint{gp mark 1}{(8.937,2.073)}
						\gppoint{gp mark 1}{(8.971,2.054)}
						\gppoint{gp mark 1}{(9.005,2.092)}
						\gppoint{gp mark 1}{(9.039,2.092)}
						\gppoint{gp mark 1}{(9.073,2.164)}
						\gppoint{gp mark 1}{(9.108,2.127)}
						\gppoint{gp mark 1}{(9.142,2.054)}
						\gppoint{gp mark 1}{(9.176,2.054)}
						\gppoint{gp mark 1}{(9.210,2.054)}
						\gppoint{gp mark 1}{(9.244,2.019)}
						\gppoint{gp mark 1}{(9.279,2.073)}
						\gppoint{gp mark 1}{(9.313,2.054)}
						\gppoint{gp mark 1}{(9.347,2.092)}
						\gppoint{gp mark 1}{(9.381,2.019)}
						\gppoint{gp mark 1}{(9.415,2.037)}
						\gppoint{gp mark 1}{(9.450,2.037)}
						\gppoint{gp mark 1}{(9.484,2.054)}
						\gppoint{gp mark 1}{(9.518,2.092)}
						\gppoint{gp mark 1}{(9.552,2.146)}
						\gppoint{gp mark 1}{(9.587,2.054)}
						\gppoint{gp mark 1}{(9.621,2.037)}
						\gppoint{gp mark 1}{(9.655,2.109)}
						\gppoint{gp mark 1}{(9.689,2.109)}
						\gppoint{gp mark 1}{(9.723,2.054)}
						\gppoint{gp mark 1}{(9.758,2.109)}
						\gppoint{gp mark 1}{(9.792,2.109)}
						\gppoint{gp mark 1}{(9.826,2.092)}
						\gppoint{gp mark 1}{(9.860,2.092)}
						\gppoint{gp mark 1}{(9.894,2.073)}
						\gppoint{gp mark 1}{(9.929,2.109)}
						\gppoint{gp mark 1}{(9.963,2.164)}
						\gppoint{gp mark 1}{(9.997,2.109)}
						\gppoint{gp mark 1}{(10.031,2.092)}
						\gppoint{gp mark 1}{(10.065,2.073)}
						\gppoint{gp mark 1}{(10.100,2.127)}
						\gppoint{gp mark 1}{(10.134,2.037)}
						\gppoint{gp mark 1}{(10.168,2.109)}
						\gppoint{gp mark 1}{(10.202,2.146)}
						\gppoint{gp mark 1}{(10.237,2.164)}
						\gppoint{gp mark 1}{(10.271,2.019)}
						\gppoint{gp mark 1}{(10.305,2.182)}
						\gppoint{gp mark 1}{(10.339,2.092)}
						\gppoint{gp mark 1}{(10.373,2.127)}
						\gppoint{gp mark 1}{(10.408,2.164)}
						\gppoint{gp mark 1}{(10.442,2.073)}
						\gppoint{gp mark 1}{(10.476,2.054)}
						\gppoint{gp mark 1}{(10.510,2.073)}
						\gppoint{gp mark 1}{(10.544,2.073)}
						\gppoint{gp mark 1}{(10.579,2.127)}
						\gppoint{gp mark 1}{(10.613,2.054)}
						\gppoint{gp mark 1}{(10.647,2.019)}
						\gppoint{gp mark 1}{(10.681,2.127)}
						\gppoint{gp mark 1}{(10.715,2.054)}
						\gppoint{gp mark 1}{(10.750,2.019)}
						\gppoint{gp mark 1}{(10.784,2.073)}
						\gppoint{gp mark 1}{(10.818,2.054)}
						\gppoint{gp mark 1}{(10.852,2.109)}
						\gppoint{gp mark 1}{(10.886,2.054)}
						\gppoint{gp mark 1}{(10.921,2.109)}
						\gppoint{gp mark 1}{(10.955,2.054)}
						\gppoint{gp mark 1}{(10.989,2.037)}
						\gppoint{gp mark 1}{(11.023,2.073)}
						\gppoint{gp mark 1}{(11.058,2.019)}
						\gppoint{gp mark 1}{(11.092,2.037)}
						\gppoint{gp mark 1}{(11.126,2.019)}
						\gppoint{gp mark 1}{(11.160,2.000)}
						\gppoint{gp mark 1}{(11.194,2.054)}
						\gppoint{gp mark 1}{(11.229,2.019)}
						\gppoint{gp mark 1}{(11.263,2.019)}
						\gppoint{gp mark 1}{(11.297,1.964)}
						\gppoint{gp mark 1}{(11.331,2.109)}
						\gppoint{gp mark 1}{(11.365,2.054)}
						\gppoint{gp mark 1}{(11.400,2.000)}
						\gppoint{gp mark 1}{(11.434,2.109)}
						\gppoint{gp mark 1}{(11.468,2.037)}
						\gpcolor{rgb color={0.000,0.000,1.000}}
						\gpsetpointsize{2.80}
						\gppoint{gp mark 7}{(1.718,7.211)}
						\gppoint{gp mark 7}{(1.752,6.213)}
						\gppoint{gp mark 7}{(1.787,4.997)}
						\gppoint{gp mark 7}{(1.923,4.978)}
						\gppoint{gp mark 7}{(1.958,4.470)}
						\gppoint{gp mark 7}{(2.060,4.088)}
						\gppoint{gp mark 7}{(2.095,3.652)}
						\gppoint{gp mark 7}{(2.129,3.598)}
						\gppoint{gp mark 7}{(2.197,3.579)}
						\gppoint{gp mark 7}{(2.266,3.544)}
						\gppoint{gp mark 7}{(2.334,3.472)}
						\gppoint{gp mark 7}{(2.471,3.453)}
						\gppoint{gp mark 7}{(2.539,3.289)}
						\gppoint{gp mark 7}{(2.573,3.126)}
						\gppoint{gp mark 7}{(2.608,3.071)}
						\gppoint{gp mark 7}{(2.745,3.017)}
						\gppoint{gp mark 7}{(2.813,2.998)}
						\gppoint{gp mark 7}{(2.847,2.818)}
						\gppoint{gp mark 7}{(2.881,2.690)}
						\gppoint{gp mark 7}{(3.292,2.618)}
						\gppoint{gp mark 7}{(3.463,2.600)}
						\gppoint{gp mark 7}{(3.497,2.581)}
						\gppoint{gp mark 7}{(3.702,2.562)}
						\gppoint{gp mark 7}{(3.737,2.417)}
						\gppoint{gp mark 7}{(3.839,2.382)}
						\gppoint{gp mark 7}{(3.873,2.272)}
						\gppoint{gp mark 7}{(3.908,2.255)}
						\gppoint{gp mark 7}{(3.942,2.199)}
						\gppoint{gp mark 7}{(4.797,2.109)}
						\gppoint{gp mark 7}{(5.310,2.092)}
						\gppoint{gp mark 7}{(5.618,2.073)}
						\gppoint{gp mark 7}{(5.652,2.037)}
						\gppoint{gp mark 7}{(6.337,1.928)}
						\gppoint{gp mark 7}{(6.371,1.909)}
						\gppoint{gp mark 7}{(8.081,1.874)}
						\gpcolor{rgb color={1.000,1.000,0.000}}
						\gpsetpointsize{5.40}
						\gppoint{gp mark 5}{(2.881,2.690)}
						\gpcolor{rgb color={0.000,0.000,0.000}}
						\gpsetpointsize{5.80}
						\gppoint{gp mark 4}{(2.881,2.690)}
						\gpsetpointsize{5.40}
						\gppoint{gp mark 5}{(1.718,1.874)}
						\gpcolor{color=gp lt color border}
						\draw[gp path] (1.684,7.702)--(1.684,1.165)--(11.947,1.165)--(11.947,7.702)--cycle;
						%% coordinates of the plot area
						\gpdefrectangularnode{gp plot 1}{\pgfpoint{1.684cm}{1.165cm}}{\pgfpoint{11.947cm}{7.702cm}}
						\end{tikzpicture}
						\subcaption{Training set of KP}
						\label{fig:TF2}
					\end{center}				
				\end{minipage}
			\end{center}
			\caption{An illustration of the performance of the proposed approach for selecting the best subset of features on the complete training set} 
			\label{fig:TrainFrontier}
		\end{figure}

		Figure~\ref{fig:TrainFrontier} shows the approximated nondominated frontier (obtained during the course of the proposed best subset selection approach) for each class of optimization problems. In this figure, small (red) plus symbols are the outputs of Algorithm~\ref{alg:Alg1}. The (black) square on the vertical axis shows the ideal point and finally the (yellow) square on the approximated nondominated frontier shows the selected point by the proposed method. First note that we introduced 313 generic features in this paper but the tail of the approximated nondominated frontier in Figure~\ref{fig:TrainFrontier} clearly shows that not all 313 features are used. This is because some of the 313 features are not applicable to all classes and will be removed automatically before running the proposed best subset selection approach.  
		
		We observe from Figure~\ref{fig:TrainFrontier} that, overall, by introducing more features the prediction error deceases for the training set. It is evident when all features are included the accuracy, i.e., $1- error$, for the training set is around 59.5\% and 70\% for AP and KP instances, respectively. Of course this is not surprising because the learning procedure will be done based on the training set and introducing more features gives a larger degree of freedom to the learning model. However, this is not necessarily the case for the testing set. Basically, by introducing more features, we may raise the issue of \textit{overfitting}, i.e., the prediction error is small for the training set but large for the testing set.
		
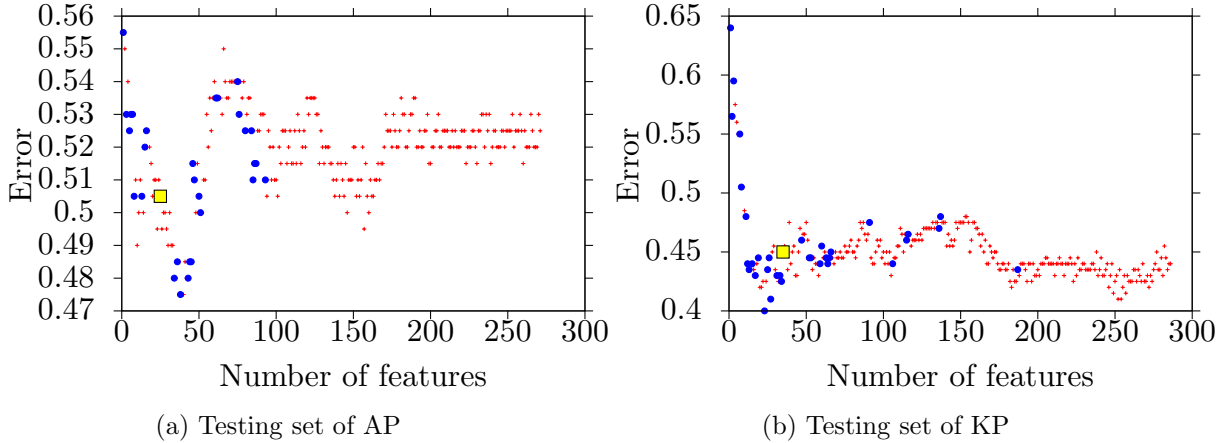
\begin{figure}[ht]
			\begin{center}
				\begin{minipage}{0.45\linewidth}
					\begin{center}
						\begin{tikzpicture}[scale=0.60,
						axis/.style={ ->, >=stealth'},line/.style={very thick},]
						\path (0.000,0.000) rectangle (12.500,8.750);
						\gpcolor{color=gp lt color border}
						\gpsetlinetype{gp lt border}
						\gpsetdashtype{gp dt solid}
						\gpsetlinewidth{1.00}
						\draw[gp path] (1.684,1.165)--(1.504,1.165);
						\draw[gp path] (11.947,1.165)--(12.127,1.165);
						\node[gp node right] at (1.320,1.165) {$0.47$};
						\draw[gp path] (1.684,1.891)--(1.504,1.891);
						\draw[gp path] (11.947,1.891)--(12.127,1.891);
						\node[gp node right] at (1.320,1.891) {$0.48$};
						\draw[gp path] (1.684,2.618)--(1.504,2.618);
						\draw[gp path] (11.947,2.618)--(12.127,2.618);
						\node[gp node right] at (1.320,2.618) {$0.49$};
						\draw[gp path] (1.684,3.344)--(1.504,3.344);
						\draw[gp path] (11.947,3.344)--(12.127,3.344);
						\node[gp node right] at (1.320,3.344) {$0.5$};
						\draw[gp path] (1.684,4.070)--(1.504,4.070);
						\draw[gp path] (11.947,4.070)--(12.127,4.070);
						\node[gp node right] at (1.320,4.070) {$0.51$};
						\draw[gp path] (1.684,4.797)--(1.504,4.797);
						\draw[gp path] (11.947,4.797)--(12.127,4.797);
						\node[gp node right] at (1.320,4.797) {$0.52$};
						\draw[gp path] (1.684,5.523)--(1.504,5.523);
						\draw[gp path] (11.947,5.523)--(12.127,5.523);
						\node[gp node right] at (1.320,5.523) {$0.53$};
						\draw[gp path] (1.684,6.249)--(1.504,6.249);
						\draw[gp path] (11.947,6.249)--(12.127,6.249);
						\node[gp node right] at (1.320,6.249) {$0.54$};
						\draw[gp path] (1.684,6.976)--(1.504,6.976);
						\draw[gp path] (11.947,6.976)--(12.127,6.976);
						\node[gp node right] at (1.320,6.976) {$0.55$};
						\draw[gp path] (1.684,7.702)--(1.504,7.702);
						\draw[gp path] (11.947,7.702)--(12.127,7.702);
						\node[gp node right] at (1.320,7.702) {$0.56$};
						\draw[gp path] (1.684,1.165)--(1.684,0.985);
						\draw[gp path] (1.684,7.702)--(1.684,7.882);
						\node[gp node center] at (1.684,0.677) {$0$};
						\draw[gp path] (3.395,1.165)--(3.395,0.985);
						\draw[gp path] (3.395,7.702)--(3.395,7.882);
						\node[gp node center] at (3.395,0.677) {$50$};
						\draw[gp path] (5.105,1.165)--(5.105,0.985);
						\draw[gp path] (5.105,7.702)--(5.105,7.882);
						\node[gp node center] at (5.105,0.677) {$100$};
						\draw[gp path] (6.816,1.165)--(6.816,0.985);
						\draw[gp path] (6.816,7.702)--(6.816,7.882);
						\node[gp node center] at (6.816,0.677) {$150$};
						\draw[gp path] (8.526,1.165)--(8.526,0.985);
						\draw[gp path] (8.526,7.702)--(8.526,7.882);
						\node[gp node center] at (8.526,0.677) {$200$};
						\draw[gp path] (10.237,1.165)--(10.237,0.985);
						\draw[gp path] (10.237,7.702)--(10.237,7.882);
						\node[gp node center] at (10.237,0.677) {$250$};
						\draw[gp path] (11.947,1.165)--(11.947,0.985);
						\draw[gp path] (11.947,7.702)--(11.947,7.882);
						\node[gp node center] at (11.947,0.677) {$300$};
						\draw[gp path] (1.684,7.702)--(1.684,1.165)--(11.947,1.165)--(11.947,7.702)--cycle;
						\node[gp node center,rotate=-270] at (-0.476,4.433) {Error};
						\node[gp node center] at (6.815,-0.315) {Number of features};
						%	 				\node[gp node center,scale=0.8,font={\fontsize{14.0pt}{16.8pt}\selectfont}] at (6.815,8.287) {\textbf{Testing set error AP instances}};
						\gpcolor{rgb color={1.000,0.000,0.000}}
						\gpsetpointsize{2.00}
						\gppoint{gp mark 1}{(1.752,6.976)}
						\gppoint{gp mark 1}{(1.821,6.249)}
						\gppoint{gp mark 1}{(1.992,4.070)}
						\gppoint{gp mark 1}{(2.026,2.618)}
						\gppoint{gp mark 1}{(2.060,3.344)}
						\gppoint{gp mark 1}{(2.095,4.070)}
						\gppoint{gp mark 1}{(2.163,3.344)}
						\gppoint{gp mark 1}{(2.266,5.160)}
						\gppoint{gp mark 1}{(2.300,4.797)}
						\gppoint{gp mark 1}{(2.334,4.433)}
						\gppoint{gp mark 1}{(2.368,3.707)}
						\gppoint{gp mark 1}{(2.402,4.070)}
						\gppoint{gp mark 1}{(2.437,4.070)}
						\gppoint{gp mark 1}{(2.471,2.981)}
						\gppoint{gp mark 1}{(2.505,4.070)}
						\gppoint{gp mark 1}{(2.573,2.981)}
						\gppoint{gp mark 1}{(2.608,3.344)}
						\gppoint{gp mark 1}{(2.642,3.344)}
						\gppoint{gp mark 1}{(2.676,2.981)}
						\gppoint{gp mark 1}{(2.710,2.618)}
						\gppoint{gp mark 1}{(2.745,3.344)}
						\gppoint{gp mark 1}{(2.779,2.618)}
						\gppoint{gp mark 1}{(2.813,2.618)}
						\gppoint{gp mark 1}{(2.881,1.891)}
						\gppoint{gp mark 1}{(2.950,2.254)}
						\gppoint{gp mark 1}{(3.018,1.528)}
						\gppoint{gp mark 1}{(3.052,1.528)}
						\gppoint{gp mark 1}{(3.087,2.254)}
						\gppoint{gp mark 1}{(3.121,1.891)}
						\gppoint{gp mark 1}{(3.326,3.344)}
						\gppoint{gp mark 1}{(3.360,4.433)}
						\gppoint{gp mark 1}{(3.463,3.344)}
						\gppoint{gp mark 1}{(3.497,4.070)}
						\gppoint{gp mark 1}{(3.531,4.070)}
						\gppoint{gp mark 1}{(3.566,5.523)}
						\gppoint{gp mark 1}{(3.600,4.797)}
						\gppoint{gp mark 1}{(3.634,5.886)}
						\gppoint{gp mark 1}{(3.668,5.160)}
						\gppoint{gp mark 1}{(3.702,5.886)}
						\gppoint{gp mark 1}{(3.737,6.249)}
						\gppoint{gp mark 1}{(3.839,5.886)}
						\gppoint{gp mark 1}{(3.873,5.886)}
						\gppoint{gp mark 1}{(3.908,5.523)}
						\gppoint{gp mark 1}{(3.942,6.976)}
						\gppoint{gp mark 1}{(3.976,6.249)}
						\gppoint{gp mark 1}{(4.010,5.886)}
						\gppoint{gp mark 1}{(4.044,5.886)}
						\gppoint{gp mark 1}{(4.079,6.249)}
						\gppoint{gp mark 1}{(4.113,6.249)}
						\gppoint{gp mark 1}{(4.147,6.249)}
						\gppoint{gp mark 1}{(4.181,6.249)}
						\gppoint{gp mark 1}{(4.216,6.249)}
						\gppoint{gp mark 1}{(4.318,6.249)}
						\gppoint{gp mark 1}{(4.352,5.886)}
						\gppoint{gp mark 1}{(4.387,6.249)}
						\gppoint{gp mark 1}{(4.455,5.523)}
						\gppoint{gp mark 1}{(4.489,5.886)}
						\gppoint{gp mark 1}{(4.523,5.886)}
						\gppoint{gp mark 1}{(4.694,5.160)}
						\gppoint{gp mark 1}{(4.729,5.160)}
						\gppoint{gp mark 1}{(4.763,5.523)}
						\gppoint{gp mark 1}{(4.797,5.523)}
						\gppoint{gp mark 1}{(4.831,5.523)}
						\gppoint{gp mark 1}{(4.900,3.707)}
						\gppoint{gp mark 1}{(4.934,5.160)}
						\gppoint{gp mark 1}{(4.968,4.797)}
						\gppoint{gp mark 1}{(5.002,4.070)}
						\gppoint{gp mark 1}{(5.037,4.797)}
						\gppoint{gp mark 1}{(5.071,4.070)}
						\gppoint{gp mark 1}{(5.105,4.070)}
						\gppoint{gp mark 1}{(5.139,3.707)}
						\gppoint{gp mark 1}{(5.173,4.797)}
						\gppoint{gp mark 1}{(5.208,4.433)}
						\gppoint{gp mark 1}{(5.242,5.160)}
						\gppoint{gp mark 1}{(5.276,5.160)}
						\gppoint{gp mark 1}{(5.310,5.160)}
						\gppoint{gp mark 1}{(5.344,4.433)}
						\gppoint{gp mark 1}{(5.379,4.797)}
						\gppoint{gp mark 1}{(5.413,5.160)}
						\gppoint{gp mark 1}{(5.447,4.433)}
						\gppoint{gp mark 1}{(5.481,4.433)}
						\gppoint{gp mark 1}{(5.516,5.160)}
						\gppoint{gp mark 1}{(5.550,4.433)}
						\gppoint{gp mark 1}{(5.584,5.523)}
						\gppoint{gp mark 1}{(5.618,5.160)}
						\gppoint{gp mark 1}{(5.652,4.433)}
						\gppoint{gp mark 1}{(5.687,4.797)}
						\gppoint{gp mark 1}{(5.721,5.160)}
						\gppoint{gp mark 1}{(5.755,4.797)}
						\gppoint{gp mark 1}{(5.789,5.886)}
						\gppoint{gp mark 1}{(5.823,5.523)}
						\gppoint{gp mark 1}{(5.858,5.886)}
						\gppoint{gp mark 1}{(5.892,5.886)}
						\gppoint{gp mark 1}{(5.926,5.886)}
						\gppoint{gp mark 1}{(5.960,5.886)}
						\gppoint{gp mark 1}{(5.994,5.523)}
						\gppoint{gp mark 1}{(6.029,5.523)}
						\gppoint{gp mark 1}{(6.063,4.433)}
						\gppoint{gp mark 1}{(6.097,5.160)}
						\gppoint{gp mark 1}{(6.131,4.797)}
						\gppoint{gp mark 1}{(6.166,4.433)}
						\gppoint{gp mark 1}{(6.200,4.797)}
						\gppoint{gp mark 1}{(6.234,4.797)}
						\gppoint{gp mark 1}{(6.268,4.797)}
						\gppoint{gp mark 1}{(6.302,4.433)}
						\gppoint{gp mark 1}{(6.337,4.433)}
						\gppoint{gp mark 1}{(6.371,4.070)}
						\gppoint{gp mark 1}{(6.405,3.707)}
						\gppoint{gp mark 1}{(6.439,4.070)}
						\gppoint{gp mark 1}{(6.473,4.433)}
						\gppoint{gp mark 1}{(6.508,3.707)}
						\gppoint{gp mark 1}{(6.542,4.433)}
						\gppoint{gp mark 1}{(6.576,4.070)}
						\gppoint{gp mark 1}{(6.610,3.707)}
						\gppoint{gp mark 1}{(6.644,4.433)}
						\gppoint{gp mark 1}{(6.679,3.344)}
						\gppoint{gp mark 1}{(6.713,4.070)}
						\gppoint{gp mark 1}{(6.747,4.070)}
						\gppoint{gp mark 1}{(6.781,4.070)}
						\gppoint{gp mark 1}{(6.816,4.797)}
						\gppoint{gp mark 1}{(6.850,4.797)}
						\gppoint{gp mark 1}{(6.884,3.707)}
						\gppoint{gp mark 1}{(6.918,4.070)}
						\gppoint{gp mark 1}{(6.952,3.707)}
						\gppoint{gp mark 1}{(6.987,4.070)}
						\gppoint{gp mark 1}{(7.021,4.797)}
						\gppoint{gp mark 1}{(7.055,2.981)}
						\gppoint{gp mark 1}{(7.089,4.433)}
						\gppoint{gp mark 1}{(7.123,3.707)}
						\gppoint{gp mark 1}{(7.158,3.344)}
						\gppoint{gp mark 1}{(7.192,4.070)}
						\gppoint{gp mark 1}{(7.226,3.707)}
						\gppoint{gp mark 1}{(7.260,3.707)}
						\gppoint{gp mark 1}{(7.294,4.070)}
						\gppoint{gp mark 1}{(7.329,4.070)}
						\gppoint{gp mark 1}{(7.363,4.433)}
						\gppoint{gp mark 1}{(7.397,4.070)}
						\gppoint{gp mark 1}{(7.431,4.433)}
						\gppoint{gp mark 1}{(7.465,4.433)}
						\gppoint{gp mark 1}{(7.500,4.797)}
						\gppoint{gp mark 1}{(7.534,5.523)}
						\gppoint{gp mark 1}{(7.568,5.160)}
						\gppoint{gp mark 1}{(7.602,5.160)}
						\gppoint{gp mark 1}{(7.637,5.160)}
						\gppoint{gp mark 1}{(7.671,4.797)}
						\gppoint{gp mark 1}{(7.705,5.523)}
						\gppoint{gp mark 1}{(7.739,5.160)}
						\gppoint{gp mark 1}{(7.773,4.797)}
						\gppoint{gp mark 1}{(7.808,4.797)}
						\gppoint{gp mark 1}{(7.842,5.160)}
						\gppoint{gp mark 1}{(7.876,5.886)}
						\gppoint{gp mark 1}{(7.910,5.523)}
						\gppoint{gp mark 1}{(7.944,5.523)}
						\gppoint{gp mark 1}{(7.979,4.797)}
						\gppoint{gp mark 1}{(8.013,4.797)}
						\gppoint{gp mark 1}{(8.047,5.160)}
						\gppoint{gp mark 1}{(8.081,5.523)}
						\gppoint{gp mark 1}{(8.115,5.523)}
						\gppoint{gp mark 1}{(8.150,5.886)}
						\gppoint{gp mark 1}{(8.184,5.886)}
						\gppoint{gp mark 1}{(8.218,4.797)}
						\gppoint{gp mark 1}{(8.252,4.797)}
						\gppoint{gp mark 1}{(8.287,5.160)}
						\gppoint{gp mark 1}{(8.321,4.797)}
						\gppoint{gp mark 1}{(8.355,4.797)}
						\gppoint{gp mark 1}{(8.389,5.160)}
						\gppoint{gp mark 1}{(8.423,5.160)}
						\gppoint{gp mark 1}{(8.458,5.523)}
						\gppoint{gp mark 1}{(8.492,4.433)}
						\gppoint{gp mark 1}{(8.526,4.433)}
						\gppoint{gp mark 1}{(8.560,4.797)}
						\gppoint{gp mark 1}{(8.594,4.797)}
						\gppoint{gp mark 1}{(8.629,5.160)}
						\gppoint{gp mark 1}{(8.663,5.160)}
						\gppoint{gp mark 1}{(8.697,5.523)}
						\gppoint{gp mark 1}{(8.731,4.797)}
						\gppoint{gp mark 1}{(8.765,4.797)}
						\gppoint{gp mark 1}{(8.800,5.160)}
						\gppoint{gp mark 1}{(8.834,5.160)}
						\gppoint{gp mark 1}{(8.868,5.160)}
						\gppoint{gp mark 1}{(8.902,5.160)}
						\gppoint{gp mark 1}{(8.937,4.797)}
						\gppoint{gp mark 1}{(8.971,4.797)}
						\gppoint{gp mark 1}{(9.005,5.160)}
						\gppoint{gp mark 1}{(9.039,4.797)}
						\gppoint{gp mark 1}{(9.073,5.160)}
						\gppoint{gp mark 1}{(9.108,4.797)}
						\gppoint{gp mark 1}{(9.142,4.797)}
						\gppoint{gp mark 1}{(9.176,4.797)}
						\gppoint{gp mark 1}{(9.210,5.160)}
						\gppoint{gp mark 1}{(9.244,5.160)}
						\gppoint{gp mark 1}{(9.279,5.160)}
						\gppoint{gp mark 1}{(9.313,4.433)}
						\gppoint{gp mark 1}{(9.347,4.797)}
						\gppoint{gp mark 1}{(9.381,4.797)}
						\gppoint{gp mark 1}{(9.415,4.797)}
						\gppoint{gp mark 1}{(9.450,5.160)}
						\gppoint{gp mark 1}{(9.484,5.160)}
						\gppoint{gp mark 1}{(9.518,4.797)}
						\gppoint{gp mark 1}{(9.552,4.433)}
						\gppoint{gp mark 1}{(9.587,5.160)}
						\gppoint{gp mark 1}{(9.621,5.160)}
						\gppoint{gp mark 1}{(9.655,5.523)}
						\gppoint{gp mark 1}{(9.689,5.523)}
						\gppoint{gp mark 1}{(9.723,4.797)}
						\gppoint{gp mark 1}{(9.758,5.523)}
						\gppoint{gp mark 1}{(9.792,5.160)}
						\gppoint{gp mark 1}{(9.826,5.523)}
						\gppoint{gp mark 1}{(9.860,4.433)}
						\gppoint{gp mark 1}{(9.894,4.797)}
						\gppoint{gp mark 1}{(9.929,5.160)}
						\gppoint{gp mark 1}{(9.963,5.160)}
						\gppoint{gp mark 1}{(9.997,4.797)}
						\gppoint{gp mark 1}{(10.031,4.797)}
						\gppoint{gp mark 1}{(10.065,4.797)}
						\gppoint{gp mark 1}{(10.100,5.160)}
						\gppoint{gp mark 1}{(10.134,5.160)}
						\gppoint{gp mark 1}{(10.168,4.797)}
						\gppoint{gp mark 1}{(10.202,5.160)}
						\gppoint{gp mark 1}{(10.237,5.160)}
						\gppoint{gp mark 1}{(10.271,5.160)}
						\gppoint{gp mark 1}{(10.305,5.160)}
						\gppoint{gp mark 1}{(10.339,5.523)}
						\gppoint{gp mark 1}{(10.373,4.797)}
						\gppoint{gp mark 1}{(10.408,5.160)}
						\gppoint{gp mark 1}{(10.442,5.160)}
						\gppoint{gp mark 1}{(10.476,5.160)}
						\gppoint{gp mark 1}{(10.510,5.523)}
						\gppoint{gp mark 1}{(10.544,5.160)}
						\gppoint{gp mark 1}{(10.579,4.433)}
						\gppoint{gp mark 1}{(10.613,5.160)}
						\gppoint{gp mark 1}{(10.647,4.797)}
						\gppoint{gp mark 1}{(10.681,4.797)}
						\gppoint{gp mark 1}{(10.715,4.797)}
						\gppoint{gp mark 1}{(10.750,5.523)}
						\gppoint{gp mark 1}{(10.784,5.160)}
						\gppoint{gp mark 1}{(10.818,4.797)}
						\gppoint{gp mark 1}{(10.852,4.797)}
						\gppoint{gp mark 1}{(10.886,4.797)}
						\gppoint{gp mark 1}{(10.921,5.523)}
						\gppoint{gp mark 1}{(10.955,5.160)}
						\gpsetpointsize{2.80}
						\gpcolor{rgb color={0.000,0.000,1.000}}
						\gppoint{gp mark 7}{(1.718,7.339)}
						\gppoint{gp mark 7}{(1.787,5.523)}
						\gppoint{gp mark 7}{(1.855,5.160)}
						\gppoint{gp mark 7}{(1.889,5.523)}
						\gppoint{gp mark 7}{(1.923,5.523)}
						\gppoint{gp mark 7}{(1.958,3.707)}
						\gppoint{gp mark 7}{(2.129,3.707)}
						\gppoint{gp mark 7}{(2.197,4.797)}
						\gppoint{gp mark 7}{(2.231,5.160)}
						\gppoint{gp mark 7}{(2.539,3.707)}
						\gppoint{gp mark 7}{(2.847,1.891)}
						\gppoint{gp mark 7}{(2.916,2.254)}
						\gppoint{gp mark 7}{(2.984,1.528)}
						\gppoint{gp mark 7}{(3.155,1.891)}
						\gppoint{gp mark 7}{(3.189,2.254)}
						\gppoint{gp mark 7}{(3.223,2.254)}
						\gppoint{gp mark 7}{(3.258,4.433)}
						\gppoint{gp mark 7}{(3.292,4.070)}
						\gppoint{gp mark 7}{(3.395,3.707)}
						\gppoint{gp mark 7}{(3.429,3.344)}
						\gppoint{gp mark 7}{(3.771,5.886)}
						\gppoint{gp mark 7}{(3.805,5.886)}
						\gppoint{gp mark 7}{(4.250,6.249)}
						\gppoint{gp mark 7}{(4.284,5.523)}
						\gppoint{gp mark 7}{(4.421,5.160)}
						\gppoint{gp mark 7}{(4.558,5.160)}
						\gppoint{gp mark 7}{(4.592,4.070)}
						\gppoint{gp mark 7}{(4.626,4.433)}
						\gppoint{gp mark 7}{(4.660,4.433)}
						\gppoint{gp mark 7}{(4.866,4.070)}
						\gpcolor{rgb color={1.000,1.000,0.000}}
						\gpsetpointsize{5.40}
						\gppoint{gp mark 5}{(2.539,3.707)}
						\gpcolor{rgb color={0.000,0.000,0.000}}
						\gpsetpointsize{5.80}
						\gppoint{gp mark 4}{(2.539,3.707)}
						\gpcolor{color=gp lt color border}
						\draw[gp path] (1.684,7.702)--(1.684,1.165)--(11.947,1.165)--(11.947,7.702)--cycle;
						%% coordinates of the plot area
						\gpdefrectangularnode{gp plot 1}{\pgfpoint{1.684cm}{1.165cm}}{\pgfpoint{11.947cm}{7.702cm}}
						\end{tikzpicture}
						\subcaption{Testing set of AP}
						\label{fig:TestF1}
					\end{center}				
				\end{minipage}
				\hfil
				\begin{minipage}{0.45\linewidth}
					\begin{center}
						\begin{tikzpicture}[scale=0.60,
						axis/.style={ ->, >=stealth'},line/.style={very thick},]
						\path (0.000,0.000) rectangle (12.500,8.750);
						\gpcolor{color=gp lt color border}
						\gpsetlinetype{gp lt border}
						\gpsetdashtype{gp dt solid}
						\gpsetlinewidth{1.00}
						\draw[gp path] (1.684,1.165)--(1.504,1.165);
						\draw[gp path] (11.947,1.165)--(12.127,1.165);
						\node[gp node right] at (1.320,1.165) {$0.4$};
						\draw[gp path] (1.684,2.472)--(1.504,2.472);
						\draw[gp path] (11.947,2.472)--(12.127,2.472);
						\node[gp node right] at (1.320,2.472) {$0.45$};
						\draw[gp path] (1.684,3.780)--(1.504,3.780);
						\draw[gp path] (11.947,3.780)--(12.127,3.780);
						\node[gp node right] at (1.320,3.780) {$0.5$};
						\draw[gp path] (1.684,5.087)--(1.504,5.087);
						\draw[gp path] (11.947,5.087)--(12.127,5.087);
						\node[gp node right] at (1.320,5.087) {$0.55$};
						\draw[gp path] (1.684,6.395)--(1.504,6.395);
						\draw[gp path] (11.947,6.395)--(12.127,6.395);
						\node[gp node right] at (1.320,6.395) {$0.6$};
						\draw[gp path] (1.684,7.702)--(1.504,7.702);
						\draw[gp path] (11.947,7.702)--(12.127,7.702);
						\node[gp node right] at (1.320,7.702) {$0.65$};
						\draw[gp path] (1.684,1.165)--(1.684,0.985);
						\draw[gp path] (1.684,7.702)--(1.684,7.882);
						\node[gp node center] at (1.684,0.677) {$0$};
						\draw[gp path] (3.395,1.165)--(3.395,0.985);
						\draw[gp path] (3.395,7.702)--(3.395,7.882);
						\node[gp node center] at (3.395,0.677) {$50$};
						\draw[gp path] (5.105,1.165)--(5.105,0.985);
						\draw[gp path] (5.105,7.702)--(5.105,7.882);
						\node[gp node center] at (5.105,0.677) {$100$};
						\draw[gp path] (6.816,1.165)--(6.816,0.985);
						\draw[gp path] (6.816,7.702)--(6.816,7.882);
						\node[gp node center] at (6.816,0.677) {$150$};
						\draw[gp path] (8.526,1.165)--(8.526,0.985);
						\draw[gp path] (8.526,7.702)--(8.526,7.882);
						\node[gp node center] at (8.526,0.677) {$200$};
						\draw[gp path] (10.237,1.165)--(10.237,0.985);
						\draw[gp path] (10.237,7.702)--(10.237,7.882);
						\node[gp node center] at (10.237,0.677) {$250$};
						\draw[gp path] (11.947,1.165)--(11.947,0.985);
						\draw[gp path] (11.947,7.702)--(11.947,7.882);
						\node[gp node center] at (11.947,0.677) {$300$};
						\draw[gp path] (1.684,7.702)--(1.684,1.165)--(11.947,1.165)--(11.947,7.702)--cycle;
						\node[gp node center,rotate=-270] at (-0.476,4.433) {Error};
						\node[gp node center] at (6.815,-0.315) {Number of features};
						%	 				\node[gp node center,scale=0.8,font={\fontsize{14.0pt}{16.8pt}\selectfont}] at (6.815,8.287) {\textbf{Testing set error KP instances}};
						\gpcolor{rgb color={1.000,0.000,0.000}}
						\gpsetpointsize{2.00}
						\gppoint{gp mark 1}{(1.821,5.741)}
						\gppoint{gp mark 1}{(1.855,5.349)}
						\gppoint{gp mark 1}{(1.889,5.087)}
						\gppoint{gp mark 1}{(1.992,3.911)}
						\gppoint{gp mark 1}{(2.026,3.388)}
						\gppoint{gp mark 1}{(2.163,2.211)}
						\gppoint{gp mark 1}{(2.231,2.080)}
						\gppoint{gp mark 1}{(2.300,2.211)}
						\gppoint{gp mark 1}{(2.368,1.688)}
						\gppoint{gp mark 1}{(2.402,1.688)}
						\gppoint{gp mark 1}{(2.437,1.819)}
						\gppoint{gp mark 1}{(2.505,1.819)}
						\gppoint{gp mark 1}{(2.642,2.472)}
						\gppoint{gp mark 1}{(2.676,2.603)}
						\gppoint{gp mark 1}{(2.710,2.080)}
						\gppoint{gp mark 1}{(2.779,1.819)}
						\gppoint{gp mark 1}{(2.916,2.603)}
						\gppoint{gp mark 1}{(2.950,2.472)}
						\gppoint{gp mark 1}{(2.984,2.080)}
						\gppoint{gp mark 1}{(3.018,3.126)}
						\gppoint{gp mark 1}{(3.052,2.211)}
						\gppoint{gp mark 1}{(3.087,2.472)}
						\gppoint{gp mark 1}{(3.121,2.603)}
						\gppoint{gp mark 1}{(3.155,1.949)}
						\gppoint{gp mark 1}{(3.189,1.949)}
						\gppoint{gp mark 1}{(3.223,2.472)}
						\gppoint{gp mark 1}{(3.258,2.995)}
						\gppoint{gp mark 1}{(3.326,2.865)}
						\gppoint{gp mark 1}{(3.360,2.865)}
						\gppoint{gp mark 1}{(3.395,3.126)}
						\gppoint{gp mark 1}{(3.429,2.734)}
						\gppoint{gp mark 1}{(3.531,2.080)}
						\gppoint{gp mark 1}{(3.566,2.342)}
						\gppoint{gp mark 1}{(3.600,2.211)}
						\gppoint{gp mark 1}{(3.634,2.211)}
						\gppoint{gp mark 1}{(3.668,1.949)}
						\gppoint{gp mark 1}{(3.771,2.342)}
						\gppoint{gp mark 1}{(3.805,2.472)}
						\gppoint{gp mark 1}{(3.976,2.342)}
						\gppoint{gp mark 1}{(4.010,2.211)}
						\gppoint{gp mark 1}{(4.044,2.472)}
						\gppoint{gp mark 1}{(4.079,2.342)}
						\gppoint{gp mark 1}{(4.113,2.342)}
						\gppoint{gp mark 1}{(4.147,2.472)}
						\gppoint{gp mark 1}{(4.181,2.342)}
						\gppoint{gp mark 1}{(4.216,2.342)}
						\gppoint{gp mark 1}{(4.250,2.603)}
						\gppoint{gp mark 1}{(4.284,2.472)}
						\gppoint{gp mark 1}{(4.318,2.472)}
						\gppoint{gp mark 1}{(4.352,2.603)}
						\gppoint{gp mark 1}{(4.387,2.734)}
						\gppoint{gp mark 1}{(4.421,2.472)}
						\gppoint{gp mark 1}{(4.455,2.472)}
						\gppoint{gp mark 1}{(4.489,2.342)}
						\gppoint{gp mark 1}{(4.523,2.603)}
						\gppoint{gp mark 1}{(4.558,2.734)}
						\gppoint{gp mark 1}{(4.592,3.126)}
						\gppoint{gp mark 1}{(4.626,2.995)}
						\gppoint{gp mark 1}{(4.660,3.126)}
						\gppoint{gp mark 1}{(4.694,2.865)}
						\gppoint{gp mark 1}{(4.729,2.734)}
						\gppoint{gp mark 1}{(4.763,2.865)}
						\gppoint{gp mark 1}{(4.831,2.342)}
						\gppoint{gp mark 1}{(4.866,2.211)}
						\gppoint{gp mark 1}{(4.900,2.211)}
						\gppoint{gp mark 1}{(4.934,2.472)}
						\gppoint{gp mark 1}{(4.968,2.734)}
						\gppoint{gp mark 1}{(5.002,2.603)}
						\gppoint{gp mark 1}{(5.037,2.734)}
						\gppoint{gp mark 1}{(5.071,2.342)}
						\gppoint{gp mark 1}{(5.105,2.472)}
						\gppoint{gp mark 1}{(5.139,2.211)}
						\gppoint{gp mark 1}{(5.173,2.472)}
						\gppoint{gp mark 1}{(5.208,2.211)}
						\gppoint{gp mark 1}{(5.242,2.472)}
						\gppoint{gp mark 1}{(5.276,1.949)}
						\gppoint{gp mark 1}{(5.344,2.342)}
						\gppoint{gp mark 1}{(5.379,2.342)}
						\gppoint{gp mark 1}{(5.413,2.603)}
						\gppoint{gp mark 1}{(5.447,2.603)}
						\gppoint{gp mark 1}{(5.481,2.995)}
						\gppoint{gp mark 1}{(5.516,2.865)}
						\gppoint{gp mark 1}{(5.550,2.472)}
						\gppoint{gp mark 1}{(5.584,2.865)}
						\gppoint{gp mark 1}{(5.687,2.865)}
						\gppoint{gp mark 1}{(5.721,2.603)}
						\gppoint{gp mark 1}{(5.755,2.734)}
						\gppoint{gp mark 1}{(5.789,2.734)}
						\gppoint{gp mark 1}{(5.823,2.734)}
						\gppoint{gp mark 1}{(5.858,2.865)}
						\gppoint{gp mark 1}{(5.892,2.734)}
						\gppoint{gp mark 1}{(5.926,2.865)}
						\gppoint{gp mark 1}{(5.960,2.734)}
						\gppoint{gp mark 1}{(5.994,2.995)}
						\gppoint{gp mark 1}{(6.029,2.995)}
						\gppoint{gp mark 1}{(6.063,2.995)}
						\gppoint{gp mark 1}{(6.097,2.995)}
						\gppoint{gp mark 1}{(6.131,2.995)}
						\gppoint{gp mark 1}{(6.166,2.995)}
						\gppoint{gp mark 1}{(6.200,3.126)}
						\gppoint{gp mark 1}{(6.234,2.995)}
						\gppoint{gp mark 1}{(6.268,3.126)}
						\gppoint{gp mark 1}{(6.302,3.126)}
						\gppoint{gp mark 1}{(6.405,3.126)}
						\gppoint{gp mark 1}{(6.439,3.126)}
						\gppoint{gp mark 1}{(6.473,3.126)}
						\gppoint{gp mark 1}{(6.508,2.603)}
						\gppoint{gp mark 1}{(6.542,2.865)}
						\gppoint{gp mark 1}{(6.576,2.603)}
						\gppoint{gp mark 1}{(6.610,2.734)}
						\gppoint{gp mark 1}{(6.644,2.603)}
						\gppoint{gp mark 1}{(6.679,2.734)}
						\gppoint{gp mark 1}{(6.713,3.126)}
						\gppoint{gp mark 1}{(6.747,2.995)}
						\gppoint{gp mark 1}{(6.781,3.126)}
						\gppoint{gp mark 1}{(6.816,3.126)}
						\gppoint{gp mark 1}{(6.850,3.126)}
						\gppoint{gp mark 1}{(6.884,3.126)}
						\gppoint{gp mark 1}{(6.918,3.257)}
						\gppoint{gp mark 1}{(6.952,3.257)}
						\gppoint{gp mark 1}{(6.987,3.126)}
						\gppoint{gp mark 1}{(7.021,2.472)}
						\gppoint{gp mark 1}{(7.055,2.734)}
						\gppoint{gp mark 1}{(7.089,3.126)}
						\gppoint{gp mark 1}{(7.123,3.126)}
						\gppoint{gp mark 1}{(7.158,2.995)}
						\gppoint{gp mark 1}{(7.192,2.995)}
						\gppoint{gp mark 1}{(7.226,2.865)}
						\gppoint{gp mark 1}{(7.260,2.603)}
						\gppoint{gp mark 1}{(7.294,2.865)}
						\gppoint{gp mark 1}{(7.329,2.472)}
						\gppoint{gp mark 1}{(7.363,2.734)}
						\gppoint{gp mark 1}{(7.397,2.211)}
						\gppoint{gp mark 1}{(7.431,2.211)}
						\gppoint{gp mark 1}{(7.465,2.603)}
						\gppoint{gp mark 1}{(7.500,2.472)}
						\gppoint{gp mark 1}{(7.534,2.342)}
						\gppoint{gp mark 1}{(7.568,2.472)}
						\gppoint{gp mark 1}{(7.602,2.472)}
						\gppoint{gp mark 1}{(7.637,2.211)}
						\gppoint{gp mark 1}{(7.671,2.080)}
						\gppoint{gp mark 1}{(7.705,2.080)}
						\gppoint{gp mark 1}{(7.739,2.211)}
						\gppoint{gp mark 1}{(7.773,2.211)}
						\gppoint{gp mark 1}{(7.808,2.080)}
						\gppoint{gp mark 1}{(7.842,2.080)}
						\gppoint{gp mark 1}{(7.876,2.211)}
						\gppoint{gp mark 1}{(7.910,1.819)}
						\gppoint{gp mark 1}{(7.944,1.688)}
						\gppoint{gp mark 1}{(7.979,2.080)}
						\gppoint{gp mark 1}{(8.013,1.819)}
						\gppoint{gp mark 1}{(8.047,1.819)}
						\gppoint{gp mark 1}{(8.115,1.949)}
						\gppoint{gp mark 1}{(8.150,2.211)}
						\gppoint{gp mark 1}{(8.184,2.211)}
						\gppoint{gp mark 1}{(8.218,2.211)}
						\gppoint{gp mark 1}{(8.252,2.080)}
						\gppoint{gp mark 1}{(8.287,1.819)}
						\gppoint{gp mark 1}{(8.321,2.211)}
						\gppoint{gp mark 1}{(8.355,2.211)}
						\gppoint{gp mark 1}{(8.389,1.819)}
						\gppoint{gp mark 1}{(8.423,2.211)}
						\gppoint{gp mark 1}{(8.458,1.949)}
						\gppoint{gp mark 1}{(8.492,1.949)}
						\gppoint{gp mark 1}{(8.526,2.080)}
						\gppoint{gp mark 1}{(8.560,1.949)}
						\gppoint{gp mark 1}{(8.594,2.211)}
						\gppoint{gp mark 1}{(8.629,2.211)}
						\gppoint{gp mark 1}{(8.663,2.211)}
						\gppoint{gp mark 1}{(8.697,2.342)}
						\gppoint{gp mark 1}{(8.731,2.342)}
						\gppoint{gp mark 1}{(8.765,2.211)}
						\gppoint{gp mark 1}{(8.800,2.211)}
						\gppoint{gp mark 1}{(8.834,2.211)}
						\gppoint{gp mark 1}{(8.868,2.211)}
						\gppoint{gp mark 1}{(8.902,1.949)}
						\gppoint{gp mark 1}{(8.937,2.211)}
						\gppoint{gp mark 1}{(8.971,2.211)}
						\gppoint{gp mark 1}{(9.005,2.342)}
						\gppoint{gp mark 1}{(9.039,2.211)}
						\gppoint{gp mark 1}{(9.073,2.211)}
						\gppoint{gp mark 1}{(9.108,1.949)}
						\gppoint{gp mark 1}{(9.142,2.211)}
						\gppoint{gp mark 1}{(9.176,2.211)}
						\gppoint{gp mark 1}{(9.210,2.211)}
						\gppoint{gp mark 1}{(9.244,2.342)}
						\gppoint{gp mark 1}{(9.279,2.211)}
						\gppoint{gp mark 1}{(9.313,1.819)}
						\gppoint{gp mark 1}{(9.347,2.211)}
						\gppoint{gp mark 1}{(9.381,1.949)}
						\gppoint{gp mark 1}{(9.415,2.342)}
						\gppoint{gp mark 1}{(9.450,2.080)}
						\gppoint{gp mark 1}{(9.484,2.080)}
						\gppoint{gp mark 1}{(9.518,2.211)}
						\gppoint{gp mark 1}{(9.552,2.211)}
						\gppoint{gp mark 1}{(9.587,2.080)}
						\gppoint{gp mark 1}{(9.621,2.080)}
						\gppoint{gp mark 1}{(9.655,2.080)}
						\gppoint{gp mark 1}{(9.689,2.080)}
						\gppoint{gp mark 1}{(9.723,2.211)}
						\gppoint{gp mark 1}{(9.758,2.211)}
						\gppoint{gp mark 1}{(9.792,2.080)}
						\gppoint{gp mark 1}{(9.826,2.472)}
						\gppoint{gp mark 1}{(9.860,1.949)}
						\gppoint{gp mark 1}{(9.894,2.080)}
						\gppoint{gp mark 1}{(9.929,2.342)}
						\gppoint{gp mark 1}{(9.963,1.949)}
						\gppoint{gp mark 1}{(9.997,1.949)}
						\gppoint{gp mark 1}{(10.031,2.080)}
						\gppoint{gp mark 1}{(10.065,2.211)}
						\gppoint{gp mark 1}{(10.100,1.688)}
						\gppoint{gp mark 1}{(10.134,1.949)}
						\gppoint{gp mark 1}{(10.168,2.211)}
						\gppoint{gp mark 1}{(10.202,1.557)}
						\gppoint{gp mark 1}{(10.237,1.819)}
						\gppoint{gp mark 1}{(10.271,1.949)}
						\gppoint{gp mark 1}{(10.305,1.426)}
						\gppoint{gp mark 1}{(10.339,2.080)}
						\gppoint{gp mark 1}{(10.373,1.426)}
						\gppoint{gp mark 1}{(10.408,1.688)}
						\gppoint{gp mark 1}{(10.442,1.949)}
						\gppoint{gp mark 1}{(10.476,1.557)}
						\gppoint{gp mark 1}{(10.510,1.949)}
						\gppoint{gp mark 1}{(10.544,1.819)}
						\gppoint{gp mark 1}{(10.579,1.819)}
						\gppoint{gp mark 1}{(10.613,1.819)}
						\gppoint{gp mark 1}{(10.647,2.080)}
						\gppoint{gp mark 1}{(10.681,2.080)}
						\gppoint{gp mark 1}{(10.715,1.819)}
						\gppoint{gp mark 1}{(10.750,2.080)}
						\gppoint{gp mark 1}{(10.784,1.949)}
						\gppoint{gp mark 1}{(10.818,1.819)}
						\gppoint{gp mark 1}{(10.852,2.211)}
						\gppoint{gp mark 1}{(10.886,1.819)}
						\gppoint{gp mark 1}{(10.921,1.819)}
						\gppoint{gp mark 1}{(10.955,1.949)}
						\gppoint{gp mark 1}{(10.989,1.688)}
						\gppoint{gp mark 1}{(11.023,1.949)}
						\gppoint{gp mark 1}{(11.058,1.819)}
						\gppoint{gp mark 1}{(11.092,2.080)}
						\gppoint{gp mark 1}{(11.126,2.211)}
						\gppoint{gp mark 1}{(11.160,2.472)}
						\gppoint{gp mark 1}{(11.194,2.080)}
						\gppoint{gp mark 1}{(11.229,2.080)}
						\gppoint{gp mark 1}{(11.263,2.342)}
						\gppoint{gp mark 1}{(11.297,1.949)}
						\gppoint{gp mark 1}{(11.331,2.603)}
						\gppoint{gp mark 1}{(11.365,2.342)}
						\gppoint{gp mark 1}{(11.400,2.080)}
						\gppoint{gp mark 1}{(11.434,2.211)}
						\gppoint{gp mark 1}{(11.468,2.211)}
						\gpsetpointsize{2.80}
						\gpcolor{rgb color={0.000,0.000,1.000}}
						\gppoint{gp mark 7}{(1.718,7.441)}
						\gppoint{gp mark 7}{(1.752,5.479)}
						\gppoint{gp mark 7}{(1.787,6.264)}
						\gppoint{gp mark 7}{(1.923,5.087)}
						\gppoint{gp mark 7}{(1.958,3.911)}
						\gppoint{gp mark 7}{(2.060,3.257)}
						\gppoint{gp mark 7}{(2.095,2.211)}
						\gppoint{gp mark 7}{(2.129,2.080)}
						\gppoint{gp mark 7}{(2.197,2.211)}
						\gppoint{gp mark 7}{(2.266,1.949)}
						\gppoint{gp mark 7}{(2.334,2.342)}
						\gppoint{gp mark 7}{(2.471,1.165)}
						\gppoint{gp mark 7}{(2.539,2.080)}
						\gppoint{gp mark 7}{(2.573,2.342)}
						\gppoint{gp mark 7}{(2.608,1.426)}
						\gppoint{gp mark 7}{(2.745,1.949)}
						\gppoint{gp mark 7}{(2.813,1.949)}
						\gppoint{gp mark 7}{(2.847,1.819)}
						\gppoint{gp mark 7}{(2.881,2.472)}
						\gppoint{gp mark 7}{(3.292,2.734)}
						\gppoint{gp mark 7}{(3.463,2.342)}
						\gppoint{gp mark 7}{(3.497,2.342)}
						\gppoint{gp mark 7}{(3.702,2.211)}
						\gppoint{gp mark 7}{(3.737,2.603)}
						\gppoint{gp mark 7}{(3.839,2.342)}
						\gppoint{gp mark 7}{(3.873,2.211)}
						\gppoint{gp mark 7}{(3.908,2.342)}
						\gppoint{gp mark 7}{(3.942,2.472)}
						\gppoint{gp mark 7}{(4.797,3.126)}
						\gppoint{gp mark 7}{(5.310,2.211)}
						\gppoint{gp mark 7}{(5.618,2.734)}
						\gppoint{gp mark 7}{(5.652,2.865)}
						\gppoint{gp mark 7}{(6.337,2.995)}
						\gppoint{gp mark 7}{(6.371,3.257)}
						\gppoint{gp mark 7}{(8.081,2.080)}
						\gpcolor{rgb color={1.000,1.000,0.000}}
						\gpsetpointsize{5.40}
						\gppoint{gp mark 5}{(2.881,2.472)}
						\gpcolor{rgb color={0.000,0.000,0.000}}
						\gpsetpointsize{5.80}
						\gppoint{gp mark 4}{(2.881,2.472)}
						\gpcolor{color=gp lt color border}
						\draw[gp path] (1.684,7.702)--(1.684,1.165)--(11.947,1.165)--(11.947,7.702)--cycle;
						%% coordinates of the plot area
						\gpdefrectangularnode{gp plot 1}{\pgfpoint{1.684cm}{1.165cm}}{\pgfpoint{11.947cm}{7.702cm}}
						\end{tikzpicture}
						\subcaption{Testing set of KP}
						\label{fig:TestF2}
					\end{center}				
				\end{minipage}
			\end{center}
			\caption{An illustration of the performance of the proposed approach for the best subset selection of features on the complete testing set} 
			\label{fig:TestFrontier}
		\end{figure}

		To show this, for each of the points (other than the ideal point)  in Figure~\ref{fig:TrainFrontier}, we have plotted its corresponding point for the testing set in Figure~\ref{fig:TestFrontier}. Specifically, for each point in Figure~\ref{fig:TrainFrontier}, we run its corresponding model on the testing set to compute the error. From Figure~\ref{fig:TestFrontier} we observe that the error highly fluctuates. In fact, it is evident that for AP instances, the prediction model that has around 40 features is the best prediction model. Similarly, from Figure~\ref{fig:TestFrontier}, it is evident that for KP instances, the prediction model that has around 25 features is the best prediction model.
		
		Note that in practice, we do not have access to the testing set. So,  we should select the best subset of features only based on the training set. Therefore, the goal of any best subset selection technique is to identify the prediction model that is (almost) optimal for the testing set based on the existing data set, i.e., the training set. From Figure~\ref{fig:TestFrontier}, we observe that our proposed best subset selection heuristic has such a desirable characteristic in practice. We see that the selected model, i.e., the (yellow) square, is nearly optimal.  In fact the proposed approach has selected a prediction model with the accuracy of around 50\% and 55\% for AP and KP instances, respectively. This implies that the absolute difference between the accuracy of the model selected by the proposed subset selection approach and the accuracy of the optimal model is almost 3\% and 5\% for AP and KP instances, respectively. We note that for both classes of instances less than 50 features exist in the model selected by the proposed approach. Overall, these results are promising due to the fact that the (expected) probability of randomly picking the correct objective function to project is $\frac{1}{p}$, i.e., around 33.3\% for the tri-objective instances. 
		
		\begin{table}[ht]
			\caption{Accuracy and average time decrease of testing set when using the proposed ML technique (for the case of the complete training and testing sets)}
			\centering \footnotesize
			%\resizebox{\textwidth}{!}{ 
			\begin{tabular}{ccclcccccc}
				\cline{7-8}
				\multicolumn{1}{l}{} & \multicolumn{1}{l}{} & \multicolumn{1}{l}{} &  &  & \multicolumn{1}{l|}{} & \multicolumn{2}{c|}{\textbf{Time Decrease}} & \multicolumn{1}{l}{} & \multicolumn{1}{l}{} \\ \cline{1-1} \cline{3-3} \cline{5-5} \cline{7-8} \cline{10-10} 
				\multicolumn{1}{|c|}{\textbf{Type}} & \multicolumn{1}{c|}{} & \multicolumn{1}{c|}{\textbf{Vars}} & \multicolumn{1}{l|}{} & \multicolumn{1}{c|}{\textbf{Accuracy}} & \multicolumn{1}{c|}{} & \multicolumn{1}{c|}{\textbf{ML vs.  Rand}} & \multicolumn{1}{c|}{\textbf{Best vs. Rand}} & \multicolumn{1}{c|}{} & \multicolumn{1}{c|}{\textbf{\makecell[c]{$\bf  \frac{\mbox{ML vs. Rand}}{\mbox{Best vs. Rand}}$}}} \\ \cline{1-1} \cline{3-3} \cline{5-5} \cline{7-8} \cline{10-10} 
				\multicolumn{1}{|c|}{\multirow{5}{*}{AP}} & \multicolumn{1}{c|}{} & \multicolumn{1}{c|}{$20 \times 20$} & \multicolumn{1}{l|}{} & \multicolumn{1}{c|}{55.56\%} & \multicolumn{1}{c|}{} & \multicolumn{1}{c|}{1.29\%} & \multicolumn{1}{c|}{2.33\%} & \multicolumn{1}{c|}{} & \multicolumn{1}{c|}{55.43\%} \\
				\multicolumn{1}{|c|}{} & \multicolumn{1}{c|}{} & \multicolumn{1}{c|}{$25 \times 25$} & \multicolumn{1}{l|}{} & \multicolumn{1}{c|}{44.74\%} & \multicolumn{1}{c|}{} & \multicolumn{1}{c|}{0.67\%} & \multicolumn{1}{c|}{1.65\%} & \multicolumn{1}{c|}{} & \multicolumn{1}{c|}{40.57\%} \\
				\multicolumn{1}{|c|}{} & \multicolumn{1}{c|}{} & \multicolumn{1}{c|}{$30 \times 30$} & \multicolumn{1}{l|}{} & \multicolumn{1}{c|}{41.30\%} & \multicolumn{1}{c|}{} & \multicolumn{1}{c|}{0.18\%} & \multicolumn{1}{c|}{1.48\%} & \multicolumn{1}{c|}{} & \multicolumn{1}{c|}{12.20\%} \\
				\multicolumn{1}{|c|}{} & \multicolumn{1}{c|}{} & \multicolumn{1}{c|}{$35 \times 35$} & \multicolumn{1}{l|}{} & \multicolumn{1}{c|}{52.08\%} & \multicolumn{1}{c|}{} & \multicolumn{1}{c|}{0.44\%} & \multicolumn{1}{c|}{1.69\%} & \multicolumn{1}{c|}{} & \multicolumn{1}{c|}{25.81\%} \\
				\multicolumn{1}{|c|}{} & \multicolumn{1}{c|}{} & \multicolumn{1}{c|}{$40 \times 40$} & \multicolumn{1}{l|}{} & \multicolumn{1}{c|}{56.25\%} & \multicolumn{1}{c|}{} & \multicolumn{1}{c|}{1.07\%} & \multicolumn{1}{c|}{1.65\%} & \multicolumn{1}{c|}{} & \multicolumn{1}{c|}{64.99\%} \\ \cline{1-1} \cline{3-3} \cline{5-5} \cline{7-8} \cline{10-10} 
				\multicolumn{1}{|c|}{\textbf{Avg}} &  &  & \multicolumn{1}{l|}{} & \multicolumn{1}{c|}{\textbf{49.50\%}} & \multicolumn{1}{c|}{} & \multicolumn{1}{c|}{\textbf{0.68\%}} & \multicolumn{1}{c|}{\textbf{1.74\%}} & \multicolumn{1}{c|}{} & \multicolumn{1}{c|}{\textbf{38.90\%}} \\ \cline{1-1} \cline{5-5} \cline{7-8} \cline{10-10} 
				&  &  &  &  &  &  &  &  &  \\ \cline{1-1} \cline{3-3} \cline{5-5} \cline{7-8} \cline{10-10} 
				\multicolumn{1}{|c|}{\multirow{5}{*}{KP}} & \multicolumn{1}{c|}{} & \multicolumn{1}{c|}{60} & \multicolumn{1}{l|}{} & \multicolumn{1}{c|}{63.64\%} & \multicolumn{1}{c|}{} & \multicolumn{1}{c|}{9.12\%} & \multicolumn{1}{c|}{11.03\%} & \multicolumn{1}{c|}{} & \multicolumn{1}{c|}{82.68\%} \\
				\multicolumn{1}{|c|}{} & \multicolumn{1}{c|}{} & \multicolumn{1}{c|}{70} & \multicolumn{1}{l|}{} & \multicolumn{1}{c|}{45.65\%} & \multicolumn{1}{c|}{} & \multicolumn{1}{c|}{2.01\%} & \multicolumn{1}{c|}{10.14\%} & \multicolumn{1}{c|}{} & \multicolumn{1}{c|}{19.79\%} \\
				\multicolumn{1}{|c|}{} & \multicolumn{1}{c|}{} & \multicolumn{1}{c|}{80} & \multicolumn{1}{l|}{} & \multicolumn{1}{c|}{56.10\%} & \multicolumn{1}{c|}{} & \multicolumn{1}{c|}{4.59\%} & \multicolumn{1}{c|}{11.42\%} & \multicolumn{1}{c|}{} & \multicolumn{1}{c|}{40.22\%} \\
				\multicolumn{1}{|c|}{} & \multicolumn{1}{c|}{} & \multicolumn{1}{c|}{90} & \multicolumn{1}{l|}{} & \multicolumn{1}{c|}{58.82\%} & \multicolumn{1}{c|}{} & \multicolumn{1}{c|}{8.05\%} & \multicolumn{1}{c|}{13.27\%} & \multicolumn{1}{c|}{} & \multicolumn{1}{c|}{60.68\%} \\
				\multicolumn{1}{|c|}{} & \multicolumn{1}{c|}{} & \multicolumn{1}{c|}{100} & \multicolumn{1}{l|}{} & \multicolumn{1}{c|}{51.43\%} & \multicolumn{1}{c|}{} & \multicolumn{1}{c|}{5.09\%} & \multicolumn{1}{c|}{10.34\%} & \multicolumn{1}{c|}{} & \multicolumn{1}{c|}{49.24\%} \\ \cline{1-1} \cline{3-3} \cline{5-5} \cline{7-8} \cline{10-10} 
				\multicolumn{1}{|c|}{\textbf{Avg}} &  &  & \multicolumn{1}{l|}{} & \multicolumn{1}{c|}{\textbf{55.00\%}} & \multicolumn{1}{c|}{} & \multicolumn{1}{c|}{\textbf{5.67\%}} & \multicolumn{1}{c|}{\textbf{11.17\%}} & \multicolumn{1}{c|}{} & \multicolumn{1}{c|}{\textbf{50.77\%}} \\ \cline{1-1} \cline{5-5} \cline{7-8} \cline{10-10} 
			\end{tabular}%}
			\label{tab:ExpTable1}
		\end{table}
		
		We now discuss about the performance of the selected model in detail for each class of optimization problems. Table~\ref{tab:ExpTable1} summarizes our findings. In this table, the column labeled `Accuracy' shows the average percentage of the prediction accuracy of the selected model for different subclasses of instances. Note that as mentioned in Introduction, each subclass has 200 instances. The column labeled `ML vs. Rand' shows the average percentage of decrease in solution time when ML technique is used compared to randomly picking an objective function for projecting. The column labeled `Best vs. Rand' shows the average percentage of decrease in solution time when the best objective function is selected for projection compared to randomly picking an objective function for projecting.  Finally,  column labeled `$\frac{\mbox{ML vs. Rand}}{\mbox{Best vs. Rand}}$' shows the percentage of `ML vs. Rand' to  `Best vs. Rand'.
		
		Overall, we observe that our ML method improves the computational time in all testing sets. For AP instances, the improvement is around 0.68\% on average which is small. However, we should note that in the ideal case, we could obtain around 1.74\%  improvement in time on average for such instances. So, the improvement obtained with the proposed ML technique is 38.9\% of the ideal case. For largest subclass of AP instances, this number is around 64.99\%.   For the KP instances, the results are even more promising since the amount of improvement in solution time is around 5.67\% on average. In the ideal case, we could obtain an average improvement of 11.17\% for such instances. So, the improvement obtained with the proposed ML technique is 50.77\% of the ideal case. 
		
		\subsection{Complete training set and reduced testing set}
		\label{subsec:ComplTrainRedTes}
		In this section, we test the performance of the model obtained in Section~\ref{subsec:CompTraingTesting} on a reduced testing set. Specifically, we remove the instances that can be considered as tie cases, i.e., those in which the solution time does not change significantly (relative to other instances) when different objective functions are selected for projection.  To reduce the testing set we apply the following steps:
		\begin{itemize}
			\item \textit{Step 1:} We compute the \textit{time range} of each instance, i.e.,  the difference between the best and worst solution times that can be obtained for the instance when different objective functions are considered for projection.
			\item \textit{Step 2:} For each subclass of instances, i.e., those with the same number of decision variables, we compute the standard deviation and the minimum of time ranges in that subclass.   
			\item \textit{Step 3:} We eliminate an instance, i.e., consider it as a tie case, if its time range is not greater than the sum of the minimum and standard deviation of time ranges in its associated subclass.
		\end{itemize}
		
		As a result of the procedure explained above, the testing set was reduced by 35.5\% for AP instances and by 17.5\% for KP instances. Table~\ref{tab:ExpTable2} summarizes our findings for the reduced testing set.
		
		\begin{table}[ht]
			\caption{Accuracy and average time decrease of testing set when using the proposed ML technique (for the case of the complete training set and the reduced testing set)}
			\centering \footnotesize
			%\resizebox{\textwidth}{!}{ 
			\begin{tabular}{ccclcccccc}
				\cline{7-8}
				\multicolumn{1}{l}{} & \multicolumn{1}{l}{} & \multicolumn{1}{l}{} &  &  & \multicolumn{1}{l|}{} & \multicolumn{2}{c|}{\textbf{Time Decrease}} & \multicolumn{1}{l}{} & \multicolumn{1}{l}{} \\ \cline{1-1} \cline{3-3} \cline{5-5} \cline{7-8} \cline{10-10} 
				\multicolumn{1}{|c|}{\textbf{Type}} & \multicolumn{1}{c|}{} & \multicolumn{1}{c|}{\textbf{Vars}} & \multicolumn{1}{l|}{} & \multicolumn{1}{c|}{\textbf{Accuracy}} & \multicolumn{1}{c|}{} & \multicolumn{1}{c|}{\textbf{ML vs.  Rand}} & \multicolumn{1}{c|}{\textbf{Best vs. Rand}} & \multicolumn{1}{c|}{} & \multicolumn{1}{c|}{\textbf{\makecell[c]{$\bf  \frac{\mbox{ML vs. Rand}}{\mbox{Best vs. Rand}}$}}} \\ \cline{1-1} \cline{3-3} \cline{5-5} \cline{7-8}
				\cline{10-10} 
				\multicolumn{1}{|c|}{\multirow{5}{*}{AP}} & \multicolumn{1}{c|}{} & \multicolumn{1}{c|}{$20 \times 20$} & \multicolumn{1}{l|}{} & \multicolumn{1}{c|}{58.06\%} & \multicolumn{1}{c|}{} & \multicolumn{1}{c|}{1.48\%} & \multicolumn{1}{c|}{2.54\%} & \multicolumn{1}{c|}{} & \multicolumn{1}{c|}{58.10\%} \\
				\multicolumn{1}{|c|}{} & \multicolumn{1}{c|}{} & \multicolumn{1}{c|}{$25 \times 25$} & \multicolumn{1}{l|}{} & \multicolumn{1}{c|}{60.00\%} & \multicolumn{1}{c|}{} & \multicolumn{1}{c|}{0.92\%} & \multicolumn{1}{c|}{2.23\%} & \multicolumn{1}{c|}{} & \multicolumn{1}{c|}{41.32\%} \\
				\multicolumn{1}{|c|}{} & \multicolumn{1}{c|}{} & \multicolumn{1}{c|}{$30 \times 30$} & \multicolumn{1}{l|}{} & \multicolumn{1}{c|}{40.74\%} & \multicolumn{1}{c|}{} & \multicolumn{1}{c|}{0.12\%} & \multicolumn{1}{c|}{1.96\%} & \multicolumn{1}{c|}{} & \multicolumn{1}{c|}{6.02\%} \\
				\multicolumn{1}{|c|}{} & \multicolumn{1}{c|}{} & \multicolumn{1}{c|}{$35 \times 35$} & \multicolumn{1}{l|}{} & \multicolumn{1}{c|}{57.89\%} & \multicolumn{1}{c|}{} & \multicolumn{1}{c|}{0.54\%} & \multicolumn{1}{c|}{1.97\%} & \multicolumn{1}{c|}{} & \multicolumn{1}{c|}{27.17\%} \\
				\multicolumn{1}{|c|}{} & \multicolumn{1}{c|}{} & \multicolumn{1}{c|}{$40 \times 40$} & \multicolumn{1}{l|}{} & \multicolumn{1}{c|}{69.57\%} & \multicolumn{1}{c|}{} & \multicolumn{1}{c|}{1.64\%} & \multicolumn{1}{c|}{2.15\%} & \multicolumn{1}{c|}{} & \multicolumn{1}{c|}{76.09\%} \\ \cline{1-1} \cline{3-3} \cline{5-5} \cline{7-8} \cline{10-10} 
				\multicolumn{1}{|c|}{\textbf{Avg}} &  &  & \multicolumn{1}{l|}{} & \multicolumn{1}{c|}{\textbf{56.59\%}} & \multicolumn{1}{c|}{} & \multicolumn{1}{c|}{\textbf{0.90\%}} & \multicolumn{1}{c|}{\textbf{2.16\%}} & \multicolumn{1}{c|}{} & \multicolumn{1}{c|}{\textbf{41.73\%}} \\ \cline{1-1} \cline{5-5} \cline{7-8} \cline{10-10} 
				&  &  &  &  &  &  &  &  &  \\ \cline{1-1} \cline{3-3} \cline{5-5} \cline{7-8} \cline{10-10} 
				\multicolumn{1}{|c|}{\multirow{5}{*}{KP}} & \multicolumn{1}{c|}{} & \multicolumn{1}{c|}{60} & \multicolumn{1}{l|}{} & \multicolumn{1}{c|}{71.05\%} & \multicolumn{1}{c|}{} & \multicolumn{1}{c|}{10.11\%} & \multicolumn{1}{c|}{11.96\%} & \multicolumn{1}{c|}{} & \multicolumn{1}{c|}{84.52\%} \\
				\multicolumn{1}{|c|}{} & \multicolumn{1}{c|}{} & \multicolumn{1}{c|}{70} & \multicolumn{1}{l|}{} & \multicolumn{1}{c|}{45.95\%} & \multicolumn{1}{c|}{} & \multicolumn{1}{c|}{2.03\%} & \multicolumn{1}{c|}{11.48\%} & \multicolumn{1}{c|}{} & \multicolumn{1}{c|}{17.67\%} \\
				\multicolumn{1}{|c|}{} & \multicolumn{1}{c|}{} & \multicolumn{1}{c|}{80} & \multicolumn{1}{l|}{} & \multicolumn{1}{c|}{57.58\%} & \multicolumn{1}{c|}{} & \multicolumn{1}{c|}{5.30\%} & \multicolumn{1}{c|}{13.48\%} & \multicolumn{1}{c|}{} & \multicolumn{1}{c|}{39.31\%} \\
				\multicolumn{1}{|c|}{} & \multicolumn{1}{c|}{} & \multicolumn{1}{c|}{90} & \multicolumn{1}{l|}{} & \multicolumn{1}{c|}{62.96\%} & \multicolumn{1}{c|}{} & \multicolumn{1}{c|}{10.21\%} & \multicolumn{1}{c|}{15.57\%} & \multicolumn{1}{c|}{} & \multicolumn{1}{c|}{65.59\%} \\
				\multicolumn{1}{|c|}{} & \multicolumn{1}{c|}{} & \multicolumn{1}{c|}{100} & \multicolumn{1}{l|}{} & \multicolumn{1}{c|}{60.00\%} & \multicolumn{1}{c|}{} & \multicolumn{1}{c|}{6.32\%} & \multicolumn{1}{c|}{11.33\%} & \multicolumn{1}{c|}{} & \multicolumn{1}{c|}{55.77\%} \\ \cline{1-1} \cline{3-3} \cline{5-5} \cline{7-8} \cline{10-10} 
				\multicolumn{1}{|c|}{\textbf{Avg}} &  &  & \multicolumn{1}{l|}{} & \multicolumn{1}{c|}{\textbf{59.39\%}} & \multicolumn{1}{c|}{} & \multicolumn{1}{c|}{\textbf{6.66\%}} & \multicolumn{1}{c|}{\textbf{12.63\%}} & \multicolumn{1}{c|}{} & \multicolumn{1}{c|}{\textbf{52.74\%}} \\ \cline{1-1} \cline{5-5} \cline{7-8} \cline{10-10} 
			\end{tabular}%}
			\label{tab:ExpTable2}
		\end{table}
		Observe that the accuracy of the prediction models has increased significantly for the reduced testing set. Specifically, it has reached to around 56.59\% and 59.39\%  on overage for AP and KP instances, respectively. Since the eliminated instances are considered as tie cases, we can assume that they are also success cases for the prediction model. So, by considering such success cases, the prediction accuracy will increase to $56.59\times (1-0.355)+35.5 \approxeq 72\%$ and $59.39\times(1-0.175)+17.5 \approxeq 66.5\%$ for AP and KP instances, respectively. In terms of computational time, we also observe (from Table~\ref{tab:ExpTable2}) an improvement of around 0.90\% and 6.66\% on average for AP and KP instances, respectively. This amount of improvement is  about $41.73\%$ and $52.74\%$ of the ideal scenarios (on average) for AP and KP instances, respectively.
		
		\subsection{Reduced training and testing sets}
		\label{subsec:RedTrainTes}
		Due to promising results obtained in Section~\ref{subsec:ComplTrainRedTes}, it is natural to ask whether we can see even more improvement if we reduce not only the testing set but also the training set. Therefore, in this section, we eliminate the tie cases using the same procedure discussed in Section~\ref{subsec:ComplTrainRedTes} from both training and testing sets. By doing so, the size of the training+testing set was reduced by 37\% and 18\% for AP and KP instances, respectively. 
		
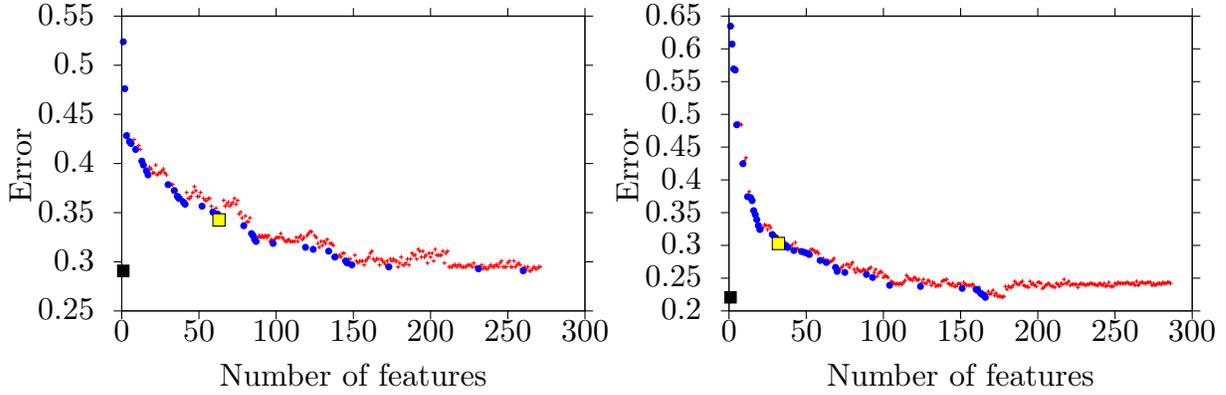
\begin{figure}[ht]
			\begin{center}
				\begin{minipage}{0.45\linewidth}
					\begin{center}
						\begin{tikzpicture}[scale=0.60,
						axis/.style={ ->, >=stealth'},line/.style={very thick},]
						\path (0.000,0.000) rectangle (12.500,8.750);
						\gpcolor{color=gp lt color border}
						\gpsetlinetype{gp lt border}
						\gpsetdashtype{gp dt solid}
						\gpsetlinewidth{1.00}
						\draw[gp path] (1.684,1.165)--(1.504,1.165);
						\draw[gp path] (11.947,1.165)--(12.127,1.165);
						\node[gp node right] at (1.320,1.165) {$0.25$};
						\draw[gp path] (1.684,2.254)--(1.504,2.254);
						\draw[gp path] (11.947,2.254)--(12.127,2.254);
						\node[gp node right] at (1.320,2.254) {$0.3$};
						\draw[gp path] (1.684,3.344)--(1.504,3.344);
						\draw[gp path] (11.947,3.344)--(12.127,3.344);
						\node[gp node right] at (1.320,3.344) {$0.35$};
						\draw[gp path] (1.684,4.433)--(1.504,4.433);
						\draw[gp path] (11.947,4.433)--(12.127,4.433);
						\node[gp node right] at (1.320,4.433) {$0.4$};
						\draw[gp path] (1.684,5.523)--(1.504,5.523);
						\draw[gp path] (11.947,5.523)--(12.127,5.523);
						\node[gp node right] at (1.320,5.523) {$0.45$};
						\draw[gp path] (1.684,6.612)--(1.504,6.612);
						\draw[gp path] (11.947,6.612)--(12.127,6.612);
						\node[gp node right] at (1.320,6.612) {$0.5$};
						\draw[gp path] (1.684,7.702)--(1.504,7.702);
						\draw[gp path] (11.947,7.702)--(12.127,7.702);
						\node[gp node right] at (1.320,7.702) {$0.55$};
						\draw[gp path] (1.684,1.165)--(1.684,0.985);
						\draw[gp path] (1.684,7.702)--(1.684,7.882);
						\node[gp node center] at (1.684,0.677) {$0$};
						\draw[gp path] (3.395,1.165)--(3.395,0.985);
						\draw[gp path] (3.395,7.702)--(3.395,7.882);
						\node[gp node center] at (3.395,0.677) {$50$};
						\draw[gp path] (5.105,1.165)--(5.105,0.985);
						\draw[gp path] (5.105,7.702)--(5.105,7.882);
						\node[gp node center] at (5.105,0.677) {$100$};
						\draw[gp path] (6.816,1.165)--(6.816,0.985);
						\draw[gp path] (6.816,7.702)--(6.816,7.882);
						\node[gp node center] at (6.816,0.677) {$150$};
						\draw[gp path] (8.526,1.165)--(8.526,0.985);
						\draw[gp path] (8.526,7.702)--(8.526,7.882);
						\node[gp node center] at (8.526,0.677) {$200$};
						\draw[gp path] (10.237,1.165)--(10.237,0.985);
						\draw[gp path] (10.237,7.702)--(10.237,7.882);
						\node[gp node center] at (10.237,0.677) {$250$};
						\draw[gp path] (11.947,1.165)--(11.947,0.985);
						\draw[gp path] (11.947,7.702)--(11.947,7.882);
						\node[gp node center] at (11.947,0.677) {$300$};
						\draw[gp path] (1.684,7.702)--(1.684,1.165)--(11.947,1.165)--(11.947,7.702)--cycle;
						\node[gp node center,rotate=-270] at (-0.476,4.433) {Error};
						\node[gp node center] at (6.815,-0.315) {Number of features};
						%					\node[gp node center,scale=0.8,font={\fontsize{14.0pt}{16.8pt}\selectfont}] at (6.815,8.287) {\textbf{Training set error AP instances}};
						\gpcolor{rgb color={1.000,0.000,0.000}}
						\gpsetpointsize{2.00}
						\gppoint{gp mark 1}{(1.821,5.094)}
						\gppoint{gp mark 1}{(1.923,4.876)}
						\gppoint{gp mark 1}{(1.958,4.963)}
						\gppoint{gp mark 1}{(2.026,4.745)}
						\gppoint{gp mark 1}{(2.060,4.832)}
						\gppoint{gp mark 1}{(2.095,4.745)}
						\gppoint{gp mark 1}{(2.197,4.399)}
						\gppoint{gp mark 1}{(2.300,4.355)}
						\gppoint{gp mark 1}{(2.334,4.311)}
						\gppoint{gp mark 1}{(2.368,4.224)}
						\gppoint{gp mark 1}{(2.402,4.224)}
						\gppoint{gp mark 1}{(2.437,4.399)}
						\gppoint{gp mark 1}{(2.471,4.181)}
						\gppoint{gp mark 1}{(2.505,4.224)}
						\gppoint{gp mark 1}{(2.539,4.224)}
						\gppoint{gp mark 1}{(2.573,4.268)}
						\gppoint{gp mark 1}{(2.608,4.311)}
						\gppoint{gp mark 1}{(2.642,4.224)}
						\gppoint{gp mark 1}{(2.676,4.181)}
						\gppoint{gp mark 1}{(2.745,3.965)}
						\gppoint{gp mark 1}{(2.779,3.965)}
						\gppoint{gp mark 1}{(2.813,3.965)}
						\gppoint{gp mark 1}{(2.881,3.834)}
						\gppoint{gp mark 1}{(2.984,3.704)}
						\gppoint{gp mark 1}{(3.018,3.660)}
						\gppoint{gp mark 1}{(3.121,3.791)}
						\gppoint{gp mark 1}{(3.155,3.660)}
						\gppoint{gp mark 1}{(3.189,3.660)}
						\gppoint{gp mark 1}{(3.223,3.704)}
						\gppoint{gp mark 1}{(3.258,3.791)}
						\gppoint{gp mark 1}{(3.292,3.921)}
						\gppoint{gp mark 1}{(3.326,3.834)}
						\gppoint{gp mark 1}{(3.360,3.704)}
						\gppoint{gp mark 1}{(3.395,3.704)}
						\gppoint{gp mark 1}{(3.429,3.791)}
						\gppoint{gp mark 1}{(3.497,3.660)}
						\gppoint{gp mark 1}{(3.531,3.616)}
						\gppoint{gp mark 1}{(3.566,3.616)}
						\gppoint{gp mark 1}{(3.600,3.660)}
						\gppoint{gp mark 1}{(3.634,3.660)}
						\gppoint{gp mark 1}{(3.668,3.488)}
						\gppoint{gp mark 1}{(3.737,3.357)}
						\gppoint{gp mark 1}{(3.771,3.444)}
						\gppoint{gp mark 1}{(3.873,3.183)}
						\gppoint{gp mark 1}{(3.908,3.270)}
						\gppoint{gp mark 1}{(3.942,3.616)}
						\gppoint{gp mark 1}{(3.976,3.488)}
						\gppoint{gp mark 1}{(4.010,3.531)}
						\gppoint{gp mark 1}{(4.044,3.575)}
						\gppoint{gp mark 1}{(4.079,3.488)}
						\gppoint{gp mark 1}{(4.113,3.531)}
						\gppoint{gp mark 1}{(4.147,3.575)}
						\gppoint{gp mark 1}{(4.181,3.660)}
						\gppoint{gp mark 1}{(4.216,3.531)}
						\gppoint{gp mark 1}{(4.250,3.616)}
						\gppoint{gp mark 1}{(4.284,3.313)}
						\gppoint{gp mark 1}{(4.318,3.357)}
						\gppoint{gp mark 1}{(4.352,3.226)}
						\gppoint{gp mark 1}{(4.421,3.270)}
						\gppoint{gp mark 1}{(4.455,3.139)}
						\gppoint{gp mark 1}{(4.489,3.139)}
						\gppoint{gp mark 1}{(4.523,3.226)}
						\gppoint{gp mark 1}{(4.694,2.793)}
						\gppoint{gp mark 1}{(4.729,2.836)}
						\gppoint{gp mark 1}{(4.763,2.793)}
						\gppoint{gp mark 1}{(4.797,2.836)}
						\gppoint{gp mark 1}{(4.831,2.793)}
						\gppoint{gp mark 1}{(4.866,2.836)}
						\gppoint{gp mark 1}{(4.900,2.793)}
						\gppoint{gp mark 1}{(4.934,2.836)}
						\gppoint{gp mark 1}{(4.968,2.706)}
						\gppoint{gp mark 1}{(5.002,2.749)}
						\gppoint{gp mark 1}{(5.071,2.749)}
						\gppoint{gp mark 1}{(5.105,2.793)}
						\gppoint{gp mark 1}{(5.139,2.793)}
						\gppoint{gp mark 1}{(5.173,2.793)}
						\gppoint{gp mark 1}{(5.208,2.880)}
						\gppoint{gp mark 1}{(5.242,2.793)}
						\gppoint{gp mark 1}{(5.276,2.749)}
						\gppoint{gp mark 1}{(5.310,2.706)}
						\gppoint{gp mark 1}{(5.344,2.706)}
						\gppoint{gp mark 1}{(5.379,2.749)}
						\gppoint{gp mark 1}{(5.413,2.706)}
						\gppoint{gp mark 1}{(5.447,2.706)}
						\gppoint{gp mark 1}{(5.481,2.706)}
						\gppoint{gp mark 1}{(5.516,2.706)}
						\gppoint{gp mark 1}{(5.550,2.793)}
						\gppoint{gp mark 1}{(5.584,2.706)}
						\gppoint{gp mark 1}{(5.618,2.793)}
						\gppoint{gp mark 1}{(5.652,2.836)}
						\gppoint{gp mark 1}{(5.687,2.749)}
						\gppoint{gp mark 1}{(5.721,2.793)}
						\gppoint{gp mark 1}{(5.789,2.836)}
						\gppoint{gp mark 1}{(5.823,2.880)}
						\gppoint{gp mark 1}{(5.858,2.923)}
						\gppoint{gp mark 1}{(5.892,2.923)}
						\gppoint{gp mark 1}{(5.960,2.836)}
						\gppoint{gp mark 1}{(5.994,2.793)}
						\gppoint{gp mark 1}{(6.029,2.662)}
						\gppoint{gp mark 1}{(6.063,2.575)}
						\gppoint{gp mark 1}{(6.097,2.706)}
						\gppoint{gp mark 1}{(6.131,2.618)}
						\gppoint{gp mark 1}{(6.166,2.706)}
						\gppoint{gp mark 1}{(6.200,2.662)}
						\gppoint{gp mark 1}{(6.234,2.749)}
						\gppoint{gp mark 1}{(6.302,2.662)}
						\gppoint{gp mark 1}{(6.337,2.662)}
						\gppoint{gp mark 1}{(6.371,2.618)}
						\gppoint{gp mark 1}{(6.439,2.403)}
						\gppoint{gp mark 1}{(6.473,2.359)}
						\gppoint{gp mark 1}{(6.508,2.403)}
						\gppoint{gp mark 1}{(6.542,2.359)}
						\gppoint{gp mark 1}{(6.576,2.446)}
						\gppoint{gp mark 1}{(6.610,2.446)}
						\gppoint{gp mark 1}{(6.713,2.316)}
						\gppoint{gp mark 1}{(6.747,2.316)}
						\gppoint{gp mark 1}{(6.816,2.185)}
						\gppoint{gp mark 1}{(6.850,2.403)}
						\gppoint{gp mark 1}{(6.884,2.272)}
						\gppoint{gp mark 1}{(6.918,2.228)}
						\gppoint{gp mark 1}{(6.952,2.272)}
						\gppoint{gp mark 1}{(6.987,2.272)}
						\gppoint{gp mark 1}{(7.021,2.272)}
						\gppoint{gp mark 1}{(7.055,2.185)}
						\gppoint{gp mark 1}{(7.089,2.272)}
						\gppoint{gp mark 1}{(7.123,2.316)}
						\gppoint{gp mark 1}{(7.158,2.359)}
						\gppoint{gp mark 1}{(7.192,2.446)}
						\gppoint{gp mark 1}{(7.226,2.490)}
						\gppoint{gp mark 1}{(7.260,2.316)}
						\gppoint{gp mark 1}{(7.294,2.272)}
						\gppoint{gp mark 1}{(7.329,2.316)}
						\gppoint{gp mark 1}{(7.363,2.316)}
						\gppoint{gp mark 1}{(7.397,2.185)}
						\gppoint{gp mark 1}{(7.431,2.316)}
						\gppoint{gp mark 1}{(7.465,2.272)}
						\gppoint{gp mark 1}{(7.500,2.228)}
						\gppoint{gp mark 1}{(7.534,2.185)}
						\gppoint{gp mark 1}{(7.568,2.272)}
						\gppoint{gp mark 1}{(7.637,2.316)}
						\gppoint{gp mark 1}{(7.671,2.272)}
						\gppoint{gp mark 1}{(7.705,2.316)}
						\gppoint{gp mark 1}{(7.739,2.272)}
						\gppoint{gp mark 1}{(7.773,2.272)}
						\gppoint{gp mark 1}{(7.808,2.141)}
						\gppoint{gp mark 1}{(7.842,2.228)}
						\gppoint{gp mark 1}{(7.876,2.359)}
						\gppoint{gp mark 1}{(7.910,2.185)}
						\gppoint{gp mark 1}{(7.944,2.185)}
						\gppoint{gp mark 1}{(7.979,2.228)}
						\gppoint{gp mark 1}{(8.013,2.316)}
						\gppoint{gp mark 1}{(8.047,2.185)}
						\gppoint{gp mark 1}{(8.081,2.316)}
						\gppoint{gp mark 1}{(8.115,2.446)}
						\gppoint{gp mark 1}{(8.150,2.403)}
						\gppoint{gp mark 1}{(8.184,2.403)}
						\gppoint{gp mark 1}{(8.218,2.316)}
						\gppoint{gp mark 1}{(8.252,2.531)}
						\gppoint{gp mark 1}{(8.287,2.490)}
						\gppoint{gp mark 1}{(8.321,2.359)}
						\gppoint{gp mark 1}{(8.355,2.490)}
						\gppoint{gp mark 1}{(8.389,2.359)}
						\gppoint{gp mark 1}{(8.423,2.403)}
						\gppoint{gp mark 1}{(8.458,2.228)}
						\gppoint{gp mark 1}{(8.492,2.446)}
						\gppoint{gp mark 1}{(8.526,2.359)}
						\gppoint{gp mark 1}{(8.560,2.446)}
						\gppoint{gp mark 1}{(8.594,2.272)}
						\gppoint{gp mark 1}{(8.629,2.531)}
						\gppoint{gp mark 1}{(8.663,2.403)}
						\gppoint{gp mark 1}{(8.697,2.403)}
						\gppoint{gp mark 1}{(8.731,2.403)}
						\gppoint{gp mark 1}{(8.765,2.403)}
						\gppoint{gp mark 1}{(8.800,2.490)}
						\gppoint{gp mark 1}{(8.834,2.228)}
						\gppoint{gp mark 1}{(8.868,2.446)}
						\gppoint{gp mark 1}{(8.902,2.490)}
						\gppoint{gp mark 1}{(8.937,2.141)}
						\gppoint{gp mark 1}{(8.971,2.141)}
						\gppoint{gp mark 1}{(9.005,2.141)}
						\gppoint{gp mark 1}{(9.039,2.185)}
						\gppoint{gp mark 1}{(9.073,2.185)}
						\gppoint{gp mark 1}{(9.108,2.141)}
						\gppoint{gp mark 1}{(9.142,2.185)}
						\gppoint{gp mark 1}{(9.176,2.185)}
						\gppoint{gp mark 1}{(9.210,2.141)}
						\gppoint{gp mark 1}{(9.244,2.141)}
						\gppoint{gp mark 1}{(9.279,2.141)}
						\gppoint{gp mark 1}{(9.313,2.141)}
						\gppoint{gp mark 1}{(9.347,2.185)}
						\gppoint{gp mark 1}{(9.381,2.185)}
						\gppoint{gp mark 1}{(9.415,2.185)}
						\gppoint{gp mark 1}{(9.450,2.141)}
						\gppoint{gp mark 1}{(9.484,2.228)}
						\gppoint{gp mark 1}{(9.518,2.141)}
						\gppoint{gp mark 1}{(9.552,2.141)}
						\gppoint{gp mark 1}{(9.621,2.141)}
						\gppoint{gp mark 1}{(9.655,2.185)}
						\gppoint{gp mark 1}{(9.689,2.098)}
						\gppoint{gp mark 1}{(9.723,2.141)}
						\gppoint{gp mark 1}{(9.758,2.098)}
						\gppoint{gp mark 1}{(9.792,2.098)}
						\gppoint{gp mark 1}{(9.826,2.141)}
						\gppoint{gp mark 1}{(9.860,2.185)}
						\gppoint{gp mark 1}{(9.894,2.141)}
						\gppoint{gp mark 1}{(9.929,2.141)}
						\gppoint{gp mark 1}{(9.963,2.185)}
						\gppoint{gp mark 1}{(9.997,2.185)}
						\gppoint{gp mark 1}{(10.031,2.141)}
						\gppoint{gp mark 1}{(10.065,2.228)}
						\gppoint{gp mark 1}{(10.100,2.272)}
						\gppoint{gp mark 1}{(10.134,2.141)}
						\gppoint{gp mark 1}{(10.168,2.141)}
						\gppoint{gp mark 1}{(10.202,2.141)}
						\gppoint{gp mark 1}{(10.237,2.228)}
						\gppoint{gp mark 1}{(10.271,2.272)}
						\gppoint{gp mark 1}{(10.305,2.141)}
						\gppoint{gp mark 1}{(10.339,2.141)}
						\gppoint{gp mark 1}{(10.373,2.272)}
						\gppoint{gp mark 1}{(10.408,2.141)}
						\gppoint{gp mark 1}{(10.442,2.185)}
						\gppoint{gp mark 1}{(10.476,2.228)}
						\gppoint{gp mark 1}{(10.510,2.185)}
						\gppoint{gp mark 1}{(10.544,2.098)}
						\gppoint{gp mark 1}{(10.613,2.141)}
						\gppoint{gp mark 1}{(10.647,2.098)}
						\gppoint{gp mark 1}{(10.681,2.141)}
						\gppoint{gp mark 1}{(10.715,2.054)}
						\gppoint{gp mark 1}{(10.750,2.054)}
						\gppoint{gp mark 1}{(10.784,2.098)}
						\gppoint{gp mark 1}{(10.818,2.141)}
						\gppoint{gp mark 1}{(10.852,2.098)}
						\gppoint{gp mark 1}{(10.886,2.098)}
						\gppoint{gp mark 1}{(10.921,2.141)}
						\gppoint{gp mark 1}{(10.955,2.141)}
						\gpsetpointsize{2.80}
						\gpcolor{rgb color={0.000,0.000,1.000}}
						\gppoint{gp mark 7}{(1.718,7.133)}
						\gppoint{gp mark 7}{(1.752,6.092)}
						\gppoint{gp mark 7}{(1.787,5.050)}
						\gppoint{gp mark 7}{(1.855,4.919)}
						\gppoint{gp mark 7}{(1.889,4.876)}
						\gppoint{gp mark 7}{(1.992,4.745)}
						\gppoint{gp mark 7}{(2.129,4.486)}
						\gppoint{gp mark 7}{(2.163,4.399)}
						\gppoint{gp mark 7}{(2.231,4.268)}
						\gppoint{gp mark 7}{(2.266,4.181)}
						\gppoint{gp mark 7}{(2.710,3.965)}
						\gppoint{gp mark 7}{(2.847,3.834)}
						\gppoint{gp mark 7}{(2.916,3.704)}
						\gppoint{gp mark 7}{(2.950,3.660)}
						\gppoint{gp mark 7}{(3.052,3.575)}
						\gppoint{gp mark 7}{(3.087,3.531)}
						\gppoint{gp mark 7}{(3.463,3.488)}
						\gppoint{gp mark 7}{(3.702,3.357)}
						\gppoint{gp mark 7}{(3.805,3.313)}
						\gppoint{gp mark 7}{(3.839,3.183)}
						\gppoint{gp mark 7}{(4.387,3.054)}
						\gppoint{gp mark 7}{(4.558,2.880)}
						\gppoint{gp mark 7}{(4.592,2.836)}
						\gppoint{gp mark 7}{(4.626,2.749)}
						\gppoint{gp mark 7}{(4.660,2.706)}
						\gppoint{gp mark 7}{(5.037,2.662)}
						\gppoint{gp mark 7}{(5.755,2.575)}
						\gppoint{gp mark 7}{(5.926,2.531)}
						\gppoint{gp mark 7}{(6.268,2.490)}
						\gppoint{gp mark 7}{(6.405,2.359)}
						\gppoint{gp mark 7}{(6.644,2.272)}
						\gppoint{gp mark 7}{(6.679,2.228)}
						\gppoint{gp mark 7}{(6.781,2.185)}
						\gppoint{gp mark 7}{(7.602,2.141)}
						\gppoint{gp mark 7}{(9.587,2.098)}
						\gppoint{gp mark 7}{(10.579,2.054)}
						\gpcolor{rgb color={1.000,1.000,0.000}}
						\gpsetpointsize{5.40}
						\gppoint{gp mark 5}{(3.839,3.183)}
						\gpcolor{rgb color={0.000,0.000,0.000}}
						\gpsetpointsize{5.80}
						\gppoint{gp mark 4}{(3.839,3.183)}
						\gpsetpointsize{5.40}
						\gppoint{gp mark 5}{(1.718,2.054)}
						\gpcolor{color=gp lt color border}
						\draw[gp path] (1.684,7.702)--(1.684,1.165)--(11.947,1.165)--(11.947,7.702)--cycle;
						%% coordinates of the plot area
						\gpdefrectangularnode{gp plot 1}{\pgfpoint{1.684cm}{1.165cm}}{\pgfpoint{11.947cm}{7.702cm}}
						\end{tikzpicture}
						\subcaption{Training set of AP}
						\label{fig:TF12}
					\end{center}				
				\end{minipage}
				\hfil
				\begin{minipage}{0.45\linewidth}
					\begin{center}
						\begin{tikzpicture}[scale=0.60,
						axis/.style={ ->, >=stealth'},line/.style={very thick},]
						\path (0.000,0.000) rectangle (12.500,8.750);
						\gpcolor{color=gp lt color border}
						\gpsetlinetype{gp lt border}
						\gpsetdashtype{gp dt solid}
						\gpsetlinewidth{1.00}
						\draw[gp path] (1.684,1.165)--(1.504,1.165);
						\draw[gp path] (11.947,1.165)--(12.127,1.165);
						\node[gp node right] at (1.320,1.165) {$0.2$};
						\draw[gp path] (1.684,1.891)--(1.504,1.891);
						\draw[gp path] (11.947,1.891)--(12.127,1.891);
						\node[gp node right] at (1.320,1.891) {$0.25$};
						\draw[gp path] (1.684,2.618)--(1.504,2.618);
						\draw[gp path] (11.947,2.618)--(12.127,2.618);
						\node[gp node right] at (1.320,2.618) {$0.3$};
						\draw[gp path] (1.684,3.344)--(1.504,3.344);
						\draw[gp path] (11.947,3.344)--(12.127,3.344);
						\node[gp node right] at (1.320,3.344) {$0.35$};
						\draw[gp path] (1.684,4.070)--(1.504,4.070);
						\draw[gp path] (11.947,4.070)--(12.127,4.070);
						\node[gp node right] at (1.320,4.070) {$0.4$};
						\draw[gp path] (1.684,4.797)--(1.504,4.797);
						\draw[gp path] (11.947,4.797)--(12.127,4.797);
						\node[gp node right] at (1.320,4.797) {$0.45$};
						\draw[gp path] (1.684,5.523)--(1.504,5.523);
						\draw[gp path] (11.947,5.523)--(12.127,5.523);
						\node[gp node right] at (1.320,5.523) {$0.5$};
						\draw[gp path] (1.684,6.249)--(1.504,6.249);
						\draw[gp path] (11.947,6.249)--(12.127,6.249);
						\node[gp node right] at (1.320,6.249) {$0.55$};
						\draw[gp path] (1.684,6.976)--(1.504,6.976);
						\draw[gp path] (11.947,6.976)--(12.127,6.976);
						\node[gp node right] at (1.320,6.976) {$0.6$};
						\draw[gp path] (1.684,7.702)--(1.504,7.702);
						\draw[gp path] (11.947,7.702)--(12.127,7.702);
						\node[gp node right] at (1.320,7.702) {$0.65$};
						\draw[gp path] (1.684,1.165)--(1.684,0.985);
						\draw[gp path] (1.684,7.702)--(1.684,7.882);
						\node[gp node center] at (1.684,0.677) {$0$};
						\draw[gp path] (3.395,1.165)--(3.395,0.985);
						\draw[gp path] (3.395,7.702)--(3.395,7.882);
						\node[gp node center] at (3.395,0.677) {$50$};
						\draw[gp path] (5.105,1.165)--(5.105,0.985);
						\draw[gp path] (5.105,7.702)--(5.105,7.882);
						\node[gp node center] at (5.105,0.677) {$100$};
						\draw[gp path] (6.816,1.165)--(6.816,0.985);
						\draw[gp path] (6.816,7.702)--(6.816,7.882);
						\node[gp node center] at (6.816,0.677) {$150$};
						\draw[gp path] (8.526,1.165)--(8.526,0.985);
						\draw[gp path] (8.526,7.702)--(8.526,7.882);
						\node[gp node center] at (8.526,0.677) {$200$};
						\draw[gp path] (10.237,1.165)--(10.237,0.985);
						\draw[gp path] (10.237,7.702)--(10.237,7.882);
						\node[gp node center] at (10.237,0.677) {$250$};
						\draw[gp path] (11.947,1.165)--(11.947,0.985);
						\draw[gp path] (11.947,7.702)--(11.947,7.882);
						\node[gp node center] at (11.947,0.677) {$300$};
						\draw[gp path] (1.684,7.702)--(1.684,1.165)--(11.947,1.165)--(11.947,7.702)--cycle;
						\node[gp node center,rotate=-270] at (-0.476,4.433) {Error};
						\node[gp node center] at (6.815,-0.315) {Number of features};
						%					\node[gp node center,scale=0.8,font={\fontsize{14.0pt}{16.8pt}\selectfont}] at (6.815,8.287) {\textbf{Training set error KP instances}};
						\gpcolor{rgb color={1.000,0.000,0.000}}
						\gpsetpointsize{2.00}
						\gppoint{gp mark 1}{(1.889,5.336)}
						\gppoint{gp mark 1}{(1.923,5.312)}
						\gppoint{gp mark 1}{(1.958,5.291)}
						\gppoint{gp mark 1}{(2.026,4.495)}
						\gppoint{gp mark 1}{(2.060,4.561)}
						\gppoint{gp mark 1}{(2.129,3.809)}
						\gppoint{gp mark 1}{(2.402,2.969)}
						\gppoint{gp mark 1}{(2.437,3.013)}
						\gppoint{gp mark 1}{(2.471,3.080)}
						\gppoint{gp mark 1}{(2.505,3.058)}
						\gppoint{gp mark 1}{(2.539,2.969)}
						\gppoint{gp mark 1}{(2.573,3.058)}
						\gppoint{gp mark 1}{(2.608,3.013)}
						\gppoint{gp mark 1}{(2.676,2.881)}
						\gppoint{gp mark 1}{(2.745,2.814)}
						\gppoint{gp mark 1}{(2.813,2.660)}
						\gppoint{gp mark 1}{(2.881,2.660)}
						\gppoint{gp mark 1}{(2.916,2.703)}
						\gppoint{gp mark 1}{(3.018,2.593)}
						\gppoint{gp mark 1}{(3.052,2.593)}
						\gppoint{gp mark 1}{(3.087,2.638)}
						\gppoint{gp mark 1}{(3.155,2.682)}
						\gppoint{gp mark 1}{(3.189,2.549)}
						\gppoint{gp mark 1}{(3.223,2.528)}
						\gppoint{gp mark 1}{(3.258,2.504)}
						\gppoint{gp mark 1}{(3.326,2.504)}
						\gppoint{gp mark 1}{(3.395,2.483)}
						\gppoint{gp mark 1}{(3.497,2.528)}
						\gppoint{gp mark 1}{(3.531,2.528)}
						\gppoint{gp mark 1}{(3.566,2.504)}
						\gppoint{gp mark 1}{(3.600,2.461)}
						\gppoint{gp mark 1}{(3.634,2.483)}
						\gppoint{gp mark 1}{(3.668,2.439)}
						\gppoint{gp mark 1}{(3.737,2.284)}
						\gppoint{gp mark 1}{(3.771,2.329)}
						\gppoint{gp mark 1}{(3.805,2.284)}
						\gppoint{gp mark 1}{(3.873,2.284)}
						\gppoint{gp mark 1}{(3.908,2.240)}
						\gppoint{gp mark 1}{(3.942,2.262)}
						\gppoint{gp mark 1}{(3.976,2.284)}
						\gppoint{gp mark 1}{(4.010,2.329)}
						\gppoint{gp mark 1}{(4.113,2.106)}
						\gppoint{gp mark 1}{(4.147,2.085)}
						\gppoint{gp mark 1}{(4.181,2.130)}
						\gppoint{gp mark 1}{(4.216,2.173)}
						\gppoint{gp mark 1}{(4.284,2.195)}
						\gppoint{gp mark 1}{(4.318,2.195)}
						\gppoint{gp mark 1}{(4.352,2.130)}
						\gppoint{gp mark 1}{(4.387,2.195)}
						\gppoint{gp mark 1}{(4.421,2.240)}
						\gppoint{gp mark 1}{(4.455,2.085)}
						\gppoint{gp mark 1}{(4.489,2.019)}
						\gppoint{gp mark 1}{(4.523,2.106)}
						\gppoint{gp mark 1}{(4.558,2.063)}
						\gppoint{gp mark 1}{(4.592,2.041)}
						\gppoint{gp mark 1}{(4.626,2.041)}
						\gppoint{gp mark 1}{(4.660,2.085)}
						\gppoint{gp mark 1}{(4.694,2.063)}
						\gppoint{gp mark 1}{(4.763,2.085)}
						\gppoint{gp mark 1}{(4.797,2.106)}
						\gppoint{gp mark 1}{(4.831,2.130)}
						\gppoint{gp mark 1}{(4.900,2.019)}
						\gppoint{gp mark 1}{(4.934,2.085)}
						\gppoint{gp mark 1}{(4.968,1.931)}
						\gppoint{gp mark 1}{(5.002,1.996)}
						\gppoint{gp mark 1}{(5.037,2.041)}
						\gppoint{gp mark 1}{(5.071,2.019)}
						\gppoint{gp mark 1}{(5.105,1.907)}
						\gppoint{gp mark 1}{(5.139,1.952)}
						\gppoint{gp mark 1}{(5.173,1.931)}
						\gppoint{gp mark 1}{(5.208,1.907)}
						\gppoint{gp mark 1}{(5.276,1.820)}
						\gppoint{gp mark 1}{(5.310,1.797)}
						\gppoint{gp mark 1}{(5.344,1.775)}
						\gppoint{gp mark 1}{(5.379,1.753)}
						\gppoint{gp mark 1}{(5.413,1.775)}
						\gppoint{gp mark 1}{(5.447,1.753)}
						\gppoint{gp mark 1}{(5.481,1.775)}
						\gppoint{gp mark 1}{(5.516,1.753)}
						\gppoint{gp mark 1}{(5.550,1.797)}
						\gppoint{gp mark 1}{(5.584,1.842)}
						\gppoint{gp mark 1}{(5.618,1.952)}
						\gppoint{gp mark 1}{(5.652,1.907)}
						\gppoint{gp mark 1}{(5.687,1.864)}
						\gppoint{gp mark 1}{(5.721,1.864)}
						\gppoint{gp mark 1}{(5.755,1.842)}
						\gppoint{gp mark 1}{(5.789,1.907)}
						\gppoint{gp mark 1}{(5.823,1.886)}
						\gppoint{gp mark 1}{(5.858,1.842)}
						\gppoint{gp mark 1}{(5.892,1.864)}
						\gppoint{gp mark 1}{(5.960,1.797)}
						\gppoint{gp mark 1}{(5.994,1.864)}
						\gppoint{gp mark 1}{(6.029,1.775)}
						\gppoint{gp mark 1}{(6.063,1.775)}
						\gppoint{gp mark 1}{(6.097,1.842)}
						\gppoint{gp mark 1}{(6.131,1.820)}
						\gppoint{gp mark 1}{(6.166,1.732)}
						\gppoint{gp mark 1}{(6.200,1.797)}
						\gppoint{gp mark 1}{(6.234,1.775)}
						\gppoint{gp mark 1}{(6.268,1.708)}
						\gppoint{gp mark 1}{(6.302,1.732)}
						\gppoint{gp mark 1}{(6.337,1.732)}
						\gppoint{gp mark 1}{(6.371,1.732)}
						\gppoint{gp mark 1}{(6.405,1.820)}
						\gppoint{gp mark 1}{(6.439,1.842)}
						\gppoint{gp mark 1}{(6.473,1.842)}
						\gppoint{gp mark 1}{(6.508,1.864)}
						\gppoint{gp mark 1}{(6.542,1.775)}
						\gppoint{gp mark 1}{(6.576,1.753)}
						\gppoint{gp mark 1}{(6.610,1.732)}
						\gppoint{gp mark 1}{(6.644,1.775)}
						\gppoint{gp mark 1}{(6.679,1.708)}
						\gppoint{gp mark 1}{(6.713,1.708)}
						\gppoint{gp mark 1}{(6.747,1.753)}
						\gppoint{gp mark 1}{(6.781,1.775)}
						\gppoint{gp mark 1}{(6.816,1.753)}
						\gppoint{gp mark 1}{(6.884,1.797)}
						\gppoint{gp mark 1}{(6.918,1.753)}
						\gppoint{gp mark 1}{(6.952,1.732)}
						\gppoint{gp mark 1}{(6.987,1.732)}
						\gppoint{gp mark 1}{(7.021,1.732)}
						\gppoint{gp mark 1}{(7.055,1.732)}
						\gppoint{gp mark 1}{(7.089,1.753)}
						\gppoint{gp mark 1}{(7.123,1.687)}
						\gppoint{gp mark 1}{(7.226,1.687)}
						\gppoint{gp mark 1}{(7.397,1.533)}
						\gppoint{gp mark 1}{(7.431,1.598)}
						\gppoint{gp mark 1}{(7.465,1.533)}
						\gppoint{gp mark 1}{(7.500,1.554)}
						\gppoint{gp mark 1}{(7.534,1.621)}
						\gppoint{gp mark 1}{(7.568,1.509)}
						\gppoint{gp mark 1}{(7.602,1.554)}
						\gppoint{gp mark 1}{(7.637,1.487)}
						\gppoint{gp mark 1}{(7.671,1.487)}
						\gppoint{gp mark 1}{(7.705,1.487)}
						\gppoint{gp mark 1}{(7.739,1.466)}
						\gppoint{gp mark 1}{(7.773,1.487)}
						\gppoint{gp mark 1}{(7.808,1.708)}
						\gppoint{gp mark 1}{(7.842,1.643)}
						\gppoint{gp mark 1}{(7.876,1.708)}
						\gppoint{gp mark 1}{(7.910,1.687)}
						\gppoint{gp mark 1}{(7.944,1.687)}
						\gppoint{gp mark 1}{(7.979,1.732)}
						\gppoint{gp mark 1}{(8.013,1.820)}
						\gppoint{gp mark 1}{(8.047,1.775)}
						\gppoint{gp mark 1}{(8.081,1.753)}
						\gppoint{gp mark 1}{(8.115,1.753)}
						\gppoint{gp mark 1}{(8.150,1.797)}
						\gppoint{gp mark 1}{(8.184,1.820)}
						\gppoint{gp mark 1}{(8.218,1.732)}
						\gppoint{gp mark 1}{(8.252,1.708)}
						\gppoint{gp mark 1}{(8.287,1.708)}
						\gppoint{gp mark 1}{(8.321,1.687)}
						\gppoint{gp mark 1}{(8.355,1.820)}
						\gppoint{gp mark 1}{(8.389,1.753)}
						\gppoint{gp mark 1}{(8.423,1.775)}
						\gppoint{gp mark 1}{(8.458,1.732)}
						\gppoint{gp mark 1}{(8.492,1.708)}
						\gppoint{gp mark 1}{(8.526,1.797)}
						\gppoint{gp mark 1}{(8.560,1.797)}
						\gppoint{gp mark 1}{(8.594,1.775)}
						\gppoint{gp mark 1}{(8.629,1.665)}
						\gppoint{gp mark 1}{(8.663,1.687)}
						\gppoint{gp mark 1}{(8.697,1.732)}
						\gppoint{gp mark 1}{(8.731,1.753)}
						\gppoint{gp mark 1}{(8.765,1.775)}
						\gppoint{gp mark 1}{(8.800,1.732)}
						\gppoint{gp mark 1}{(8.834,1.708)}
						\gppoint{gp mark 1}{(8.868,1.708)}
						\gppoint{gp mark 1}{(8.902,1.687)}
						\gppoint{gp mark 1}{(8.937,1.775)}
						\gppoint{gp mark 1}{(8.971,1.753)}
						\gppoint{gp mark 1}{(9.005,1.708)}
						\gppoint{gp mark 1}{(9.039,1.775)}
						\gppoint{gp mark 1}{(9.073,1.753)}
						\gppoint{gp mark 1}{(9.108,1.864)}
						\gppoint{gp mark 1}{(9.142,1.775)}
						\gppoint{gp mark 1}{(9.176,1.775)}
						\gppoint{gp mark 1}{(9.210,1.820)}
						\gppoint{gp mark 1}{(9.244,1.732)}
						\gppoint{gp mark 1}{(9.279,1.753)}
						\gppoint{gp mark 1}{(9.313,1.753)}
						\gppoint{gp mark 1}{(9.347,1.753)}
						\gppoint{gp mark 1}{(9.381,1.753)}
						\gppoint{gp mark 1}{(9.415,1.687)}
						\gppoint{gp mark 1}{(9.450,1.753)}
						\gppoint{gp mark 1}{(9.484,1.775)}
						\gppoint{gp mark 1}{(9.518,1.753)}
						\gppoint{gp mark 1}{(9.552,1.753)}
						\gppoint{gp mark 1}{(9.587,1.732)}
						\gppoint{gp mark 1}{(9.621,1.753)}
						\gppoint{gp mark 1}{(9.655,1.753)}
						\gppoint{gp mark 1}{(9.689,1.775)}
						\gppoint{gp mark 1}{(9.723,1.775)}
						\gppoint{gp mark 1}{(9.758,1.753)}
						\gppoint{gp mark 1}{(9.792,1.753)}
						\gppoint{gp mark 1}{(9.826,1.732)}
						\gppoint{gp mark 1}{(9.860,1.732)}
						\gppoint{gp mark 1}{(9.894,1.775)}
						\gppoint{gp mark 1}{(9.929,1.753)}
						\gppoint{gp mark 1}{(9.963,1.732)}
						\gppoint{gp mark 1}{(9.997,1.775)}
						\gppoint{gp mark 1}{(10.031,1.708)}
						\gppoint{gp mark 1}{(10.065,1.732)}
						\gppoint{gp mark 1}{(10.100,1.797)}
						\gppoint{gp mark 1}{(10.134,1.708)}
						\gppoint{gp mark 1}{(10.168,1.708)}
						\gppoint{gp mark 1}{(10.202,1.708)}
						\gppoint{gp mark 1}{(10.237,1.775)}
						\gppoint{gp mark 1}{(10.271,1.753)}
						\gppoint{gp mark 1}{(10.305,1.797)}
						\gppoint{gp mark 1}{(10.339,1.775)}
						\gppoint{gp mark 1}{(10.373,1.775)}
						\gppoint{gp mark 1}{(10.408,1.775)}
						\gppoint{gp mark 1}{(10.442,1.775)}
						\gppoint{gp mark 1}{(10.476,1.775)}
						\gppoint{gp mark 1}{(10.510,1.797)}
						\gppoint{gp mark 1}{(10.544,1.775)}
						\gppoint{gp mark 1}{(10.579,1.753)}
						\gppoint{gp mark 1}{(10.613,1.775)}
						\gppoint{gp mark 1}{(10.647,1.775)}
						\gppoint{gp mark 1}{(10.681,1.732)}
						\gppoint{gp mark 1}{(10.715,1.820)}
						\gppoint{gp mark 1}{(10.750,1.732)}
						\gppoint{gp mark 1}{(10.784,1.775)}
						\gppoint{gp mark 1}{(10.818,1.797)}
						\gppoint{gp mark 1}{(10.852,1.797)}
						\gppoint{gp mark 1}{(10.886,1.775)}
						\gppoint{gp mark 1}{(10.921,1.775)}
						\gppoint{gp mark 1}{(10.955,1.797)}
						\gppoint{gp mark 1}{(10.989,1.797)}
						\gppoint{gp mark 1}{(11.023,1.797)}
						\gppoint{gp mark 1}{(11.058,1.732)}
						\gppoint{gp mark 1}{(11.092,1.775)}
						\gppoint{gp mark 1}{(11.126,1.732)}
						\gppoint{gp mark 1}{(11.160,1.753)}
						\gppoint{gp mark 1}{(11.194,1.775)}
						\gppoint{gp mark 1}{(11.229,1.775)}
						\gppoint{gp mark 1}{(11.263,1.753)}
						\gppoint{gp mark 1}{(11.297,1.797)}
						\gppoint{gp mark 1}{(11.331,1.775)}
						\gppoint{gp mark 1}{(11.365,1.775)}
						\gppoint{gp mark 1}{(11.400,1.797)}
						\gppoint{gp mark 1}{(11.434,1.797)}
						\gppoint{gp mark 1}{(11.468,1.775)}
						\gpsetpointsize{2.80}
						\gpcolor{rgb color={0.000,0.000,1.000}}
						\gppoint{gp mark 7}{(1.718,7.480)}
						\gppoint{gp mark 7}{(1.752,7.082)}
						\gppoint{gp mark 7}{(1.787,6.530)}
						\gppoint{gp mark 7}{(1.821,6.506)}
						\gppoint{gp mark 7}{(1.855,5.291)}
						\gppoint{gp mark 7}{(1.992,4.429)}
						\gppoint{gp mark 7}{(2.095,3.698)}
						\gppoint{gp mark 7}{(2.163,3.677)}
						\gppoint{gp mark 7}{(2.197,3.610)}
						\gppoint{gp mark 7}{(2.231,3.389)}
						\gppoint{gp mark 7}{(2.266,3.300)}
						\gppoint{gp mark 7}{(2.300,3.190)}
						\gppoint{gp mark 7}{(2.334,3.058)}
						\gppoint{gp mark 7}{(2.368,2.969)}
						\gppoint{gp mark 7}{(2.642,2.859)}
						\gppoint{gp mark 7}{(2.710,2.792)}
						\gppoint{gp mark 7}{(2.779,2.660)}
						\gppoint{gp mark 7}{(2.847,2.638)}
						\gppoint{gp mark 7}{(2.950,2.615)}
						\gppoint{gp mark 7}{(2.984,2.571)}
						\gppoint{gp mark 7}{(3.121,2.504)}
						\gppoint{gp mark 7}{(3.292,2.483)}
						\gppoint{gp mark 7}{(3.360,2.461)}
						\gppoint{gp mark 7}{(3.429,2.439)}
						\gppoint{gp mark 7}{(3.463,2.416)}
						\gppoint{gp mark 7}{(3.702,2.284)}
						\gppoint{gp mark 7}{(3.839,2.240)}
						\gppoint{gp mark 7}{(4.044,2.130)}
						\gppoint{gp mark 7}{(4.079,2.041)}
						\gppoint{gp mark 7}{(4.250,2.019)}
						\gppoint{gp mark 7}{(4.729,1.974)}
						\gppoint{gp mark 7}{(4.866,1.907)}
						\gppoint{gp mark 7}{(5.242,1.732)}
						\gppoint{gp mark 7}{(5.926,1.708)}
						\gppoint{gp mark 7}{(6.850,1.665)}
						\gppoint{gp mark 7}{(7.158,1.643)}
						\gppoint{gp mark 7}{(7.192,1.621)}
						\gppoint{gp mark 7}{(7.260,1.554)}
						\gppoint{gp mark 7}{(7.294,1.533)}
						\gppoint{gp mark 7}{(7.329,1.509)}
						\gppoint{gp mark 7}{(7.363,1.466)}
						\gpcolor{rgb color={1.000,1.000,0.000}}
						\gpsetpointsize{5.40}
						\gppoint{gp mark 5}{(2.779,2.660)}
						\gpcolor{rgb color={0.000,0.000,0.000}}
						\gpsetpointsize{5.80}
						\gppoint{gp mark 4}{(2.779,2.660)}
						\gpsetpointsize{5.40}
						\gppoint{gp mark 5}{(1.718,1.466)}
						\gpcolor{color=gp lt color border}
						\draw[gp path] (1.684,7.702)--(1.684,1.165)--(11.947,1.165)--(11.947,7.702)--cycle;
						%% coordinates of the plot area
						\gpdefrectangularnode{gp plot 1}{\pgfpoint{1.684cm}{1.165cm}}{\pgfpoint{11.947cm}{7.702cm}}
						\end{tikzpicture}
						\subcaption{Training set of KP}
						\label{fig:TF22}
					\end{center}				
				\end{minipage}
			\end{center}
			\caption{An illustration of the performance of the proposed approach for selecting the best subset of features on the reduced training set} 
			\label{fig:TrainFrontier2}
		\end{figure}

		It is evident that due to the change in the training set, we need to apply our proposed approach for best subset selection of features again. So, similar to Section~\ref{subsec:CompTraingTesting}, Figure~\ref{fig:TrainFrontier2} shows the approximated nondominated frontier for each class of optimization problems based on the reduced training set. By comparing the ideal points in Figures~\ref{fig:TrainFrontier} and \ref{fig:TrainFrontier2}, an immediate improvement in the (ideal) accuracy can be observed. In fact the absolute difference between the error of the ideal points (in these figures) is around 12\% and 7\% for AP and KP instances, respectively. Similar improvements can be observed by comparing the selected approximated nondominated points in Figures~\ref{fig:TrainFrontier} and \ref{fig:TrainFrontier2}.
		
\begin{figure}[ht]
			\begin{center}
				\begin{minipage}{0.45\linewidth}
					\begin{center}
						\begin{tikzpicture}[scale=0.60,
						axis/.style={ ->, >=stealth'},line/.style={very thick},]
						\path (0.000,0.000) rectangle (12.500,8.750);
						\gpcolor{color=gp lt color border}
						\gpsetlinetype{gp lt border}
						\gpsetdashtype{gp dt solid}
						\gpsetlinewidth{1.00}
						\draw[gp path] (1.684,1.165)--(1.504,1.165);
						\draw[gp path] (11.947,1.165)--(12.127,1.165);
						\node[gp node right] at (1.320,1.165) {$0.43$};
						\draw[gp path] (1.684,1.819)--(1.504,1.819);
						\draw[gp path] (11.947,1.819)--(12.127,1.819);
						\node[gp node right] at (1.320,1.819) {$0.44$};
						\draw[gp path] (1.684,2.472)--(1.504,2.472);
						\draw[gp path] (11.947,2.472)--(12.127,2.472);
						\node[gp node right] at (1.320,2.472) {$0.45$};
						\draw[gp path] (1.684,3.126)--(1.504,3.126);
						\draw[gp path] (11.947,3.126)--(12.127,3.126);
						\node[gp node right] at (1.320,3.126) {$0.46$};
						\draw[gp path] (1.684,3.780)--(1.504,3.780);
						\draw[gp path] (11.947,3.780)--(12.127,3.780);
						\node[gp node right] at (1.320,3.780) {$0.47$};
						\draw[gp path] (1.684,4.434)--(1.504,4.434);
						\draw[gp path] (11.947,4.434)--(12.127,4.434);
						\node[gp node right] at (1.320,4.434) {$0.48$};
						\draw[gp path] (1.684,5.087)--(1.504,5.087);
						\draw[gp path] (11.947,5.087)--(12.127,5.087);
						\node[gp node right] at (1.320,5.087) {$0.49$};
						\draw[gp path] (1.684,5.741)--(1.504,5.741);
						\draw[gp path] (11.947,5.741)--(12.127,5.741);
						\node[gp node right] at (1.320,5.741) {$0.5$};
						\draw[gp path] (1.684,6.395)--(1.504,6.395);
						\draw[gp path] (11.947,6.395)--(12.127,6.395);
						\node[gp node right] at (1.320,6.395) {$0.51$};
						\draw[gp path] (1.684,7.048)--(1.504,7.048);
						\draw[gp path] (11.947,7.048)--(12.127,7.048);
						\node[gp node right] at (1.320,7.048) {$0.52$};
						\draw[gp path] (1.684,7.702)--(1.504,7.702);
						\draw[gp path] (11.947,7.702)--(12.127,7.702);
						\node[gp node right] at (1.320,7.702) {$0.53$};
						\draw[gp path] (1.684,1.165)--(1.684,0.985);
						\draw[gp path] (1.684,7.702)--(1.684,7.882);
						\node[gp node center] at (1.684,0.677) {$0$};
						\draw[gp path] (3.395,1.165)--(3.395,0.985);
						\draw[gp path] (3.395,7.702)--(3.395,7.882);
						\node[gp node center] at (3.395,0.677) {$50$};
						\draw[gp path] (5.105,1.165)--(5.105,0.985);
						\draw[gp path] (5.105,7.702)--(5.105,7.882);
						\node[gp node center] at (5.105,0.677) {$100$};
						\draw[gp path] (6.816,1.165)--(6.816,0.985);
						\draw[gp path] (6.816,7.702)--(6.816,7.882);
						\node[gp node center] at (6.816,0.677) {$150$};
						\draw[gp path] (8.526,1.165)--(8.526,0.985);
						\draw[gp path] (8.526,7.702)--(8.526,7.882);
						\node[gp node center] at (8.526,0.677) {$200$};
						\draw[gp path] (10.237,1.165)--(10.237,0.985);
						\draw[gp path] (10.237,7.702)--(10.237,7.882);
						\node[gp node center] at (10.237,0.677) {$250$};
						\draw[gp path] (11.947,1.165)--(11.947,0.985);
						\draw[gp path] (11.947,7.702)--(11.947,7.882);
						\node[gp node center] at (11.947,0.677) {$300$};
						\draw[gp path] (1.684,7.702)--(1.684,1.165)--(11.947,1.165)--(11.947,7.702)--cycle;
						\node[gp node center,rotate=-270] at (-0.476,4.433) {Error};
						\node[gp node center] at (6.815,-0.315) {Number of features};
						%					\node[gp node center,scale=0.8,font={\fontsize{14.0pt}{16.8pt}\selectfont}] at (6.815,8.287) {\textbf{Testing set error AP instances}};
						\gpcolor{rgb color={1.000,0.000,0.000}}
						\gpsetpointsize{2.00}
						\gppoint{gp mark 1}{(1.821,2.629)}
						\gppoint{gp mark 1}{(1.923,3.146)}
						\gppoint{gp mark 1}{(1.958,1.590)}
						\gppoint{gp mark 1}{(2.026,3.669)}
						\gppoint{gp mark 1}{(2.060,4.185)}
						\gppoint{gp mark 1}{(2.095,5.224)}
						\gppoint{gp mark 1}{(2.197,5.224)}
						\gppoint{gp mark 1}{(2.300,5.224)}
						\gppoint{gp mark 1}{(2.334,3.669)}
						\gppoint{gp mark 1}{(2.368,4.702)}
						\gppoint{gp mark 1}{(2.402,4.702)}
						\gppoint{gp mark 1}{(2.437,6.257)}
						\gppoint{gp mark 1}{(2.471,5.741)}
						\gppoint{gp mark 1}{(2.505,5.224)}
						\gppoint{gp mark 1}{(2.539,4.185)}
						\gppoint{gp mark 1}{(2.573,4.702)}
						\gppoint{gp mark 1}{(2.608,3.669)}
						\gppoint{gp mark 1}{(2.642,4.185)}
						\gppoint{gp mark 1}{(2.676,4.185)}
						\gppoint{gp mark 1}{(2.745,4.185)}
						\gppoint{gp mark 1}{(2.779,4.702)}
						\gppoint{gp mark 1}{(2.813,4.702)}
						\gppoint{gp mark 1}{(2.881,4.702)}
						\gppoint{gp mark 1}{(2.984,2.629)}
						\gppoint{gp mark 1}{(3.018,2.106)}
						\gppoint{gp mark 1}{(3.121,2.629)}
						\gppoint{gp mark 1}{(3.155,3.146)}
						\gppoint{gp mark 1}{(3.189,2.629)}
						\gppoint{gp mark 1}{(3.223,2.106)}
						\gppoint{gp mark 1}{(3.258,3.146)}
						\gppoint{gp mark 1}{(3.292,3.669)}
						\gppoint{gp mark 1}{(3.326,2.106)}
						\gppoint{gp mark 1}{(3.360,2.106)}
						\gppoint{gp mark 1}{(3.395,2.106)}
						\gppoint{gp mark 1}{(3.429,3.669)}
						\gppoint{gp mark 1}{(3.497,3.669)}
						\gppoint{gp mark 1}{(3.531,4.185)}
						\gppoint{gp mark 1}{(3.566,3.669)}
						\gppoint{gp mark 1}{(3.600,3.669)}
						\gppoint{gp mark 1}{(3.634,2.629)}
						\gppoint{gp mark 1}{(3.668,2.106)}
						\gppoint{gp mark 1}{(3.737,2.106)}
						\gppoint{gp mark 1}{(3.771,1.590)}
						\gppoint{gp mark 1}{(3.873,4.185)}
						\gppoint{gp mark 1}{(3.908,3.669)}
						\gppoint{gp mark 1}{(3.942,4.185)}
						\gppoint{gp mark 1}{(3.976,5.741)}
						\gppoint{gp mark 1}{(4.010,6.780)}
						\gppoint{gp mark 1}{(4.044,6.257)}
						\gppoint{gp mark 1}{(4.079,6.257)}
						\gppoint{gp mark 1}{(4.113,6.257)}
						\gppoint{gp mark 1}{(4.147,6.780)}
						\gppoint{gp mark 1}{(4.181,5.741)}
						\gppoint{gp mark 1}{(4.216,5.224)}
						\gppoint{gp mark 1}{(4.250,5.224)}
						\gppoint{gp mark 1}{(4.284,5.741)}
						\gppoint{gp mark 1}{(4.318,5.224)}
						\gppoint{gp mark 1}{(4.352,6.257)}
						\gppoint{gp mark 1}{(4.421,5.741)}
						\gppoint{gp mark 1}{(4.455,6.780)}
						\gppoint{gp mark 1}{(4.489,5.741)}
						\gppoint{gp mark 1}{(4.523,6.257)}
						\gppoint{gp mark 1}{(4.694,6.780)}
						\gppoint{gp mark 1}{(4.729,6.780)}
						\gppoint{gp mark 1}{(4.763,6.780)}
						\gppoint{gp mark 1}{(4.797,6.780)}
						\gppoint{gp mark 1}{(4.831,6.780)}
						\gppoint{gp mark 1}{(4.866,6.780)}
						\gppoint{gp mark 1}{(4.900,6.780)}
						\gppoint{gp mark 1}{(4.934,6.780)}
						\gppoint{gp mark 1}{(4.968,6.780)}
						\gppoint{gp mark 1}{(5.002,6.780)}
						\gppoint{gp mark 1}{(5.071,6.780)}
						\gppoint{gp mark 1}{(5.105,7.297)}
						\gppoint{gp mark 1}{(5.139,7.297)}
						\gppoint{gp mark 1}{(5.173,7.297)}
						\gppoint{gp mark 1}{(5.208,7.297)}
						\gppoint{gp mark 1}{(5.242,7.297)}
						\gppoint{gp mark 1}{(5.276,7.297)}
						\gppoint{gp mark 1}{(5.310,6.780)}
						\gppoint{gp mark 1}{(5.344,6.780)}
						\gppoint{gp mark 1}{(5.379,6.780)}
						\gppoint{gp mark 1}{(5.413,7.297)}
						\gppoint{gp mark 1}{(5.447,6.780)}
						\gppoint{gp mark 1}{(5.481,7.297)}
						\gppoint{gp mark 1}{(5.516,6.780)}
						\gppoint{gp mark 1}{(5.550,6.780)}
						\gppoint{gp mark 1}{(5.584,6.257)}
						\gppoint{gp mark 1}{(5.618,6.780)}
						\gppoint{gp mark 1}{(5.652,6.780)}
						\gppoint{gp mark 1}{(5.687,6.257)}
						\gppoint{gp mark 1}{(5.721,6.780)}
						\gppoint{gp mark 1}{(5.789,4.702)}
						\gppoint{gp mark 1}{(5.823,5.741)}
						\gppoint{gp mark 1}{(5.858,5.741)}
						\gppoint{gp mark 1}{(5.892,5.741)}
						\gppoint{gp mark 1}{(5.960,6.257)}
						\gppoint{gp mark 1}{(5.994,6.257)}
						\gppoint{gp mark 1}{(6.029,6.257)}
						\gppoint{gp mark 1}{(6.063,5.224)}
						\gppoint{gp mark 1}{(6.097,5.741)}
						\gppoint{gp mark 1}{(6.131,5.741)}
						\gppoint{gp mark 1}{(6.166,5.741)}
						\gppoint{gp mark 1}{(6.200,5.741)}
						\gppoint{gp mark 1}{(6.234,5.741)}
						\gppoint{gp mark 1}{(6.302,6.257)}
						\gppoint{gp mark 1}{(6.337,5.741)}
						\gppoint{gp mark 1}{(6.371,5.741)}
						\gppoint{gp mark 1}{(6.439,6.257)}
						\gppoint{gp mark 1}{(6.473,6.257)}
						\gppoint{gp mark 1}{(6.508,6.257)}
						\gppoint{gp mark 1}{(6.542,6.257)}
						\gppoint{gp mark 1}{(6.576,6.257)}
						\gppoint{gp mark 1}{(6.610,6.257)}
						\gppoint{gp mark 1}{(6.713,6.257)}
						\gppoint{gp mark 1}{(6.747,5.741)}
						\gppoint{gp mark 1}{(6.816,6.257)}
						\gppoint{gp mark 1}{(6.850,7.297)}
						\gppoint{gp mark 1}{(6.884,6.780)}
						\gppoint{gp mark 1}{(6.918,7.297)}
						\gppoint{gp mark 1}{(6.952,7.297)}
						\gppoint{gp mark 1}{(6.987,6.780)}
						\gppoint{gp mark 1}{(7.021,7.297)}
						\gppoint{gp mark 1}{(7.055,6.780)}
						\gppoint{gp mark 1}{(7.089,7.297)}
						\gppoint{gp mark 1}{(7.123,7.297)}
						\gppoint{gp mark 1}{(7.158,7.297)}
						\gppoint{gp mark 1}{(7.192,7.297)}
						\gppoint{gp mark 1}{(7.226,6.780)}
						\gppoint{gp mark 1}{(7.260,7.297)}
						\gppoint{gp mark 1}{(7.294,6.257)}
						\gppoint{gp mark 1}{(7.329,6.780)}
						\gppoint{gp mark 1}{(7.363,6.780)}
						\gppoint{gp mark 1}{(7.397,6.780)}
						\gppoint{gp mark 1}{(7.431,6.780)}
						\gppoint{gp mark 1}{(7.465,7.297)}
						\gppoint{gp mark 1}{(7.500,6.257)}
						\gppoint{gp mark 1}{(7.534,6.257)}
						\gppoint{gp mark 1}{(7.568,6.257)}
						\gppoint{gp mark 1}{(7.637,5.741)}
						\gppoint{gp mark 1}{(7.671,5.741)}
						\gppoint{gp mark 1}{(7.705,5.741)}
						\gppoint{gp mark 1}{(7.739,6.257)}
						\gppoint{gp mark 1}{(7.773,6.780)}
						\gppoint{gp mark 1}{(7.808,5.741)}
						\gppoint{gp mark 1}{(7.842,6.257)}
						\gppoint{gp mark 1}{(7.876,6.257)}
						\gppoint{gp mark 1}{(7.910,5.741)}
						\gppoint{gp mark 1}{(7.944,5.224)}
						\gppoint{gp mark 1}{(7.979,5.741)}
						\gppoint{gp mark 1}{(8.013,6.257)}
						\gppoint{gp mark 1}{(8.047,6.780)}
						\gppoint{gp mark 1}{(8.081,6.780)}
						\gppoint{gp mark 1}{(8.115,6.257)}
						\gppoint{gp mark 1}{(8.150,6.257)}
						\gppoint{gp mark 1}{(8.184,6.257)}
						\gppoint{gp mark 1}{(8.218,6.257)}
						\gppoint{gp mark 1}{(8.252,6.780)}
						\gppoint{gp mark 1}{(8.287,6.257)}
						\gppoint{gp mark 1}{(8.321,6.780)}
						\gppoint{gp mark 1}{(8.355,6.257)}
						\gppoint{gp mark 1}{(8.389,6.257)}
						\gppoint{gp mark 1}{(8.423,6.257)}
						\gppoint{gp mark 1}{(8.458,6.780)}
						\gppoint{gp mark 1}{(8.492,6.257)}
						\gppoint{gp mark 1}{(8.526,7.297)}
						\gppoint{gp mark 1}{(8.560,6.780)}
						\gppoint{gp mark 1}{(8.594,6.257)}
						\gppoint{gp mark 1}{(8.629,6.780)}
						\gppoint{gp mark 1}{(8.663,6.257)}
						\gppoint{gp mark 1}{(8.697,6.780)}
						\gppoint{gp mark 1}{(8.731,6.257)}
						\gppoint{gp mark 1}{(8.765,6.780)}
						\gppoint{gp mark 1}{(8.800,7.297)}
						\gppoint{gp mark 1}{(8.834,6.780)}
						\gppoint{gp mark 1}{(8.868,6.257)}
						\gppoint{gp mark 1}{(8.902,6.780)}
						\gppoint{gp mark 1}{(8.937,6.257)}
						\gppoint{gp mark 1}{(8.971,6.257)}
						\gppoint{gp mark 1}{(9.005,6.257)}
						\gppoint{gp mark 1}{(9.039,6.257)}
						\gppoint{gp mark 1}{(9.073,6.257)}
						\gppoint{gp mark 1}{(9.108,6.780)}
						\gppoint{gp mark 1}{(9.142,6.257)}
						\gppoint{gp mark 1}{(9.176,6.257)}
						\gppoint{gp mark 1}{(9.210,6.257)}
						\gppoint{gp mark 1}{(9.244,6.257)}
						\gppoint{gp mark 1}{(9.279,6.257)}
						\gppoint{gp mark 1}{(9.313,6.257)}
						\gppoint{gp mark 1}{(9.347,6.257)}
						\gppoint{gp mark 1}{(9.381,6.257)}
						\gppoint{gp mark 1}{(9.415,6.257)}
						\gppoint{gp mark 1}{(9.450,6.257)}
						\gppoint{gp mark 1}{(9.484,6.257)}
						\gppoint{gp mark 1}{(9.518,6.257)}
						\gppoint{gp mark 1}{(9.552,6.257)}
						\gppoint{gp mark 1}{(9.621,6.257)}
						\gppoint{gp mark 1}{(9.655,6.780)}
						\gppoint{gp mark 1}{(9.689,6.780)}
						\gppoint{gp mark 1}{(9.723,6.780)}
						\gppoint{gp mark 1}{(9.758,6.257)}
						\gppoint{gp mark 1}{(9.792,6.780)}
						\gppoint{gp mark 1}{(9.826,6.780)}
						\gppoint{gp mark 1}{(9.860,6.780)}
						\gppoint{gp mark 1}{(9.894,6.257)}
						\gppoint{gp mark 1}{(9.929,6.257)}
						\gppoint{gp mark 1}{(9.963,6.780)}
						\gppoint{gp mark 1}{(9.997,6.780)}
						\gppoint{gp mark 1}{(10.031,6.257)}
						\gppoint{gp mark 1}{(10.065,6.257)}
						\gppoint{gp mark 1}{(10.100,6.257)}
						\gppoint{gp mark 1}{(10.134,6.257)}
						\gppoint{gp mark 1}{(10.168,6.257)}
						\gppoint{gp mark 1}{(10.202,6.257)}
						\gppoint{gp mark 1}{(10.237,6.780)}
						\gppoint{gp mark 1}{(10.271,6.257)}
						\gppoint{gp mark 1}{(10.305,6.257)}
						\gppoint{gp mark 1}{(10.339,6.257)}
						\gppoint{gp mark 1}{(10.373,6.257)}
						\gppoint{gp mark 1}{(10.408,6.780)}
						\gppoint{gp mark 1}{(10.442,6.257)}
						\gppoint{gp mark 1}{(10.476,6.780)}
						\gppoint{gp mark 1}{(10.510,6.780)}
						\gppoint{gp mark 1}{(10.544,6.257)}
						\gppoint{gp mark 1}{(10.613,6.257)}
						\gppoint{gp mark 1}{(10.647,6.257)}
						\gppoint{gp mark 1}{(10.681,6.257)}
						\gppoint{gp mark 1}{(10.715,6.257)}
						\gppoint{gp mark 1}{(10.750,6.257)}
						\gppoint{gp mark 1}{(10.784,6.257)}
						\gppoint{gp mark 1}{(10.818,6.257)}
						\gppoint{gp mark 1}{(10.852,6.257)}
						\gppoint{gp mark 1}{(10.886,6.780)}
						\gppoint{gp mark 1}{(10.921,6.257)}
						\gppoint{gp mark 1}{(10.955,6.257)}
						\gpsetpointsize{2.80}
						\gpcolor{rgb color={0.000,0.000,1.000}}
						\gppoint{gp mark 7}{(1.718,3.669)}
						\gppoint{gp mark 7}{(1.752,5.224)}
						\gppoint{gp mark 7}{(1.787,4.702)}
						\gppoint{gp mark 7}{(1.855,3.669)}
						\gppoint{gp mark 7}{(1.889,4.185)}
						\gppoint{gp mark 7}{(1.992,3.669)}
						\gppoint{gp mark 7}{(2.129,2.106)}
						\gppoint{gp mark 7}{(2.163,2.629)}
						\gppoint{gp mark 7}{(2.231,4.702)}
						\gppoint{gp mark 7}{(2.266,3.669)}
						\gppoint{gp mark 7}{(2.710,4.185)}
						\gppoint{gp mark 7}{(2.847,4.702)}
						\gppoint{gp mark 7}{(2.916,4.185)}
						\gppoint{gp mark 7}{(2.950,2.629)}
						\gppoint{gp mark 7}{(3.052,2.106)}
						\gppoint{gp mark 7}{(3.087,2.629)}
						\gppoint{gp mark 7}{(3.463,4.185)}
						\gppoint{gp mark 7}{(3.702,2.629)}
						\gppoint{gp mark 7}{(3.805,3.146)}
						\gppoint{gp mark 7}{(3.839,4.185)}
						\gppoint{gp mark 7}{(4.387,5.741)}
						\gppoint{gp mark 7}{(4.558,6.257)}
						\gppoint{gp mark 7}{(4.592,6.257)}
						\gppoint{gp mark 7}{(4.626,6.780)}
						\gppoint{gp mark 7}{(4.660,6.780)}
						\gppoint{gp mark 7}{(5.037,6.780)}
						\gppoint{gp mark 7}{(5.755,7.297)}
						\gppoint{gp mark 7}{(5.926,6.780)}
						\gppoint{gp mark 7}{(6.268,5.741)}
						\gppoint{gp mark 7}{(6.405,6.257)}
						\gppoint{gp mark 7}{(6.644,5.741)}
						\gppoint{gp mark 7}{(6.679,6.257)}
						\gppoint{gp mark 7}{(6.781,6.257)}
						\gppoint{gp mark 7}{(7.602,6.257)}
						\gppoint{gp mark 7}{(9.587,6.780)}
						\gppoint{gp mark 7}{(10.579,6.257)}
						\gpcolor{rgb color={1.000,1.000,0.000}}
						\gpsetpointsize{5.40}
						\gppoint{gp mark 5}{(3.839,4.185)}
						\gpcolor{rgb color={0.000,0.000,0.000}}
						\gpsetpointsize{5.80}
						\gppoint{gp mark 4}{(3.839,4.185)}
						\gpcolor{color=gp lt color border}
						\draw[gp path] (1.684,7.702)--(1.684,1.165)--(11.947,1.165)--(11.947,7.702)--cycle;
						%% coordinates of the plot area
						\gpdefrectangularnode{gp plot 1}{\pgfpoint{1.684cm}{1.165cm}}{\pgfpoint{11.947cm}{7.702cm}}
						\end{tikzpicture}
						\subcaption{Testing set of AP} 					
						\label{fig:TestF12}
					\end{center}				
				\end{minipage}
				\hfil
				\begin{minipage}{0.45\linewidth}
					\begin{center}
						\begin{tikzpicture}[scale=0.60,
						axis/.style={ ->, >=stealth'},line/.style={very thick},]
						\path (0.000,0.000) rectangle (12.500,8.750);
						\gpcolor{color=gp lt color border}
						\gpsetlinetype{gp lt border}
						\gpsetdashtype{gp dt solid}
						\gpsetlinewidth{1.00}
						\draw[gp path] (1.684,1.165)--(1.504,1.165);
						\draw[gp path] (11.947,1.165)--(12.127,1.165);
						\node[gp node right] at (1.320,1.165) {$0.35$};
						\draw[gp path] (1.684,2.099)--(1.504,2.099);
						\draw[gp path] (11.947,2.099)--(12.127,2.099);
						\node[gp node right] at (1.320,2.099) {$0.4$};
						\draw[gp path] (1.684,3.033)--(1.504,3.033);
						\draw[gp path] (11.947,3.033)--(12.127,3.033);
						\node[gp node right] at (1.320,3.033) {$0.45$};
						\draw[gp path] (1.684,3.967)--(1.504,3.967);
						\draw[gp path] (11.947,3.967)--(12.127,3.967);
						\node[gp node right] at (1.320,3.967) {$0.5$};
						\draw[gp path] (1.684,4.900)--(1.504,4.900);
						\draw[gp path] (11.947,4.900)--(12.127,4.900);
						\node[gp node right] at (1.320,4.900) {$0.55$};
						\draw[gp path] (1.684,5.834)--(1.504,5.834);
						\draw[gp path] (11.947,5.834)--(12.127,5.834);
						\node[gp node right] at (1.320,5.834) {$0.6$};
						\draw[gp path] (1.684,6.768)--(1.504,6.768);
						\draw[gp path] (11.947,6.768)--(12.127,6.768);
						\node[gp node right] at (1.320,6.768) {$0.65$};
						\draw[gp path] (1.684,7.702)--(1.504,7.702);
						\draw[gp path] (11.947,7.702)--(12.127,7.702);
						\node[gp node right] at (1.320,7.702) {$0.7$};
						\draw[gp path] (1.684,1.165)--(1.684,0.985);
						\draw[gp path] (1.684,7.702)--(1.684,7.882);
						\node[gp node center] at (1.684,0.677) {$0$};
						\draw[gp path] (3.395,1.165)--(3.395,0.985);
						\draw[gp path] (3.395,7.702)--(3.395,7.882);
						\node[gp node center] at (3.395,0.677) {$50$};
						\draw[gp path] (5.105,1.165)--(5.105,0.985);
						\draw[gp path] (5.105,7.702)--(5.105,7.882);
						\node[gp node center] at (5.105,0.677) {$100$};
						\draw[gp path] (6.816,1.165)--(6.816,0.985);
						\draw[gp path] (6.816,7.702)--(6.816,7.882);
						\node[gp node center] at (6.816,0.677) {$150$};
						\draw[gp path] (8.526,1.165)--(8.526,0.985);
						\draw[gp path] (8.526,7.702)--(8.526,7.882);
						\node[gp node center] at (8.526,0.677) {$200$};
						\draw[gp path] (10.237,1.165)--(10.237,0.985);
						\draw[gp path] (10.237,7.702)--(10.237,7.882);
						\node[gp node center] at (10.237,0.677) {$250$};
						\draw[gp path] (11.947,1.165)--(11.947,0.985);
						\draw[gp path] (11.947,7.702)--(11.947,7.882);
						\node[gp node center] at (11.947,0.677) {$300$};
						\draw[gp path] (1.684,7.702)--(1.684,1.165)--(11.947,1.165)--(11.947,7.702)--cycle;
						\node[gp node center,rotate=-270] at (-0.476,4.433) {Error};
						\node[gp node center] at (6.815,-0.315) {Number of features};
						%					\node[gp node center,scale=0.8,font={\fontsize{14.0pt}{16.8pt}\selectfont}] at (6.815,8.287) {\textbf{Testing set error KP instances}};
						\gpcolor{rgb color={1.000,0.000,0.000}}
						\gpsetpointsize{2.00}
						\gppoint{gp mark 1}{(1.889,3.967)}
						\gppoint{gp mark 1}{(1.923,4.194)}
						\gppoint{gp mark 1}{(1.958,4.422)}
						\gppoint{gp mark 1}{(2.026,3.283)}
						\gppoint{gp mark 1}{(2.060,3.511)}
						\gppoint{gp mark 1}{(2.129,2.827)}
						\gppoint{gp mark 1}{(2.402,2.485)}
						\gppoint{gp mark 1}{(2.437,2.372)}
						\gppoint{gp mark 1}{(2.471,1.916)}
						\gppoint{gp mark 1}{(2.505,2.030)}
						\gppoint{gp mark 1}{(2.539,2.030)}
						\gppoint{gp mark 1}{(2.573,1.916)}
						\gppoint{gp mark 1}{(2.608,2.827)}
						\gppoint{gp mark 1}{(2.676,1.348)}
						\gppoint{gp mark 1}{(2.745,1.576)}
						\gppoint{gp mark 1}{(2.813,1.916)}
						\gppoint{gp mark 1}{(2.881,1.688)}
						\gppoint{gp mark 1}{(2.916,2.144)}
						\gppoint{gp mark 1}{(3.018,2.030)}
						\gppoint{gp mark 1}{(3.052,2.144)}
						\gppoint{gp mark 1}{(3.087,2.144)}
						\gppoint{gp mark 1}{(3.155,2.485)}
						\gppoint{gp mark 1}{(3.189,2.485)}
						\gppoint{gp mark 1}{(3.223,2.485)}
						\gppoint{gp mark 1}{(3.258,2.485)}
						\gppoint{gp mark 1}{(3.326,2.372)}
						\gppoint{gp mark 1}{(3.395,2.258)}
						\gppoint{gp mark 1}{(3.497,1.916)}
						\gppoint{gp mark 1}{(3.531,2.258)}
						\gppoint{gp mark 1}{(3.566,2.030)}
						\gppoint{gp mark 1}{(3.600,1.916)}
						\gppoint{gp mark 1}{(3.634,1.916)}
						\gppoint{gp mark 1}{(3.668,1.916)}
						\gppoint{gp mark 1}{(3.737,1.916)}
						\gppoint{gp mark 1}{(3.771,2.030)}
						\gppoint{gp mark 1}{(3.805,2.030)}
						\gppoint{gp mark 1}{(3.873,2.144)}
						\gppoint{gp mark 1}{(3.908,2.372)}
						\gppoint{gp mark 1}{(3.942,2.258)}
						\gppoint{gp mark 1}{(3.976,2.372)}
						\gppoint{gp mark 1}{(4.010,2.599)}
						\gppoint{gp mark 1}{(4.113,2.372)}
						\gppoint{gp mark 1}{(4.147,2.599)}
						\gppoint{gp mark 1}{(4.181,2.485)}
						\gppoint{gp mark 1}{(4.216,2.144)}
						\gppoint{gp mark 1}{(4.284,2.372)}
						\gppoint{gp mark 1}{(4.318,2.485)}
						\gppoint{gp mark 1}{(4.352,2.485)}
						\gppoint{gp mark 1}{(4.387,2.258)}
						\gppoint{gp mark 1}{(4.421,2.485)}
						\gppoint{gp mark 1}{(4.455,2.599)}
						\gppoint{gp mark 1}{(4.489,2.599)}
						\gppoint{gp mark 1}{(4.523,2.485)}
						\gppoint{gp mark 1}{(4.558,2.372)}
						\gppoint{gp mark 1}{(4.592,2.372)}
						\gppoint{gp mark 1}{(4.626,2.372)}
						\gppoint{gp mark 1}{(4.660,2.258)}
						\gppoint{gp mark 1}{(4.694,2.485)}
						\gppoint{gp mark 1}{(4.763,2.599)}
						\gppoint{gp mark 1}{(4.797,2.372)}
						\gppoint{gp mark 1}{(4.831,2.599)}
						\gppoint{gp mark 1}{(4.900,2.372)}
						\gppoint{gp mark 1}{(4.934,2.713)}
						\gppoint{gp mark 1}{(4.968,2.713)}
						\gppoint{gp mark 1}{(5.002,2.713)}
						\gppoint{gp mark 1}{(5.037,2.941)}
						\gppoint{gp mark 1}{(5.071,1.802)}
						\gppoint{gp mark 1}{(5.105,2.030)}
						\gppoint{gp mark 1}{(5.139,2.030)}
						\gppoint{gp mark 1}{(5.173,1.802)}
						\gppoint{gp mark 1}{(5.208,2.030)}
						\gppoint{gp mark 1}{(5.276,2.258)}
						\gppoint{gp mark 1}{(5.310,2.485)}
						\gppoint{gp mark 1}{(5.344,2.258)}
						\gppoint{gp mark 1}{(5.379,2.485)}
						\gppoint{gp mark 1}{(5.413,2.485)}
						\gppoint{gp mark 1}{(5.447,2.372)}
						\gppoint{gp mark 1}{(5.481,2.485)}
						\gppoint{gp mark 1}{(5.516,2.485)}
						\gppoint{gp mark 1}{(5.550,2.372)}
						\gppoint{gp mark 1}{(5.584,2.485)}
						\gppoint{gp mark 1}{(5.618,2.372)}
						\gppoint{gp mark 1}{(5.652,2.372)}
						\gppoint{gp mark 1}{(5.687,2.372)}
						\gppoint{gp mark 1}{(5.721,2.258)}
						\gppoint{gp mark 1}{(5.755,2.485)}
						\gppoint{gp mark 1}{(5.789,2.599)}
						\gppoint{gp mark 1}{(5.823,2.599)}
						\gppoint{gp mark 1}{(5.858,2.713)}
						\gppoint{gp mark 1}{(5.892,2.485)}
						\gppoint{gp mark 1}{(5.960,2.485)}
						\gppoint{gp mark 1}{(5.994,2.713)}
						\gppoint{gp mark 1}{(6.029,2.713)}
						\gppoint{gp mark 1}{(6.063,2.713)}
						\gppoint{gp mark 1}{(6.097,2.372)}
						\gppoint{gp mark 1}{(6.131,2.485)}
						\gppoint{gp mark 1}{(6.166,2.372)}
						\gppoint{gp mark 1}{(6.200,2.372)}
						\gppoint{gp mark 1}{(6.234,2.599)}
						\gppoint{gp mark 1}{(6.268,2.485)}
						\gppoint{gp mark 1}{(6.302,2.599)}
						\gppoint{gp mark 1}{(6.337,2.713)}
						\gppoint{gp mark 1}{(6.371,2.485)}
						\gppoint{gp mark 1}{(6.405,2.599)}
						\gppoint{gp mark 1}{(6.439,2.599)}
						\gppoint{gp mark 1}{(6.473,2.485)}
						\gppoint{gp mark 1}{(6.508,2.713)}
						\gppoint{gp mark 1}{(6.542,2.599)}
						\gppoint{gp mark 1}{(6.576,2.599)}
						\gppoint{gp mark 1}{(6.610,2.713)}
						\gppoint{gp mark 1}{(6.644,2.827)}
						\gppoint{gp mark 1}{(6.679,2.713)}
						\gppoint{gp mark 1}{(6.713,2.599)}
						\gppoint{gp mark 1}{(6.747,2.372)}
						\gppoint{gp mark 1}{(6.781,2.599)}
						\gppoint{gp mark 1}{(6.816,2.485)}
						\gppoint{gp mark 1}{(6.884,2.599)}
						\gppoint{gp mark 1}{(6.918,2.485)}
						\gppoint{gp mark 1}{(6.952,2.372)}
						\gppoint{gp mark 1}{(6.987,2.485)}
						\gppoint{gp mark 1}{(7.021,2.485)}
						\gppoint{gp mark 1}{(7.055,2.372)}
						\gppoint{gp mark 1}{(7.089,2.485)}
						\gppoint{gp mark 1}{(7.123,2.485)}
						\gppoint{gp mark 1}{(7.226,2.144)}
						\gppoint{gp mark 1}{(7.397,1.688)}
						\gppoint{gp mark 1}{(7.431,1.576)}
						\gppoint{gp mark 1}{(7.465,1.576)}
						\gppoint{gp mark 1}{(7.500,1.576)}
						\gppoint{gp mark 1}{(7.534,1.576)}
						\gppoint{gp mark 1}{(7.568,1.576)}
						\gppoint{gp mark 1}{(7.602,1.576)}
						\gppoint{gp mark 1}{(7.637,1.802)}
						\gppoint{gp mark 1}{(7.671,1.688)}
						\gppoint{gp mark 1}{(7.705,1.802)}
						\gppoint{gp mark 1}{(7.739,1.802)}
						\gppoint{gp mark 1}{(7.773,1.916)}
						\gppoint{gp mark 1}{(7.808,2.258)}
						\gppoint{gp mark 1}{(7.842,2.258)}
						\gppoint{gp mark 1}{(7.876,2.485)}
						\gppoint{gp mark 1}{(7.910,2.599)}
						\gppoint{gp mark 1}{(7.944,2.485)}
						\gppoint{gp mark 1}{(7.979,2.485)}
						\gppoint{gp mark 1}{(8.013,2.485)}
						\gppoint{gp mark 1}{(8.047,2.485)}
						\gppoint{gp mark 1}{(8.081,2.372)}
						\gppoint{gp mark 1}{(8.115,2.258)}
						\gppoint{gp mark 1}{(8.150,2.258)}
						\gppoint{gp mark 1}{(8.184,2.258)}
						\gppoint{gp mark 1}{(8.218,2.372)}
						\gppoint{gp mark 1}{(8.252,2.485)}
						\gppoint{gp mark 1}{(8.287,2.372)}
						\gppoint{gp mark 1}{(8.321,2.372)}
						\gppoint{gp mark 1}{(8.355,2.372)}
						\gppoint{gp mark 1}{(8.389,2.485)}
						\gppoint{gp mark 1}{(8.423,2.372)}
						\gppoint{gp mark 1}{(8.458,2.372)}
						\gppoint{gp mark 1}{(8.492,2.372)}
						\gppoint{gp mark 1}{(8.526,2.372)}
						\gppoint{gp mark 1}{(8.560,2.372)}
						\gppoint{gp mark 1}{(8.594,2.258)}
						\gppoint{gp mark 1}{(8.629,2.372)}
						\gppoint{gp mark 1}{(8.663,2.485)}
						\gppoint{gp mark 1}{(8.697,2.372)}
						\gppoint{gp mark 1}{(8.731,2.372)}
						\gppoint{gp mark 1}{(8.765,2.372)}
						\gppoint{gp mark 1}{(8.800,2.372)}
						\gppoint{gp mark 1}{(8.834,2.485)}
						\gppoint{gp mark 1}{(8.868,2.372)}
						\gppoint{gp mark 1}{(8.902,2.372)}
						\gppoint{gp mark 1}{(8.937,2.372)}
						\gppoint{gp mark 1}{(8.971,2.372)}
						\gppoint{gp mark 1}{(9.005,2.258)}
						\gppoint{gp mark 1}{(9.039,2.144)}
						\gppoint{gp mark 1}{(9.073,2.258)}
						\gppoint{gp mark 1}{(9.108,2.258)}
						\gppoint{gp mark 1}{(9.142,2.372)}
						\gppoint{gp mark 1}{(9.176,2.258)}
						\gppoint{gp mark 1}{(9.210,2.258)}
						\gppoint{gp mark 1}{(9.244,2.372)}
						\gppoint{gp mark 1}{(9.279,2.258)}
						\gppoint{gp mark 1}{(9.313,2.485)}
						\gppoint{gp mark 1}{(9.347,2.372)}
						\gppoint{gp mark 1}{(9.381,2.258)}
						\gppoint{gp mark 1}{(9.415,2.258)}
						\gppoint{gp mark 1}{(9.450,2.144)}
						\gppoint{gp mark 1}{(9.484,2.030)}
						\gppoint{gp mark 1}{(9.518,2.258)}
						\gppoint{gp mark 1}{(9.552,2.258)}
						\gppoint{gp mark 1}{(9.587,2.372)}
						\gppoint{gp mark 1}{(9.621,2.372)}
						\gppoint{gp mark 1}{(9.655,2.258)}
						\gppoint{gp mark 1}{(9.689,2.144)}
						\gppoint{gp mark 1}{(9.723,2.144)}
						\gppoint{gp mark 1}{(9.758,2.372)}
						\gppoint{gp mark 1}{(9.792,2.258)}
						\gppoint{gp mark 1}{(9.826,2.144)}
						\gppoint{gp mark 1}{(9.860,2.258)}
						\gppoint{gp mark 1}{(9.894,2.258)}
						\gppoint{gp mark 1}{(9.929,2.144)}
						\gppoint{gp mark 1}{(9.963,2.258)}
						\gppoint{gp mark 1}{(9.997,2.258)}
						\gppoint{gp mark 1}{(10.031,2.030)}
						\gppoint{gp mark 1}{(10.065,2.144)}
						\gppoint{gp mark 1}{(10.100,2.144)}
						\gppoint{gp mark 1}{(10.134,2.372)}
						\gppoint{gp mark 1}{(10.168,2.258)}
						\gppoint{gp mark 1}{(10.202,2.144)}
						\gppoint{gp mark 1}{(10.237,2.258)}
						\gppoint{gp mark 1}{(10.271,2.144)}
						\gppoint{gp mark 1}{(10.305,2.144)}
						\gppoint{gp mark 1}{(10.339,2.372)}
						\gppoint{gp mark 1}{(10.373,2.258)}
						\gppoint{gp mark 1}{(10.408,2.144)}
						\gppoint{gp mark 1}{(10.442,2.258)}
						\gppoint{gp mark 1}{(10.476,2.144)}
						\gppoint{gp mark 1}{(10.510,2.258)}
						\gppoint{gp mark 1}{(10.544,2.258)}
						\gppoint{gp mark 1}{(10.579,2.258)}
						\gppoint{gp mark 1}{(10.613,1.916)}
						\gppoint{gp mark 1}{(10.647,2.372)}
						\gppoint{gp mark 1}{(10.681,2.144)}
						\gppoint{gp mark 1}{(10.715,2.144)}
						\gppoint{gp mark 1}{(10.750,2.372)}
						\gppoint{gp mark 1}{(10.784,2.258)}
						\gppoint{gp mark 1}{(10.818,2.258)}
						\gppoint{gp mark 1}{(10.852,2.258)}
						\gppoint{gp mark 1}{(10.886,2.372)}
						\gppoint{gp mark 1}{(10.921,2.372)}
						\gppoint{gp mark 1}{(10.955,2.144)}
						\gppoint{gp mark 1}{(10.989,2.144)}
						\gppoint{gp mark 1}{(11.023,2.258)}
						\gppoint{gp mark 1}{(11.058,2.258)}
						\gppoint{gp mark 1}{(11.092,2.144)}
						\gppoint{gp mark 1}{(11.126,2.144)}
						\gppoint{gp mark 1}{(11.160,2.485)}
						\gppoint{gp mark 1}{(11.194,2.144)}
						\gppoint{gp mark 1}{(11.229,2.144)}
						\gppoint{gp mark 1}{(11.263,2.144)}
						\gppoint{gp mark 1}{(11.297,2.258)}
						\gppoint{gp mark 1}{(11.331,2.258)}
						\gppoint{gp mark 1}{(11.365,2.258)}
						\gppoint{gp mark 1}{(11.400,2.144)}
						\gppoint{gp mark 1}{(11.434,2.258)}
						\gppoint{gp mark 1}{(11.468,2.144)}
						\gpsetpointsize{2.80}
						\gpcolor{rgb color={0.000,0.000,1.000}}
						\gppoint{gp mark 7}{(1.718,6.813)}
						\gppoint{gp mark 7}{(1.752,4.878)}
						\gppoint{gp mark 7}{(1.787,6.017)}
						\gppoint{gp mark 7}{(1.821,5.334)}
						\gppoint{gp mark 7}{(1.855,4.194)}
						\gppoint{gp mark 7}{(1.992,3.283)}
						\gppoint{gp mark 7}{(2.095,2.827)}
						\gppoint{gp mark 7}{(2.163,3.169)}
						\gppoint{gp mark 7}{(2.197,2.372)}
						\gppoint{gp mark 7}{(2.231,2.372)}
						\gppoint{gp mark 7}{(2.266,2.030)}
						\gppoint{gp mark 7}{(2.300,2.258)}
						\gppoint{gp mark 7}{(2.334,2.258)}
						\gppoint{gp mark 7}{(2.368,2.258)}
						\gppoint{gp mark 7}{(2.642,2.030)}
						\gppoint{gp mark 7}{(2.710,1.802)}
						\gppoint{gp mark 7}{(2.779,1.802)}
						\gppoint{gp mark 7}{(2.847,1.916)}
						\gppoint{gp mark 7}{(2.950,2.030)}
						\gppoint{gp mark 7}{(2.984,2.144)}
						\gppoint{gp mark 7}{(3.121,2.485)}
						\gppoint{gp mark 7}{(3.292,2.599)}
						\gppoint{gp mark 7}{(3.360,2.258)}
						\gppoint{gp mark 7}{(3.429,2.030)}
						\gppoint{gp mark 7}{(3.463,2.030)}
						\gppoint{gp mark 7}{(3.702,2.030)}
						\gppoint{gp mark 7}{(3.839,1.916)}
						\gppoint{gp mark 7}{(4.044,2.258)}
						\gppoint{gp mark 7}{(4.079,2.144)}
						\gppoint{gp mark 7}{(4.250,2.258)}
						\gppoint{gp mark 7}{(4.729,2.372)}
						\gppoint{gp mark 7}{(4.866,2.258)}
						\gppoint{gp mark 7}{(5.242,2.258)}
						\gppoint{gp mark 7}{(5.926,2.485)}
						\gppoint{gp mark 7}{(6.850,2.713)}
						\gppoint{gp mark 7}{(7.158,2.485)}
						\gppoint{gp mark 7}{(7.192,2.144)}
						\gppoint{gp mark 7}{(7.260,2.030)}
						\gppoint{gp mark 7}{(7.294,1.688)}
						\gppoint{gp mark 7}{(7.329,1.688)}
						\gppoint{gp mark 7}{(7.363,1.802)}
						\gpcolor{rgb color={1.000,1.000,0.000}}
						\gpsetpointsize{5.40}
						\gppoint{gp mark 5}{(2.779,1.802)}
						\gpcolor{rgb color={0.000,0.000,0.000}}
						\gpsetpointsize{5.80}
						\gppoint{gp mark 4}{(2.779,1.802)}
						\gpcolor{color=gp lt color border}
						\draw[gp path] (1.684,7.702)--(1.684,1.165)--(11.947,1.165)--(11.947,7.702)--cycle;
						%% coordinates of the plot area
						\gpdefrectangularnode{gp plot 1}{\pgfpoint{1.684cm}{1.165cm}}{\pgfpoint{11.947cm}{7.702cm}}
						\end{tikzpicture}
						\subcaption{Testing set of KP}					
						\label{fig:TestF22}
					\end{center}				
				\end{minipage}
			\end{center}
			\caption{An illustration of the performance of the proposed approach for the best subset selection of features on the reduced testing set} 
			\label{fig:TestFrontier2}
		\end{figure}
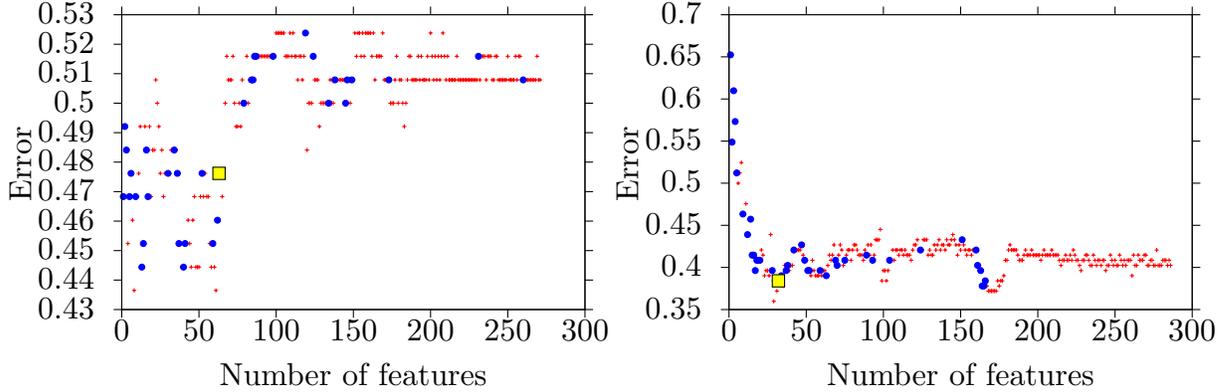

		Similar to Section~\ref{subsec:CompTraingTesting}, for each of the points (other than the ideal point)  in Figure~\ref{fig:TrainFrontier2}, we have plotted its corresponding point for the testing set in Figure~\ref{fig:TestFrontier2}. We again observe that the selected model, i.e., the (yellow) square, is nearly optimal for both classes of optimization problems.  In fact the proposed approach has selected a prediction model with the accuracy of around 52\% and 62\% for AP and KP instances, respectively. This implies that the absolute difference between the accuracy of the model selected by the proposed approach  and the accuracy of the optimal model is almost 5\% and 3\% for AP and KP instances, respectively.

		\begin{table}[ht]
			\caption{Accuracy and average time decrease of testing set when using the proposed ML technique (for the case of the reduced training and testing sets)}
			\centering \footnotesize
			%\resizebox{\textwidth}{!}{ 
			\begin{tabular}{ccclcccccc}
				\cline{7-8}
				\multicolumn{1}{l}{} & \multicolumn{1}{l}{} & \multicolumn{1}{l}{} &  &  & \multicolumn{1}{l|}{} & \multicolumn{2}{c|}{\textbf{Time Decrease}} & \multicolumn{1}{l}{} & \multicolumn{1}{l}{} \\ \cline{1-1} \cline{3-3} \cline{5-5} \cline{7-8} \cline{10-10} 
				\multicolumn{1}{|c|}{\textbf{Type}} & \multicolumn{1}{c|}{} & \multicolumn{1}{c|}{\textbf{Vars}} & \multicolumn{1}{l|}{} & \multicolumn{1}{c|}{\textbf{Accuracy}} & \multicolumn{1}{c|}{} & \multicolumn{1}{c|}{\textbf{ML vs.  Rand}} & \multicolumn{1}{c|}{\textbf{Best vs. Rand}} & \multicolumn{1}{c|}{} & \multicolumn{1}{c|}{\textbf{\makecell[c]{$\bf  \frac{\mbox{ML vs. Rand}}{\mbox{Best vs. Rand}}$}}} \\ \cline{1-1} \cline{3-3} \cline{5-5} \cline{7-8} \cline{10-10} 
				\multicolumn{1}{|c|}{\multirow{5}{*}{AP}} & \multicolumn{1}{c|}{} & \multicolumn{1}{c|}{$20 \times 20$} & \multicolumn{1}{l|}{} & \multicolumn{1}{c|}{58.33\%} & \multicolumn{1}{c|}{} & \multicolumn{1}{c|}{0.82\%} & \multicolumn{1}{c|}{2.14\%} & \multicolumn{1}{c|}{} & \multicolumn{1}{c|}{38.41\%} \\
				\multicolumn{1}{|c|}{} & \multicolumn{1}{c|}{} & \multicolumn{1}{c|}{$25 \times 25$} & \multicolumn{1}{l|}{} & \multicolumn{1}{c|}{52.38\%} & \multicolumn{1}{c|}{} & \multicolumn{1}{c|}{1.17\%} & \multicolumn{1}{c|}{2.39\%} & \multicolumn{1}{c|}{} & \multicolumn{1}{c|}{48.76\%} \\
				\multicolumn{1}{|c|}{} & \multicolumn{1}{c|}{} & \multicolumn{1}{c|}{$30 \times 30$} & \multicolumn{1}{l|}{} & \multicolumn{1}{c|}{55.17\%} & \multicolumn{1}{c|}{} & \multicolumn{1}{c|}{0.67\%} & \multicolumn{1}{c|}{1.93\%} & \multicolumn{1}{c|}{} & \multicolumn{1}{c|}{34.46\%} \\
				\multicolumn{1}{|c|}{} & \multicolumn{1}{c|}{} & \multicolumn{1}{c|}{$35 \times 35$} & \multicolumn{1}{l|}{} & \multicolumn{1}{c|}{45.45\%} & \multicolumn{1}{c|}{} & \multicolumn{1}{c|}{1.10\%} & \multicolumn{1}{c|}{2.53\%} & \multicolumn{1}{c|}{} & \multicolumn{1}{c|}{43.35\%} \\
				\multicolumn{1}{|c|}{} & \multicolumn{1}{c|}{} & \multicolumn{1}{c|}{$40 \times 40$} & \multicolumn{1}{l|}{} & \multicolumn{1}{c|}{52.63\%} & \multicolumn{1}{c|}{} & \multicolumn{1}{c|}{1.60\%} & \multicolumn{1}{c|}{2.87\%} & \multicolumn{1}{c|}{} & \multicolumn{1}{c|}{55.61\%} \\ \cline{1-1} \cline{3-3} \cline{5-5} \cline{7-8} \cline{10-10} 
				\multicolumn{1}{|c|}{\textbf{Avg}} &  &  & \multicolumn{1}{l|}{} & \multicolumn{1}{c|}{\textbf{52.38\%}} & \multicolumn{1}{c|}{} & \multicolumn{1}{c|}{\textbf{1.03\%}} & \multicolumn{1}{c|}{\textbf{2.35\%}} & \multicolumn{1}{c|}{} & \multicolumn{1}{c|}{\textbf{43.99\%}} \\ \cline{1-1} \cline{5-5} \cline{7-8} \cline{10-10} 
				&  &  &  &  &  &  &  &  &  \\ \cline{1-1} \cline{3-3} \cline{5-5} \cline{7-8} \cline{10-10} 
				\multicolumn{1}{|c|}{\multirow{5}{*}{KP}} & \multicolumn{1}{c|}{} & \multicolumn{1}{c|}{60} & \multicolumn{1}{l|}{} & \multicolumn{1}{c|}{60.61\%} & \multicolumn{1}{c|}{} & \multicolumn{1}{c|}{6.75\%} & \multicolumn{1}{c|}{12.16\%} & \multicolumn{1}{c|}{} & \multicolumn{1}{c|}{55.51\%} \\
				\multicolumn{1}{|c|}{} & \multicolumn{1}{c|}{} & \multicolumn{1}{c|}{70} & \multicolumn{1}{l|}{} & \multicolumn{1}{c|}{55.56\%} & \multicolumn{1}{c|}{} & \multicolumn{1}{c|}{6.79\%} & \multicolumn{1}{c|}{12.03\%} & \multicolumn{1}{c|}{} & \multicolumn{1}{c|}{56.43\%} \\
				\multicolumn{1}{|c|}{} & \multicolumn{1}{c|}{} & \multicolumn{1}{c|}{80} & \multicolumn{1}{l|}{} & \multicolumn{1}{c|}{60.00\%} & \multicolumn{1}{c|}{} & \multicolumn{1}{c|}{12.14\%} & \multicolumn{1}{c|}{16.91\%} & \multicolumn{1}{c|}{} & \multicolumn{1}{c|}{71.79\%} \\
				\multicolumn{1}{|c|}{} & \multicolumn{1}{c|}{} & \multicolumn{1}{c|}{90} & \multicolumn{1}{l|}{} & \multicolumn{1}{c|}{72.73\%} & \multicolumn{1}{c|}{} & \multicolumn{1}{c|}{9.22\%} & \multicolumn{1}{c|}{12.26\%} & \multicolumn{1}{c|}{} & \multicolumn{1}{c|}{75.17\%} \\
				\multicolumn{1}{|c|}{} & \multicolumn{1}{c|}{} & \multicolumn{1}{c|}{100} & \multicolumn{1}{l|}{} & \multicolumn{1}{c|}{62.50\%} & \multicolumn{1}{c|}{} & \multicolumn{1}{c|}{4.70\%} & \multicolumn{1}{c|}{10.56\%} & \multicolumn{1}{c|}{} & \multicolumn{1}{c|}{44.50\%} \\ \cline{1-1} \cline{3-3} \cline{5-5} \cline{7-8} \cline{10-10} 
				\multicolumn{1}{|c|}{\textbf{Avg}} &  &  & \multicolumn{1}{l|}{} & \multicolumn{1}{c|}{\textbf{61.59\%}} & \multicolumn{1}{c|}{} & \multicolumn{1}{c|}{\textbf{7.61\%}} & \multicolumn{1}{c|}{\textbf{12.64\%}} & \multicolumn{1}{c|}{} & \multicolumn{1}{c|}{\textbf{60.18\%}} \\ \cline{1-1} \cline{5-5} \cline{7-8} \cline{10-10} 
			\end{tabular}%}
			\label{tab:ExpTable3}
		\end{table}
		
		A summary of the results of this last experiment can be found in Table \ref{tab:ExpTable3}. Observe that the average prediction accuracy on the testing set for the experimental setting in this section, i.e., reduced training and testing sets, has improved significantly for KP instances compared to the results given in Sections~\ref{subsec:CompTraingTesting} and \ref{subsec:ComplTrainRedTes}. However for the instances of AP problem, the average prediction accuracy in this section is only better than the one presented in Section~\ref{subsec:CompTraingTesting}. Overall, the average prediction accuracy is $52.38\%$ and $61.59\%$ for AP and KP instances when using reduced training and testing sets. By considering the tie cases as success events, the projected accuracy increases up to $70\%$ and $68.5\%$ for AP and KP instances, respectively. The importance of such an increase in the accuracy is highlighted by the time decrease percentages given in Table \ref{tab:ExpTable3}, which is over 1\% for AP instances and is near 8\% for KP instances. In fact, for the largest subclass of AP instances, the average time improvement of 1.6\%  is equivalent to almost 110 seconds on average. Similarly, for the largest subclass of KP instances, the time improvement of 4.7\% is around 490 seconds on average.
		
		\subsection{Replacing MSVM by Random Forest}
		\label{subsec: RandomForest}
		One main reason that we used MSVM in this study is that (as shown in the previous sections) it performs well in practice for the purpose of this study. However, another critical reason is the fact that MSVM creates a matrix of parameters denoted by $W^t$ in each iteration. This matrix has $p$ rows where $p$ is the number of objective functions. In other words, for each objective function, MSVM creates a specific model for predicting which one should be used for projection in KSA. This characteristic is desirable because it allowed us to develop a custom-built bi-objective heuristic for selecting the best subset of features. Specifically, as discussed in Section~\ref{sec:bestfeatures}, this characteristic is essential for identifying the least important feature in each iteration of Algorithm~\ref{alg:Alg1}. However, applying such a procedure on other ML techniques is not trivial.   
		
		In light of the above, in this section we replace MSVM by Random Forest within the proposed machine learning framework. However, we simply use the best subset of features selected by MSVM and then feed it to Random Forest for training and predicting. To implement Random Forest we use \texttt{scikit-learn} library in Python \citep{scikit-learn}. Table \ref{tab:ExpTable4} shows a comparison between the prediction accuracy of MSVM and Random Forest under three experimental settings described in Sections~\ref{subsec:CompTraingTesting}-\ref{subsec:RedTrainTes}. In other words, Setting 1 corresponds to the complete training and testing sets; Setting 2 corresponds to the complete training set and reduced testing sets; Finally, Setting 3 refers to the reduced training and testing sets. 
		
		In this table, columns labeled `Increase' show the average percentage of increase in the prediction accuracy of the Random Forest compared to MSVM. Observe from these columns that the reported numbers are mostly negative. This implies that, in general, MSVM outperforms Random Forest in terms of prediction accuracy. For example, in Setting 3, we observe that the accuracy of Random Forest is around 18.63\% and 22.88\% worse than the accuracy of MSVM on average. This experiment clearly shows the advantage of using MSVM in the proposed ML framework. 
		
		\begin{table}[ht]
			\caption{A performance comparison between MSVM and Random Forest on a testing set}
			\centering \footnotesize
			\resizebox{\textwidth}{!}{ 
				\begin{tabular}{cccccccccccc}
					\cline{5-6} \cline{8-9} \cline{11-12}
					&  &  & \multicolumn{1}{c|}{} & \multicolumn{2}{c|}{\textbf{Setting 1}} & \multicolumn{1}{c|}{} & \multicolumn{2}{c|}{\textbf{Setting 2}} & \multicolumn{1}{c|}{} & \multicolumn{2}{c|}{\textbf{Setting 3}} \\ \cline{1-1} \cline{3-3} \cline{5-6} \cline{8-9} \cline{11-12} 
					\multicolumn{1}{|c|}{\textbf{Type}} & \multicolumn{1}{c|}{} & \multicolumn{1}{c|}{\textbf{Vars}} & \multicolumn{1}{c|}{} & \multicolumn{1}{c|}{\textbf{Accuracy}} & \multicolumn{1}{c|}{\textbf{Increase}} & \multicolumn{1}{c|}{} & \multicolumn{1}{c|}{\textbf{Accuracy}} & \multicolumn{1}{c|}{\textbf{Increase}} & \multicolumn{1}{c|}{} & \multicolumn{1}{c|}{\textbf{Accuracy}} & \multicolumn{1}{c|}{\textbf{ Increase}} \\ \cline{1-1} \cline{3-3} \cline{5-6} \cline{8-9} \cline{11-12} 
					\multicolumn{1}{|c|}{\multirow{5}{*}{AP}} & \multicolumn{1}{c|}{} & \multicolumn{1}{c|}{$20\times20$} & \multicolumn{1}{c|}{} & \multicolumn{1}{c|}{61.11\%} & \multicolumn{1}{c|}{9.99\%} & \multicolumn{1}{c|}{} & \multicolumn{1}{c|}{64.52\%} & \multicolumn{1}{c|}{11.12\%} & \multicolumn{1}{c|}{} & \multicolumn{1}{c|}{36.36\%} & \multicolumn{1}{c|}{-37.66\%} \\
					\multicolumn{1}{|c|}{} & \multicolumn{1}{c|}{} & \multicolumn{1}{c|}{$25\times25$} & \multicolumn{1}{c|}{} & \multicolumn{1}{c|}{42.11\%} & \multicolumn{1}{c|}{-5.89\%} & \multicolumn{1}{c|}{} & \multicolumn{1}{c|}{50.00\%} & \multicolumn{1}{c|}{-16.67\%} & \multicolumn{1}{c|}{} & \multicolumn{1}{c|}{26.67\%} & \multicolumn{1}{c|}{-49.09\%} \\
					\multicolumn{1}{|c|}{} & \multicolumn{1}{c|}{} & \multicolumn{1}{c|}{$30\times30$} & \multicolumn{1}{c|}{} & \multicolumn{1}{c|}{45.65\%} & \multicolumn{1}{c|}{10.54\%} & \multicolumn{1}{c|}{} & \multicolumn{1}{c|}{37.04\%} & \multicolumn{1}{c|}{-9.09\%} & \multicolumn{1}{c|}{} & \multicolumn{1}{c|}{52.63\%} & \multicolumn{1}{c|}{-4.60\%} \\
					\multicolumn{1}{|c|}{} & \multicolumn{1}{c|}{} & \multicolumn{1}{c|}{$35\times35$} & \multicolumn{1}{c|}{} & \multicolumn{1}{c|}{62.50\%} & \multicolumn{1}{c|}{20.01\%} & \multicolumn{1}{c|}{} & \multicolumn{1}{c|}{65.79\%} & \multicolumn{1}{c|}{13.65\%} & \multicolumn{1}{c|}{} & \multicolumn{1}{c|}{70.59\%} & \multicolumn{1}{c|}{55.31\%} \\
					\multicolumn{1}{|c|}{} & \multicolumn{1}{c|}{} & \multicolumn{1}{c|}{$40\times40$} & \multicolumn{1}{c|}{} & \multicolumn{1}{c|}{50.00\%} & \multicolumn{1}{c|}{-11.11\%} & \multicolumn{1}{c|}{} & \multicolumn{1}{c|}{73.91\%} & \multicolumn{1}{c|}{6.24\%} & \multicolumn{1}{c|}{} & \multicolumn{1}{c|}{50.00\%} & \multicolumn{1}{c|}{-5.00\%} \\ \cline{1-1} \cline{3-3} \cline{5-6} \cline{8-9} \cline{11-12} 
					\multicolumn{1}{|c|}{\textbf{Avg}} &  &  & \multicolumn{1}{c|}{} & \multicolumn{1}{c|}{\textbf{52.50\%}} & \multicolumn{1}{c|}{\textbf{6.06\%}} & \multicolumn{1}{c|}{} & \multicolumn{1}{c|}{\textbf{59.69\%}} & \multicolumn{1}{c|}{\textbf{5.48\%}} & \multicolumn{1}{c|}{} & \multicolumn{1}{c|}{\textbf{42.62\%}} & \multicolumn{1}{c|}{\textbf{-18.63\%}} \\ \cline{1-1} \cline{5-6} \cline{8-9} \cline{11-12} 
					&  &  &  &  &  &  &  &  &  &  &  \\ \cline{1-1} \cline{3-3} \cline{5-6} \cline{8-9} \cline{11-12} 
					\multicolumn{1}{|c|}{\multirow{5}{*}{KP}} & \multicolumn{1}{c|}{} & \multicolumn{1}{c|}{60} & \multicolumn{1}{c|}{} & \multicolumn{1}{c|}{47.73\%} & \multicolumn{1}{c|}{-25.00\%} & \multicolumn{1}{c|}{} & \multicolumn{1}{c|}{47.37\%} & \multicolumn{1}{c|}{-33.33\%} & \multicolumn{1}{c|}{} & \multicolumn{1}{c|}{41.18\%} & \multicolumn{1}{c|}{-32.06\%} \\
					\multicolumn{1}{|c|}{} & \multicolumn{1}{c|}{} & \multicolumn{1}{c|}{70} & \multicolumn{1}{c|}{} & \multicolumn{1}{c|}{50.00\%} & \multicolumn{1}{c|}{9.53\%} & \multicolumn{1}{c|}{} & \multicolumn{1}{c|}{56.76\%} & \multicolumn{1}{c|}{23.52\%} & \multicolumn{1}{c|}{} & \multicolumn{1}{c|}{50.00\%} & \multicolumn{1}{c|}{-10.01\%} \\
					\multicolumn{1}{|c|}{} & \multicolumn{1}{c|}{} & \multicolumn{1}{c|}{80} & \multicolumn{1}{c|}{} & \multicolumn{1}{c|}{48.78\%} & \multicolumn{1}{c|}{-13.06\%} & \multicolumn{1}{c|}{} & \multicolumn{1}{c|}{54.55\%} & \multicolumn{1}{c|}{-5.27\%} & \multicolumn{1}{c|}{} & \multicolumn{1}{c|}{61.90\%} & \multicolumn{1}{c|}{3.17\%} \\
					\multicolumn{1}{|c|}{} & \multicolumn{1}{c|}{} & \multicolumn{1}{c|}{90} & \multicolumn{1}{c|}{} & \multicolumn{1}{c|}{44.12\%} & \multicolumn{1}{c|}{-25.00\%} & \multicolumn{1}{c|}{} & \multicolumn{1}{c|}{55.56\%} & \multicolumn{1}{c|}{-11.76\%} & \multicolumn{1}{c|}{} & \multicolumn{1}{c|}{47.06\%} & \multicolumn{1}{c|}{-35.30\%} \\
					\multicolumn{1}{|c|}{} & \multicolumn{1}{c|}{} & \multicolumn{1}{c|}{100} & \multicolumn{1}{c|}{} & \multicolumn{1}{c|}{37.14\%} & \multicolumn{1}{c|}{-27.78\%} & \multicolumn{1}{c|}{} & \multicolumn{1}{c|}{43.33\%} & \multicolumn{1}{c|}{-27.78\%} & \multicolumn{1}{c|}{} & \multicolumn{1}{c|}{45.95\%} & \multicolumn{1}{c|}{-26.49\%} \\ \cline{1-1} \cline{3-3} \cline{5-6} \cline{8-9} \cline{11-12} 
					\multicolumn{1}{|c|}{\textbf{Avg}} &  &  & \multicolumn{1}{c|}{} & \multicolumn{1}{c|}{\textbf{46.00\%}} & \multicolumn{1}{c|}{\textbf{-16.36\%}} & \multicolumn{1}{c|}{} & \multicolumn{1}{c|}{\textbf{51.52\%}} & \multicolumn{1}{c|}{\textbf{-13.26\%}} & \multicolumn{1}{c|}{} & \multicolumn{1}{c|}{\textbf{47.50\%}} & \multicolumn{1}{c|}{\textbf{-22.88\%}} \\ \cline{1-1} \cline{5-6} \cline{8-9} \cline{11-12} 
				\end{tabular}}
				\label{tab:ExpTable4}
			\end{table}
			
			\section{Conclusions and future research}
			\label{sec:Conclusions}
			We presented a multi-class support vector machine based approach to enhance exact multi-objective binary linear programming algorithms. Our approach simulates the best selection of objective function to be used for projection in the KSA in order to improve its computational time. We introduced a pre-ordering approach for the objective functions in the input file for the purpose of standardizing the vector of features. Moreover, we introduced a bi-objective optimization approach for selecting the best subset of features in order to overcome overfitting. By conducting an extensive computational , we showed that reaching to the prediction accuracy of around 70\% is possible for instances of tri-objective AP and KP.  It was shown that such a prediction accuracy results in a decrease of over 12\% in the computational time for some instances.   
			
			Overall, we hope that the simplicity of our proposed ML technique and its promising results encourage more researchers to use ML techniques for improving multi-objective optimization solvers. Note that, in this paper, we studied the problem of learning to project in a static setting, i.e., before solving an instance we predict the best objective function and use it during the course of the search. So, one future research direction of this study would be finding a way to employ the proposed learning-to-project technique in a dynamic setting, i.e., at each iteration in the search process we predict the best projected objective and use it. Evidently, this may result in developing new algorithms that have not yet been studied in the literature of multi-objective optimization.
			
\bibliographystyle{informs2014} 
\bibliography{references.bib}   

\newpage
\begin{APPENDICES}
			\section{List of Features}
			\label{app:Features}
			In this section, we present all $5p^2+106p-50$ features that we used in this study. Note that since $p=3$ in our computational study, the total number of features are 313 features. For convenience, we partition all the proposed features into some subsets and present them next. We use the letter $F$ to represent each subset.     
			
			The first subset of features is $F^1=\{\boldsymbol{c}_1^\intercal\boldsymbol{\tilde{x}},\boldsymbol{c}_2^\intercal\boldsymbol{\tilde{x}},\dots,\boldsymbol{c}_p^\intercal\boldsymbol{\tilde{x}}\}$, which will be automatically computed during the pre-ordering process. To incorporate the impact of the size of an instance in the learning process, we  introduce  $F^2=\{n\}$, $F^3=\{m\}$, and $F^4=\{\mbox{density}(A)\}$. To incorporate the impact of zero coefficients of the objective functions in the learning process, for each $i\in\{1,\dots,p\}$ we introduce:
			$$
			F_{i}^{5}=\{\mbox{size}(S^5_i)\},
			$$
			where $S^5_i=\{c\in\boldsymbol{c}_{i}:c=0\}$. To incorporate the impact of positive coefficients of the objective functions in the learning process, for each $i\in\{1,\dots,p\}$ we introduce: 
			$$
			F_i^{6}=\{\mbox{size}(S^6_i),\mbox{Avg}(S^6_i),\mbox{Min}(S^6_i),\mbox{Max}(S^6_i),\mbox{Std}(S^6_i),\mbox{Median}(S^6_i)\},
			$$
			where $S^6_i=\{c\in\boldsymbol{c}_{i}:c>0\}$. To incorporate the impact of negative coefficients of the objective functions in the learning process, for each $i\in\{1,\dots,p\}$ we introduce: 
			$$
			F_i^{7}=\{\mbox{size}(S^7_i),\mbox{Avg}(S^7_i),\mbox{Min}(S^7_i),\mbox{Max}(S^7_i),\mbox{Std}(S^7_i),\mbox{Median}(S^7_i)\},
			$$
			where $S^7_i=\{c\in\boldsymbol{c}_{i}:c<0\}$. To establish a relation between the objective functions and $A$ in the learning process, for each $i\in\{1,\dots,p\}$ we introduce:
			
			$$
			F_i^{8}=\{\mbox{Avg}(S^8_i),\mbox{Min}(S^8_i),\mbox{Max}(S^8_i),\mbox{Std}(S^8_i),\mbox{Median}(S^8_i)\},
			$$
			
			where $S^8_i=\cup_{j\in\{1,\dots,m\}}\{\boldsymbol{c}^\intercal_i \boldsymbol{a}^\intercal_{j.}\}$ and $\boldsymbol{a}_{j.}$ represents row $j$ of matrix $A$. For the same purpose, for each $i\in\{1,\dots,p\}$ we also introduce: 
			$$F^{9}_{i}=\boldsymbol{c}^\intercal_i\times A^\intercal\times\boldsymbol{b}.$$
			
			For each $j\in \{1,\dots,m\}$, let $b'_j:=b_j+1$ if $b_j \ge 0$ and $b'_j:=b_j-1$ otherwise. To establish a relation between the positive and negative coefficients in the objective functions and $\boldsymbol{b}$ in the learning process, for each $i\in\{1,\dots,p\}$ and $k\in\{10,11\}$ we introduce:
			$$
			F_i^{k}=\{\mbox{Avg}(S^{k}_i),\mbox{Min}(S^{k}_i),\mbox{Max}(S^{k}_i),\mbox{Std}(S^{k}_i),\mbox{Median}(S^{k}_i)\},
			$$
			where 
			$S^{10}_i=\cup_{j\in\{1,\dots,m\}}\{\frac{\sum_{c\in\boldsymbol{S}_{i}^{6}}{c}}{b'_j}\}$ and $S^{11}_i=\cup_{j\in\{1,\dots,m\}}\{\frac{\sum_{c\in\boldsymbol{S}_{i}^{7}}{c}}{b'_j}\}$.
			
			For each $i\in \{1,\dots,p\}$, let $l_{i}:=\min_{\boldsymbol{x} \in \mathcal{X}_{LR}} \ z_i(\boldsymbol{x})$ and  $u_{i}:=\max_{\boldsymbol{x} \in \mathcal{X}_{LR}}  \ z_i(\boldsymbol{x})$ where $\mathcal{X}_{LR}$ is the linear programming relaxation of $\mathcal{X}$. To incorporate the impact of the volume of the search region in a projected criterion space, i.e., a $(p-1)$-dimensional criterion space, in the learning process, for each $i\in\{1,\dots,p\}$ we introduce:
			\begin{equation*}
			F_i^{12}=\prod\limits_{j\in\{1,\dots,p\}\setminus \{i\}}(u_{j}-l_{j}).
			\end{equation*}
			
			Let $\bar{c}'_i:=\frac{\sum_{c\in\boldsymbol{c}_i}{|c|}}{n}$ be the average of the absolute values of the elements in the objective $i\in\{1,\dots,p\}$. Note that we understand that $\boldsymbol{c}_i$ is a vector (and not a set) and hence $c\in\boldsymbol{c}_i$ is not a well-defined mathematical notation. However, for simplicity, we keep this notation as is and basically treat each component of the vector as an element. 
			
			We also introduce some features that measures the size of an instance in an indirect way. Specifically, for each $i\in\{1,\dots,p\}$ and $k\in\{13,14,15\}$ we introduce:
			$$
			F_i^{k}=\{\mbox{Avg}(S^{k}_i),\mbox{Min}(S^{k}_i),\mbox{Max}(S^{k}_i),\mbox{Std}(S^{k}_i),\mbox{Median}(S^{k}_i)\},
			$$
			to measure $n$, $p$ and $m$, respectively, where
			$$S^{13}_i:=\cup_{j\in\{1,\dots,p\}\backslash\{i\}}\{\frac{\sum_{c\in\boldsymbol{c}_i}{|c|}}{\bar{c}'_j+1}\},$$ 
			$$S^{14}_i:=\cup_{k\in\{1,\dots,n\}}\{\frac{\sum_{j\in\{1,\dots,p\}\setminus \{i\}}{|c_{jk}|}}{|c_{ik}|+1}\},$$ 
			and $$
			S^{15}_i:=\cup_{k\in\{1,\dots,n\}}\{\frac{\sum_{j=1}^{m}{|a_{jk}|}}{|c_{ik}|+1}\}.$$

			Motivated by the idea using the product of two variables for studying the interaction effect between them, for each $i\in\{1,\dots,p\}$, we introduce:  
			$$
			F_i^{16}=\{\mbox{Avg}(S^{16}_i),\mbox{Min}(S^{16}_i),\mbox{Max}(S^{16}_i),\mbox{Std}(S^{16}_i),\mbox{Median}(S^{16}_i)\},
			$$
			where $S_i^{16}=\cup_{j\in\{1,\dots,p\}\backslash\{i\}}\{\sum_{l=1}^{n} c_{il}c_{jl}\}$.  Similarly, we also define a subset of features based on the \textit{leverage score} $LS_j$ of the variable $j\in \{1,\dots,n\}$ in the matrix $A$. Specifically, for each $i\in\{1\dots,p\}$, we introduce:
			$$F_i^{17}=\sum_{j=1}^n c_{ij} LS_j.$$
			where $LS_j := \frac{||\boldsymbol{a}_{j}||^2}{\sum_{l=1}^{n} ||\boldsymbol{a}_{l}||^2}$ and $\boldsymbol{a}_{j}$ represents column $j$ of matrix $A$ for each $j\in\{1,\dots,n\}$. Let $\text{Avg}(C):=\text{Avg}(\boldsymbol{c}_1,\dots,\boldsymbol{c}_p)$,  $\text{Std}(C):=\text{Std}(\boldsymbol{c}_1,\dots,\boldsymbol{c}_p)$, and $$
			O:=\{(-\infty,-1),(-1,-0.5),(-0.5, 0),(0,0.5),(0.5,1),(1,\infty)\}.
			$$ 
			For each $i\in\{1\dots,p\}$, we define  
			$$F_i^{18}=\cup_{(l,u)\in O}\{car(\boldsymbol{c}_i^{l,u})\}
			$$
			where
			$$\boldsymbol{c}_i^{l,u}:=\{c\in\boldsymbol{c}_i:\text{Avg}(C)+l\ \text{Std}(C)\le c\le\text{Avg}(C)+u\ \text{std}(C)\}.$$ The following observation creates the basis of the remaining features.
			
			\begin{obs}
				\label{obs:scaling}
				Let $\alpha_i>0$ where $i\in\{1,\dots,p\}$ and $\beta_k$ where $k\in\{1,\dots,m\}$ be positive constants. For a MOBLP, its equivalent problem can be constructed as follows:
				\begin{equation}
				\label{eq:Problem2}
				\begin{aligned}
				\min\ &\{\sum_{j=1}^n\alpha_1c_{1j}x_j,\dots,\sum_{j=1}^n\alpha_pc_{pj}x_j\}\\
				\mbox{s.t. } & \sum_{j=1}^n\beta_k a_{kj}x_{j}\le \beta_kb_k && \forall\ k\in\{1,\dots,m\},\\
				&x_j\in \{0,1\}  && \forall\ j\in\{1,\dots,n\}
				\end{aligned}
				\end{equation} 
			\end{obs}
			
			Observation \ref{obs:scaling} is critical because it shows that our ML approach should not be sensitive to positive scaling. So, the remaining features are specifically designed to address this issue. Note that the remaining features are similar to the ones that we have already seen before but they are less sensitive to a positive scaling. 
			
			Let $c^{max}_{i}=\max\{|c_{i1}|,\dots, |c_{in}|\}$ and $\bar{\boldsymbol{c}}_{i}=(\frac{c_{i1}}{c^{max}_{i}},\dots,\frac{c_{in}}{c^{max}_{i}})$. To incorporate the impact of the relative number of zeros, positive, and negative coefficients of the objective functions in the learning process, for each $i\in\{2,\dots,p\}$, we introduce:
			$$F_{i}^{19}=\{\ln\Big(1+\frac{car(\bar{\boldsymbol{c}}_1)^0}{1+ car(\bar{\boldsymbol{c}}_i)^0}\Big)\},$$
			$$F_{i}^{20}=\{\ln\Big(1+\frac{car(\bar{\boldsymbol{c}}_1)^+}{1+car(\bar{\boldsymbol{c}}_i)^+}\Big)\},$$
			and
			$$F_{i}^{21}=\{\ln\Big(1+\frac{car(\bar{\boldsymbol{c}}_1)^-}{1+car(\bar{\boldsymbol{c}}_i)^-}\Big)\},$$
			
			where $car(\bar{\boldsymbol{c}}_i)^0$ is the number of elements in $\bar{\boldsymbol{c}}_i$ with zero values. Also,  $car(\bar{\boldsymbol{c}}_i)^+$ is the number of elements in $\bar{\boldsymbol{c}}_i$ with positive values. Finally, $car(\bar{\boldsymbol{c}}_i)^-$ is the number of elements in $\bar{\boldsymbol{c}}_i$ with negative values.
			
			The following function is helpful for introducing some other features:
			\[
			g(a)=
			\left \{
			\begin{tabular}{cc}
			$a+1$ &  \mbox{ if } $a\ge 0$ \\
			$a-1$ & \mbox{ otherwise}
			\end{tabular}
			\right \}.
			\]
			
			For each $l\in\{1,\dots,m\}$, let $a^{max}_{l}=\max\{|a_{l1}|,\dots, |a_{ln}|\}$, $\bar{\boldsymbol{a}}_{l}=(\frac{a_{l1}}{a^{max}_{l}},\dots,\frac{a_{ln}}{a^{max}_{l}})$ and $\bar{b}_l=\frac{b_l}{a^{max}_{l}}$. To incorporate the relative impact of the magnitude of objective function coefficients, and constraints in the learning process, for each $i\in\{2,\dots,p\}$ and $k\in\{22,23,24,25,26,27\}$, we introduce:
			$$
			F_i^{k}=\{\mbox{Avg}(S^{k}_i),\mbox{Min}(S^{k}_i),\mbox{Max}(S^{k}_i),\mbox{Std}(S^{k}_i),\mbox{Median}(S^{k}_i)\},
			$$
			where  
			$$S_i^{22}=\{\frac{\bar{c}_{11}}{g(\bar{c}_{i1})},\dots, \frac{\bar{c}_{1n}}{g(\bar{c}_{in})}\},$$ $$S_i^{23}=\{\frac{\bar{c}^2_{11}}{g(\bar{c}^2_{i1})},\dots, \frac{\bar{c}^2_{1n}}{g(\bar{c}^2_{in})}\},$$ $$S_i^{24}=\{\sum_{j=1}^n \frac{\bar{c}_{1j}\bar{a}_{1j}}{ng(\bar{c}_{ij})},\dots, \sum_{j=1}^n \frac{\bar{c}_{1j}\bar{a}_{mj}}{ng(\bar{c}_{ij})}\},$$ $$S_i^{25}=\{\sum_{j=1}^n \frac{\bar{c}_{1j}\bar{a}_{1j}}{ng(\bar{c}_{ij})}-\bar{b}_1,\dots, \sum_{j=1}^n \frac{\bar{c}_{1j}\bar{a}_{mj}}{ng(\bar{c}_{ij})}-\bar{b}_m\},$$ $$S_i^{26}=\{\sum_{j=1}^n \frac{\bar{c}^2_{1j}\bar{a}_{1j}}{ng(\bar{c}^2_{ij})},\dots, \sum_{j=1}^n \frac{\bar{c}^2_{1j}\bar{a}_{mj}}{ng(\bar{c}^2_{ij})}\},$$ and $$S_i^{27}=\{\sum_{j=1}^n \frac{\bar{c}^2_{1j}\bar{a}_{1j}}{ng(\bar{c}^2_{ij})}-\bar{b}_1,\dots, \sum_{j=1}^n \frac{\bar{c}^2_{1j}\bar{a}_{mj}}{ng(\bar{c}^2_{ij})}-\bar{b}_m\}.$$
			For the same reason, for each $i\in \{1,\dots,p\}$ and $l\in\{1\dots,p\}\backslash\{i\}$, we introduce:
			$$
			F_{il}^{28}=\big\{\mbox{Avg}(S^{28}_{il}),\mbox{Min}(S^{28}_{il}),\mbox{Max}(S^{28}_{il}),\mbox{Std}(S^{28}_{il}),\mbox{Median}(S^{28}_{il})\big\},
			$$ where 
			$$S_{il}^{28}=\{\frac{\bar{c}_{i1}}{g(\bar{c}_{l1})},\dots, \frac{\bar{c}_{in}}{g(\bar{c}_{ln})}\}.$$ 
			
			Finally, let $\bar{A}_j=\{\bar{a}_{1j},\dots,\bar{a}_{mj}\}$ for each $j\in \{1,\dots,n\}$. For each $i\in\{2,\dots,p\}$, the following subsets of features are also defined for linking the constraints and objective functions:
			
			$$F_i^{29}=\{\sum_{j=1}^n \frac{\bar{c}_{1j}\mbox{Avg}(\bar{A}_j)}{ng(\bar{c}_{ij})},\sum_{j=1}^n \frac{\bar{c}_{1j}\mbox{Min}(\bar{A}_j)}{ng(\bar{c}_{ij})},\sum_{j=1}^n \frac{\bar{c}_{1j}\mbox{Max}(\bar{A}_j)}{ng(\bar{c}_{ij})},\sum_{j=1}^n \frac{\bar{c}_{1j}\mbox{Std}(\bar{A}_j)}{ng(\bar{c}_{ij})},\sum_{j=1}^n \frac{\bar{c}_{1j}\mbox{Median}(\bar{A}_j)}{ng(\bar{c}_{ij})}\},$$	
			\begin{align*}
			F_i^{30}=&\{\sum_{j=1}^n \frac{\bar{c}_{1j}\mbox{Avg}(\bar{A}_j)}{ng(\bar{c}_{ij})}-\mbox{Avg}(\bar{b}_j),\sum_{j=1}^n \frac{\bar{c}_{1j}\mbox{Min}(\bar{A}_j)}{ng(\bar{c}_{ij})}-\mbox{Min}(\bar{b}_j),\sum_{j=1}^n \frac{\bar{c}_{1j}\mbox{Max}(\bar{A}_j)}{ng(\bar{c}_{ij})}-\mbox{Max}(\bar{b}_j),\\
			&\sum_{j=1}^n \frac{\bar{c}_{1j}\mbox{Std}(\bar{A}_j)}{ng(\bar{c}_{ij})}-\mbox{Std}(\bar{b}_j),\sum_{j=1}^n \frac{\bar{c}_{1j}\mbox{Median}(\bar{A}_j)}{ng(\bar{c}_{ij})}-\mbox{Median}(\bar{b}_j)\},
			\end{align*}		
			$$F_i^{31}=\{\sum_{j=1}^n \frac{\bar{c}^2_{1j}\mbox{Avg}(\bar{A}_j)}{ng(\bar{c}^2_{ij})},\sum_{j=1}^n \frac{\bar{c}^2_{1j}\mbox{Min}(\bar{A}_j)}{ng(\bar{c}^2_{ij})},\sum_{j=1}^n \frac{\bar{c}^2_{1j}\mbox{Max}(\bar{A}_j)}{ng(\bar{c}^2_{ij})},\sum_{j=1}^n \frac{\bar{c}^2_{1j}\mbox{Std}(\bar{A}_j)}{ng(\bar{c}^2_{ij})},\sum_{j=1}^n \frac{\bar{c}^@_{1j}\mbox{Median}(\bar{A}_j)}{ng(\bar{c}^2_{ij})}\},$$	
			\begin{align*}
			F_i^{32}=&\{\sum_{j=1}^n \frac{\bar{c}^2_{1j}\mbox{Avg}(\bar{A}_j)}{ng(\bar{c}^2_{ij})}-\mbox{Avg}(\bar{b}_j),\sum_{j=1}^n \frac{\bar{c}^2_{1j}\mbox{Min}(\bar{A}_j)}{ng(\bar{c}^2_{ij})}-\mbox{Min}(\bar{b}_j),\sum_{j=1}^n \frac{\bar{c}^2_{1j}\mbox{Max}(\bar{A}_j)}{ng(\bar{c}^2_{ij})}-\mbox{Max}(\bar{b}_j),\\
			&\sum_{j=1}^n \frac{\bar{c}^2_{1j}\mbox{Std}(\bar{A}_j)}{ng(\bar{c}^2_{ij})}-\mbox{Std}(\bar{b}_j),\sum_{j=1}^n \frac{\bar{c}^2_{1j}\mbox{Median}(\bar{A}_j)}{ng(\bar{c}^2_{ij})}-\mbox{Median}(\bar{b}_j)\},
			\end{align*}
			where $\bar{c}^2_{ij}=\bar{c}_{ij}\bar{c}_{ij}$ for each $i\in\{1,\dots,p\}$ and $j\in\{1\dots,n\}$. 
\end{APPENDICES}	
		
% Appendix here
% Options are (1) APPENDIX (with or without general title) or 
%             (2) APPENDICES (if it has more than one unrelated sections)
% Outcomment the appropriate case if necessary
%
% \begin{APPENDIX}{<Title of the Appendix>}
% \end{APPENDIX}
%
%   or 
%
% \begin{APPENDICES}
% \section{<Title of Section A>}
% \section{<Title of Section B>}
% etc
% \end{APPENDICES}

% References here (outcomment the appropriate case) 

% CASE 1: BiBTeX used to constantly update the references 
%   (while the paper is being written).
%\bibliographystyle{informs2014} % outcomment this and next line in Case 1
%\bibliography{<your bib file(s)>} % if more than one, comma separated

% CASE 2: BiBTeX used to generate mypaper.bbl (to be further fine tuned)
%\input{mypaper.bbl} % outcomment this line in Case 2

\end{document}